\documentclass{article}

\PassOptionsToPackage{square,numbers}{natbib}

    \usepackage[final]{neurips_2022}

\usepackage[utf8]{inputenc} % allow utf-8 input
\usepackage[T1]{fontenc}    % use 8-bit T1 fonts
\usepackage[hidelinks]{hyperref}       % hyperlinks

\usepackage{url}            % simple URL typesetting
\usepackage{booktabs}       % professional-quality tables
\usepackage{amsfonts}       % blackboard math symbols
\usepackage{nicefrac}       % compact symbols for 1/2, etc.
\usepackage{microtype}      % microtypography
\usepackage[table]{xcolor}         % colors
\usepackage{pifont}     
\usepackage{siunitx}
\usepackage{wrapfig}

\usepackage{amsmath}
\usepackage{amsfonts}
\usepackage[T1]{fontenc}
\usepackage{xspace}
\usepackage{graphicx}
\usepackage[caption=false,position=top]{subfig}
\usepackage{multirow}
\usepackage{geometry}
\usepackage{tabularx}
\usepackage{bbm}
\usepackage{booktabs}% http://ctan.org/pkg/booktabs
\usepackage{makecell}% http://ctan.org/pkg/makecell
\usepackage{thmtools}
\usepackage{thm-restate}

\newcommand{\bN}{\ensuremath{\mathbb N}\xspace}

\newcommand{\bE}{\ensuremath{\mathbb E}\xspace}

\newcommand{\cS}{\ensuremath{\mathcal S}\xspace}
\newcommand{\cA}{\ensuremath{\mathcal A}\xspace}

\newcommand{\cO}{\ensuremath{\mathcal O}\xspace}
\newcommand{\rl}{reinforcement learning\xspace}
\newcommand{\bsuite}{\texttt{\textbf{bsuite}}\xspace}
\newcommand{\colosseum}{\texttt{\textbf{Colosseum}}\xspace}
\newcommand{\prand}{\texttt{p\_rand}\xspace}
\newcommand{\plazy}{\texttt{p\_lazy}\xspace}
\newcommand{\subgaps}{sum of the reciprocals of the sub-optimality gaps\xspace}

\DeclareMathOperator*{\var}{Var}

\DeclareMathOperator*{\argmin}{arg\,min}

\title{Hardness in Markov Decision Processes:\\Theory and Practice}
\author{%
  Michelangelo Conserva\\
  Queen Mary University of London\\
  London, United Kingdom \\
  \texttt{m.conserva@qmul.ac.uk} \\
  \And
  Paulo Rauber\\
  Queen Mary University of London\\
  London, United Kingdom \\
  \texttt{p.rauber@qmul.ac.uk} \\
}

\begin{document}

\maketitle

\begin{abstract}
Meticulously analysing the empirical strengths and weaknesses of reinforcement learning methods in hard (challenging) environments is essential to inspire innovations and assess progress in the field. In tabular reinforcement learning, there is no well-established standard selection of environments to conduct such analysis, which is partially due to the lack of a widespread understanding of the rich theory of hardness of environments. The goal of this paper is to unlock the practical usefulness of this theory through four main contributions.
First, we present a systematic survey of the theory of hardness, which also identifies promising research directions.
Second, we introduce \texttt{Colosseum}, a pioneering package that enables empirical hardness analysis and implements a principled benchmark composed of environments that are diverse with respect to different measures of hardness.
Third, we present an empirical analysis that provides new insights into computable measures.
Finally, we benchmark five tabular agents in our newly proposed benchmark.
While advancing the theoretical understanding of hardness in non-tabular reinforcement learning remains essential, our contributions in the tabular setting are intended as solid steps towards a principled non-tabular benchmark. Accordingly, we benchmark four agents in non-tabular versions of \colosseum environments, obtaining results that demonstrate the generality of tabular hardness measures.
\end{abstract}

\section{Introduction}

Reinforcement learning studies a setting where an agent interacts with an environment by observing states, receiving rewards, and selecting actions with the objective of optimizing a reward-based criterion.
The field has attracted significant interest in recent years after striking performances obtained in board games \citep{silver2018general} and video games \citep{vinyals2019grandmaster, berner2019dota}.
Solving these grand challenges constitutes a pivotal milestone in the field.
However, the corresponding agents require efficient simulators due to their high sample complexity, i.e., the number of observations that they require to optimize a reward-based criterion in an unknown environment.
Outside of games, many important applications in healthcare, robotics, logistics, finance, and advertising can also be naturally formulated as \rl problems. However, simulators for these scenarios may not be available, reliable, or efficient.

The development of \rl methods that explore efficiently has long been considered one of the most crucial efforts to reduce sample complexity. Meticulously evaluating the strengths and weaknesses of such methods is essential to assess progress and inspire new developments in the field. Such empirical evaluations are performed through benchmarks composed of a selection of environments and evaluation criteria. Ideally, this selection should be based on theoretically principled reasoning that considers the \textit{hardness} of the environments and the \textit{soundness} of the evaluation criteria.

In non-tabular \rl, where the number of states is large (and potentially infinite), there is no theory of hardness except for a few restricted settings. Consequently, the selection of environments in current benchmarks \citep{osband2020bsuite, rajan2021mdp} relies solely on the experience of their authors. Although such benchmarks are certainly valuable, there is no guarantee that they contain a sufficiently diverse range of environments and that they are effectively able to quantify agent capabilities.
In contrast, in tabular \rl, where the number of states and actions is finite, a rich theory of hardness of environments is available. Perhaps surprisingly, a principled benchmark based on this theory has so far been absent. There are at least two reasons for this absence. First, these hardness measures have been developed to provide theoretical guarantees for \rl methods, and not considered as directly useful for practical purposes. Second, the lack of a unifying presentation and critical comparison between different measures has limited their potential impact.

The goal of this paper is to establish the importance of hardness measures outside the context of theoretical \rl and unlock their practical usefulness with four main contributions.
In Section~\ref{sec:theory}, we present a systematic survey of the theory of hardness for Markov decision processes. This survey serves as an introduction, tutorial, and, equally importantly, identifies gaps in the current theoretical landscape that suggest promising directions for future work.
In Section~\ref{sec:colosseum}, we introduce \colosseum, a pioneering Python package that enables the empirical investigation of hardness and implements a principled benchmark for the four most widely studied tabular \rl settings. The selected environments aim to maximize diversity with respect to two important measures of hardness, thus providing a varied set of challenges for which a \textit{precise} characterization of hardness is available. 
In Section~\ref{sec:analysis}, we present an empirical comparison between three theoretical, yet efficiently computable, measures of hardness. Our analysis provides insights into which aspects of hardness are best captured by each of the measures and identifies desirable qualities for future measures.
Finally, in Section~\ref{sec:benchmark}, we report the results of five agents with theoretical guarantees in our novel (principled) benchmark, which allows us to empirically validate the quality of the selection methodology by demonstrating that harder environments effectively lead to worse performances.

Although this paper is concerned with the \emph{tabular} setting, for which principled measures of hardness are available, we intend our contributions as milestones towards the future development of theoretical and empirical measures of hardness for non-tabular \rl.
Accordingly, Section~\ref{sec:colosseum} shows how tabular hardness measures can be used in a widespread class of environments that includes non-tabular versions of the environments in the \colosseum benchmark, which enables studying how these measures relate to the performance of four agents from the non-tabular $\texttt{bsuite}$ benchmark \citep{osband2020bsuite}.

\section{Hardness in Theory} \label{sec:theory}

Section~\ref{sec:preliminaries} introduces our notation and important definitions. Section~\ref{sec:characterization} presents our survey of the theoretical landscape of measures of hardness, which includes a novel categorization of existing measures. Section~\ref{sec:complete} highlights the weaknesses of existing measures and introduces the concept of a \emph{complete} measure of hardness, which we believe should be the focus of future developments.

\subsection{Preliminaries} \label{sec:preliminaries}

Let $\Delta(\mathcal{X})$ denote the set of probability distributions over a set $\mathcal{X}$.
A finite Markov decision process (MDP) is a tuple $\text{M} = \left(\cS, \cA, P, P_0, R\right)$, where 
\cS is the finite set of states,
\cA is the finite set of actions,
$P : \cS \times \cA \to \Delta(\cS)$ is the transition kernel,
$P_0 \in \Delta(\cS)$ is the initial state distribution, and
$R : \cS \times \cA \to \Delta\left([0,1]\right)$ is the reward kernel \citep{puterman2014markov}.
Given an optimization horizon $T$, a \rl agent aims to find a (possibly stochastic) policy $\pi : \cS \to \Delta(\cA)$ that optimizes a reward-based criterion. The MDP M is unknown to the agent but can be learned through experience. The interaction between the agent and the environment starts when the initial state $s_{0} \sim P_0$ is drawn. For any $0 \leq t < T$, the agent samples an action $a_{t} \sim \pi_t(s_t)$ from its current policy $\pi_t$, and the environment draws the next state $s_{t+1} \sim P(s_t,a_t)$ and reward $r_{t+1} \sim R(s_t,a_t)$.
An MDP is \textit{episodic} with time horizon $H$ when the state $s_{t+1}$ is drawn from the initial state distribution whenever $t+1 \equiv 0 \pmod H$ and \textit{continuous} otherwise. In the continuous setting, a factor $\gamma \in (0,1)$ is used to discount future rewards in the \emph{discounted} setting, and future rewards are averaged across time in the \emph{undiscounted} setting.
For a policy $\pi$, an MDP M induces a Markov chain \citep{levin2017markov} with transition probabilities $\text P_{s\to s'}^\pi(\text{M}) = \sum_{a \in \cA}\pi(a\mid s) P(s'\mid s, a)$ whose stationary distribution is denoted by $\mu^\pi$.
MDPs can be classified into three communication classes \citep{kallenberg2002classification}.
An MDP is \textit{ergodic} if the Markov chain induced by any deterministic policy is ergodic.
An MDP is \textit{communicating} if, for every two states $s,s' \in \cS$, there is a deterministic policy that induces a Markov chain where $s$ is accessible from $s'$ and vice versa.
An MDP is \textit{weakly communicating} when there is a partition $(\mathcal{C}, \cS \setminus \mathcal{C})$ of the set of states \cS such that, for every two states $s,s' \in \mathcal{C}$, there is a deterministic policy that induces a Markov chain where $s$ is accessible from $s'$ and vice versa, and every state $s \in \cS \setminus \mathcal{C}$ is transient in every Markov chain induced by any deterministic policy.
Every ergodic MDP is communicating, and every communicating MDP is weakly communicating.

The episodic state-action value function is given by $Q^\pi_{h, \texttt{epi}}(s, a) := \bE \left[ \sum_{t=h + 1}^H r_t | s_h = s, a_h = a\right]$, and the episodic state value function is given by $V^\pi_{h, \texttt{epi}}(s) := \sum_{a} \pi(a \mid s) Q^\pi_{h, \texttt{epi}}(s, a)$. The discounted state-action value function is given by $Q^\pi_\gamma(s, a) := \bE \left[ \sum_{t=1}^\infty \gamma^{t-1} r_t \mid s_{0} = s, a_0 = a \right]$, and the discounted state value function is given by $V^\pi_\gamma(s) := \sum_{a} \pi(a \mid s) Q^\pi_\gamma(s, a)$. These expectations are taken w.r.t. the policy $\pi$, the transition kernel $P$, and the reward kernel $R$.
In the undiscounted setting, the value of every state(-action) is the same, since  rewards are averaged across infinite time steps. In that case, we define the expected average reward as $\rho^\pi := \lim_{T\to\infty} \frac{1}{T} \sum_{t=1}^T \bE\left[ r_t \right]$, which is also given by $\rho^\pi = \langle \mu^\pi, \text{R}^\pi \rangle$, where $\text{R}^\pi$ is a vector such that $\text{R}^\pi_s = \bE_{a \sim \pi(s)} R(s, a)$ is the expected reward obtained when following $\pi$ from state $s$.
An optimal policy $\pi^*$ obtains maximum value for every state.
We assume with little loss of generality that the optimal policy is unique.
We drop the subscripts when the
setting is clear from the context, and write $*$ to denote $\pi^*$ in superscripts.

The most widely studied performance criteria are the \textit{expected cumulative regret}, which yields the \textit{regret minimization} setting \citep{auer2006logarithmic}, and the \textit{sample efficiency of exploration}, which yields the Probably Approximately Correct Reinforcement Learning (PAC-RL) setting \citep{kakade2003sample}.
In simple terms, the regret measures the loss in reward due to the execution of a sub-optimal policy.
In contrast, the sample efficiency measures how many interactions the agent requires to approximately learn $\pi^*$ with high probability.
The main difference between the two criteria is that the rewards obtained during the interactions with the MDP are not considered in PAC-RL, whereas all rewards contribute to the cumulative regret. Therefore, PAC-RL agents can generally afford more aggressive exploration. 

\subsection{Characterization of hardness} \label{sec:characterization}

We distinguish the hardness of MDPs into two kinds of complexity, the \textit{visitation} complexity and the \textit{estimation} complexity.
The visitation complexity relates to the difficulty of visiting all the states, and the estimation complexity relates to the discrepancy between the optimal policy and the best policy an agent can derive from a given estimate of the transition and reward kernels.
The two complexities are complementary in the sense that the former quantifies the hardness of gathering samples from the state space while the latter quantifies the hardness of producing a highly rewarding policy given the samples.
In the literature, we identify two approaches that aim to capture the mentioned complexities.
The first approach, which we call \textit{Markov chain-based}, considers that an MDP is an extension of a Markov chain where transition probabilities can be changed based on direct interventions by an agent \citep{bellman1957markovian}.
This approach is well suited to capture the hardness that comes from the visitation complexity since it considers the properties of transition kernels.
The second approach, which we call \textit{value-based}, considers the discrepancy between the optimal policy and the best policy an agent can derive from a value function point estimate.
Note that the information contained in such point estimate is lower than the one in kernels estimate and that the difficulty of obtaining an accurate estimate of the value function is not considered in this approach. 
Therefore, the value-based approach is only able to partially capture the estimation complexity.
A striking fact that highlights this shortcoming is that almost every value-based measure of hardness is independent of the variability of the reward kernel.
Therefore, given an MDP $\text{M}'$ that is obtained by increasing the reward kernel variability of an MDP $\text{M}$, value-based measures of hardness assign the same level of hardness to $\text{M}$ and $\text{M}'$.

Markov chain properties and value functions depend on a fixed policy, which presents two natural choices to derive measures of hardness. The first choice considers the optimal policy, which typically leads to a measure that considers a best-case scenario. The second choice considers a policy that maximizes a criterion that characterizes a worst-case scenario. For instance, a policy that maximizes such criterion may spend its time in a region of the state space that is not relevant for learning $\pi^*$.

\newpage
\subsubsection{Markov chain-based measures of hardness}

\paragraph{Mixing time.}
The mixing time of a Markov chain with stationary distribution $\mu$ is defined as
\begin{equation} \label{eq:mtmc}
    t_\mu := \inf\{n \mid \sup\limits_{s\in\cS} d_{\text{TV}}\left( 
    p_{s}^n
    , \mu\right) \leq 0.25 \},
\end{equation}
where $d_{\text{TV}}$ is the total variation distance between distributions
and the vector $p_{s}^n$ represents the distribution over states after $n$ steps starting from state $s$.
The value $0.25$ is conventionally established in the Markov chain literature for the definition of the mixing time.
For ergodic and aperiodic Markov chains, $\lim_{n\to\infty} p_{s}^n = \mu$ for every state $s$, so the mixing time is the number of steps a Markov chain takes to produce samples that are \emph{close} to being distributed according to the stationary distribution $\mu$.

For instance, a Markov chain with a non-negligible probability of transitioning from every state to every state is quickly mixing. In contrast, the mixing time can be very long in chains where the state space has several distinct regions each of which is well connected but where transitions between regions have low probability \cite{koller2009probabilistic}.
\citet{kearns2002near} propose an extension of the mixing time to MDPs that considers the maximum mixing time across policies $t := \sup_\pi t_{\mu^\pi}$ in the undiscounted setting. Although the mixing time plays an important role in  that setting, since the average reward obtained by policy $\pi$ is given by $\rho^\pi = \langle \mu^\pi, \text{R}^\pi \rangle$, it is not a generally good measure of hardness. % since it provides a worst-case indication of the number of time steps necessary for a policy to reach its stationary distribution, i.e. the moment starting from which the agent can estimate its average reward.
First, it does not capture the visitation complexity, since it neglects the fact that the agent may direct its exploration through a choice of policy. Second, it does not capture any significant aspect of optimal policy estimation, which does not require stationary distribution samples from every policy.

\paragraph{Diameter.} The diameter is fundamentally related to the number of time steps required to transition between states. In the continuous setting, the diameter $D$ is most commonly defined as
\begin{equation}
    D := \sup\limits_{s_1 \neq s_2} \inf_\pi T^\pi_{s_1 \to s_2},
\end{equation}
where $T^\pi_{s_1 \to s_2}$ is the expected number of time steps required to reach state $s_2$ from state $s_1$ when following policy $\pi$ \citep{jaksch2010near}.
Intuitively, $D$ is the worst-case expected number of time steps required to transition between two states when following the best policy for that purpose.
Related definitions are 
$D_{\text{worst}} := \sup\limits_\pi\sup\limits_{s_1 \neq s_2} T^\pi_{s_1 \to s_2}$ and
$D_{\text{opt}} := \inf\limits_\pi\sup\limits_{s_1 \neq s_2} T^\pi_{s_1 \to s_2}$,
which imply  $D \leq D_{\text{opt}} \leq D_{\text{worst}}$  \citep{bartlett2012regal}.

In the episodic setting, we define the diameter by augmenting each state $s$ with the current in-episode time step $h$ and considering the diameter of this augmented MDP in the continuous setting. Note that $ T^\pi_{(s_1,h_1) \to (s_2,h_2)}$ can be larger than the episode length $H$, which means that (on average) more than one episode may be required to transition from state $(s_1, h_1)$ to state $(s_2, h_2)$. This happens whenever $h_2 < h_1$, which may be undesirable if the intent is to focus on the expected number of time steps required to transition between states from the same episode. The diameter is always infinite in weakly-communicating MDPs if the supremum is not restricted to states in the recurrent class $\mathcal{C}$ and is always finite in the episodic setting (where every state is reachable within an episode).
A large diameter can be caused by high stochasticity. The diameter is very apt at measuring visitation complexity, since it captures the effort required to deliberately move between states. % in the worst case scenario.
However, it neglects the reward kernel, and so has limited capacity to measure the estimation complexity.

\paragraph{Distribution mismatch coefficient.}
The \textit{distribution mismatch coefficient} (DMC) has been defined for the continuous undiscounted \citep{wei2020model} and continuous discounted \citep{agarwal2021theory} cases respectively as
\begin{equation*}
    \text{DMC} := \sup\limits_\pi\sum_{s\in\cS} \frac{\mu^*_s}{\mu^\pi_s}  \quad\quad \text{and} \quad\quad \text{DMC}_{s_0} := \sup\limits_{s\in\cS} \frac{d^*_{s_0}(s)}{P_0(s)},
\end{equation*}
where $d^*_{s_0}(s) = (1 - \gamma)\sum_{t=0}^\infty \gamma^t \text{Pr}(s_t = s \mid s_0)$ is the discounted state
visitation distribution of the optimal policy given an initial state $s_0$.
Note that the DMC is guaranteed to be finite only for ergodic MDPs.
In communicating MDPs, there is at least one policy $\pi$ whose stationary distribution assigns probability zero to some states.
MDPs whose optimal stationary distribution $\mu^*$ has its probability mass concentrated on a few states tend to have a large DMC. In contrast,  when $\mu^*$ is closer to being uniformly distributed across states, the DMC tends to be small.
For small values of DMC,
as every $\mu^\pi$ is \textit{close} to $\mu^*$,
the agent will gather samples from the optimal stationary distribution regardless of its current policy, which may enable quick learning.
In contrast, for large values of DMC, the agent needs to actively seek a policy that gathers such samples.
The DMC is not well suited to quantify the visitation complexity, since it fails to capture the difficulty of visiting all states. The DMC also does not capture the estimation complexity, since it does not account for the stochasticity of the environment, which is related to the number of samples required to make accurate estimations.

\subsubsection{Value-based measures of hardness}

\paragraph{Action-gap regularity.}
Given an estimate $\hat Q^*$ of the optimal state-action value function $Q^*$, the \emph{greedy policy} with respect to $\hat Q^*$ always chooses an action associated with the highest state-action value.
Whether or not such a policy is optimal depends exclusively on the ordering of the estimates for a given state rather than their accuracy.
For instance, assuming that $a^*$ is the optimal action for every state $s$, a greedy agent would act optimally if $\hat Q(s,a^*) > \hat Q(s,a')$ for every action $a' \neq a^*$, even if $|Q^*(s,a) - \hat Q(s,a)| \gg 0$ for every action $a$. The action-gap regularity $\zeta$ is a measure of hardness that leverages this principle through the theory of hardness for classification algorithms \citep{farahmand2011action}. However, this measure is only defined for two actions, and so has exceptionally limited applicability.

\paragraph{Environmental value norm.} The (discounted) environmental value norm  $C^\pi_{\gamma}$ is defined as 
\begin{equation*}
    C^\pi_{\gamma} := \sup\limits_{(s,a)} \sqrt{\var\limits_{s' \sim P(s,a)} V^{\pi}_\gamma(s')}.
\end{equation*}
This quantity can be similarly defined in the undiscounted setting \cite{maillard2014hard}. In words, the environmental value norm captures the one-step variance of the value function $V^{\pi}$ for a given policy $\pi$. In the episodic setting, a closely related measure called \emph{maximum per-step conditional variance} $C^\pi_\text{H}$ \citep{zanette2019tighter} is defined as 
\begin{equation*}
    C^\pi_\text{H} := \sup\limits_{(s,a,h)}\left( \var R(s,a) + \var\limits_{s'\sim P(s,a)} V^\pi_{h+1}(s') \right).
\end{equation*}
Alternatively, as with the diameter, it is also possible to define this quantity by augmenting each state $s$ with the current in-episode time step $h$ and considering the environmental value norm of this augmented MDP in the continuous setting. For every policy, the environmental value norm is equal to zero when the transition kernel is deterministic. However, a highly stochastic MDP may still have a small environmental value norm, since this norm captures the variance of the state value function rather than
the stochasticity of the transition kernel.
\citet{maillard2014hard} suggest using the environmental value norm of the optimal policy $\pi^{*}$ as a measure of hardness. The main strength of such measure is that the variability of the optimal value function captures an important aspect of 
the estimation complexity.
However, this measure of hardness neglects the visitation complexity.

\paragraph{Sub-optimality gap.} The sub-optimality gap is defined in the continuous and episodic settings as
\begin{equation*}
    \Delta(s,a) := V^*(s) - Q^*(s,a)
    \quad\text{and}\quad
    \Delta_h(s,a) := V_h^*(s) - Q_h^*(s,a),
\end{equation*}
respectively. Since $V^*(s) = \max_a Q^*(s,a)$ for every state $s$, the sub-optimality gap $\Delta(s,a)$ measures the difference in expected return between selecting the optimal action for state $s$ and selecting the action $a$.
Intuitively, identifying a suboptimal action $a'$ in a given state $s$ is easier if the gap $\Delta(s,a')$ is large. \citet{simchowitz2019non} identifies the \subgaps $\sum_{(s,a) \mid \Delta(s,a) \neq 0} \frac{1}{\Delta(s,a)}$ as a measure of hardness.
They also demonstrate the importance of the sub-optimality gaps by showing that recent optimistic algorithms necessarily incur in a cumulative regret proportional to the smallest nonzero sub-optimality gap in the episodic setting. However, note that approximating the optimal value function (and identifying a near-optimal policy) is particularly easy when every sub-optimality gap is small. Consequently, the (PAC-RL) sample complexity is likely to decrease when the \subgaps increases. Furthermore, this measure does not explicitly capture visitation complexity and is prone to severe numerical issues.

\begin{table}[h!]
    \centering
    \caption{Computational complexity of generally applicable measures up to logarithmic factors.}
    \resizebox{\textwidth}{!}{%
    \begin{tabular}{ccccc}
        \toprule
        \multicolumn{3}{c}{Markov chain-based measures} & \multicolumn{2}{c}{Value-based measures}\\
        \arrayrulecolor{black!35}\cmidrule(lr){1-3}\cmidrule(lr){4-5}%
        Mixing time & Diameter & Distribution mismatch coefficient & Environmental value norm & Sub-optimality gaps\\
        \arrayrulecolor{black!15}\cmidrule(lr){1-3}\cmidrule(lr){4-5}%
        $\boldsymbol ?$ & $\tilde O(|\cS|^{3.5}|\cA|)$ & $\boldsymbol ?$ &  $\tilde O(|\cS|^2|\cA|(1 - \gamma)^{-1})$ & $\tilde O(|\cS|^2|\cA|(1 - \gamma)^{-1})$\\
        \bottomrule
    \end{tabular}}
    \label{tab:hmcc}
\end{table}

\subsubsection{Future directions} \label{sec:complete}

Current hardness measures suffer from three principal issues.
They are not designed to be efficiently computable (see Table \ref{tab:hmcc} and Appendix \ref{app:cc}),
they are limited in their ability to simultaneously capture visitation complexity and estimation complexity, and
they are oblivious to the distinct challenges presented by different performance criteria.
For instance, while regret minimizing agents must be cautious not to incur in large regret during learning (for example, by not revisiting lowly rewarding states that are not followed by highly rewarding states), PAC-RL agents have the flexibility to incur in large regret as long as they end up with a near-optimal policy.
Therefore, the visitation complexity should differ across settings. The fact that current measures disregard this distinction is concerning, since they should account for the specific difficulty of the optimization task. These issues are not discussed in previous work, since measures of hardness have not been considered relevant outside the context of deriving theoretical performance guarantees for \rl agents.

In order to address these issues, we believe that future work should focus on developing efficiently computable (non-trivial) hardness measures that (approximately) meet the following novel definition.

\textbf{Definition \ref{sec:theory}.1 (Complete measure of hardness)} A measure $\theta : \mathcal{M} \to \mathbb{R}^+$ is \textit{complete} for an MDP class $\mathcal{M} $ and criterion $\psi$ (sample complexity or cumulative regret) if, for every pair $\text{M}_1, \text{M}_2 \in \mathcal{M}$ and \emph{near-optimal agent} $A^*$ that achieves the criterion lower bound of class $\mathcal{M}$ up to logarithmic factors\footnote{
For example, the $\Omega(|\cS||\cA|H^2\epsilon^{-2})$ sample complexity bound for the episodic communicating setting \citep{dann2015sample}.
}%
\hspace{-.35em},\hspace{.25em}$\theta(\text{M}_1) > \theta(\text{M}_2)$ implies $\tilde \psi(\text{M}_1, A^*) > \tilde \psi(\text{M}_2, A^*)$,
where $\tilde \psi$ hides logarithmic factors.

Combining existing measures that capture visitation complexity and estimation complexity is a viable first step in that direction.
Recently, \citet{wagenmaker2021beyond} have pioneered this approach in the episodic setting
by proposing the \textit{gap-visitation complexity},
\begin{equation}
    \text{GVP}(\epsilon) := \sum_{h=0}^H\inf\limits_\pi\sup\limits_{s,a}\inf\left( \frac{1}{w^\pi_h(s,a)\Delta_h(s,a)^2}, \frac{W_h(s)^2}{w^\pi_h(s,a)\epsilon^2} \right),
\end{equation}
where $w_h^\pi(s,a) := P_\pi(s_h = s, a_h = a)$ is the probability of visiting state-action pairs $(s,a)$ at in-episode time step $h$ when following policy $\pi$, $W_h(s) := \sup_\pi P_\pi(s_h = s)$ is the maximum reachability of state $s$ at in-episode time step $h$, and $\epsilon$ is a parameter related to the optimality of the output policy in the PAC-RL setting.
The strength of this measure is that it weights the sub-optimality action gaps with measures of visitation complexity, $w_h^\pi$ and $W_h$.
This captures the difficulty induced by the critical states that are both hard to reach and for which it is hard to estimate the best action.
However, the gap-visitation complexity fails to be a generally applicable hardness measure.
It depends on the PAC-RL setting-specific parameter $\epsilon$, it is restricted to the finite horizon setting (and can not be extended to the continuous setting), and is not efficiently computable.

\section{Hardness in Practice} \label{sec:practice}

\subsection{Colosseum} \label{sec:colosseum}

This section briefly introduces \colosseum, a pioneering Python package that bridges theory and practice in tabular reinforcement learning while also being applicable in the non-tabular setting. More details about the package can be found in Appendix \ref{app:colosseum} and in the project website.\footnote{Available at \url{https://michelangeloconserva.github.io/Colosseum}.}

As a hardness analysis tool, \colosseum 
identifies the communication class of MDPs,
assembles insightful visualizations and logs of interactions between agents and MDPs, computes three measures of hardness (environmental value norm, \subgaps, and diameter, whose computation requires a novel solution described in App. \ref{app:diameter}).
Eight MDP families are available for experimentation. Some are traditional families (\texttt{RiverSwim} \citep{strehl2008analysis}, \texttt{Taxi} \citep{dietterich2000hierarchical}, and \texttt{FrozenLake}) while others are more recent (\texttt{MiniGid} environments \citep{gym_minigrid}). Additionally, \texttt{DeepSea} \citep{osband2019deep} was included as a hard exploration family of problems, and the \texttt{SimpleGrid} family is composed of simplified versions of the \texttt{MG--Empty} environment. By controlling the parameters of MDPs from each family (further detailed in App. \ref{app:mdps}), it is easy to create an MDP with any desired hardness.

As a benchmarking tool, \colosseum is unique in its strong connection with theory.
For instance, in contrast to non-tabular benchmarks, \colosseum computes theoretical evaluation criteria such as the expected cumulative regret and the expected average future reward, which can be used to exactly evaluate the performance criterion of regret minimizing agents.
The benchmark covers the most commonly studied \rl settings: \emph{episodic ergodic}, \emph{episodic communicating}, \emph{continuous ergodic}, and \emph{continuous communicating}.
For each setting, we have selected twenty MDPs that are diverse with respect to their diameters and environmental value norms as proxies for different combinations of visitation complexity and estimation complexity.
Figure~\ref{fig:benchmark} in Appendix \ref{app:benchmark} shows how each of these MDPs varies according to these measures, and Section \ref{sec:benchmark} empirically validates this selection by showing that harder MDPs correspond to worse agent performance.
Notably, the theoretically backed selection of MDPs and the rigorous evaluation criteria make the \colosseum benchmark the most exhaustive in tabular \rl, since previous evaluations were conducted empirically in a few MDPs (such as Taxi or RiverSwim).

\colosseum also allows testing of non-tabular agents by leveraging the BlockMDP model \citep{du2019provably}.
BlockMDPs equip tabular MDPs with an \textit{emission map} that is a (possibly stochastic) mapping $q : \cS \to \Delta(\cO)$ from the finite state space $\mathcal{S}$ to a (possibly infinite) \textit{observation space} $\cO$.
Agents interacting with BlockMDPs are only provided with observations, so non-tabular methods are generally required.
Many commonly used non-tabular MDPs (such as Minecraft \citep{misra2020kinematic}) can be straightforwardly encoded as BlockMDPs using the \colosseum MDP families.
\colosseum implements a diverse set of deterministic emission maps and allows combining them with different sources of noise.
Appendix \ref{app:bmdp} further details BlockMDPs and the available emission maps.

\subsection{Empirical analysis of hardness measures} \label{sec:analysis} 

For brevity, this section only presents results of hardness measures in the \texttt{MiniGridEmpty} family of environments in the episodic setting. Appendix~\ref{app:analysis} presents the full outcome of the empirical analysis.

A \texttt{MiniGridEmpty} MDP is a grid world where an agent has three available actions: moving forward, rotating left, and rotating right. An agent is rewarded for being in a few specific states and receives no reward in every other state. Appendix \ref{app:minigrid_mdps} provides more details about this family of environments.

In our investigation, we consider four scenarios that highlight the different aspects of MDP hardness. 

\textbf{Scenario 1.} We vary the probability \prand that an MDP executes a random action instead of the action selected by an agent. As \prand approaches one, value estimation becomes easier, since outcomes depend less on agent choices. However, intentionally visiting states becomes harder.

\textbf{Scenario 2.} We vary the probability \plazy that an MDP stays in the same state instead of executing the action selected by an agent. Contrary to increasing \prand, increasing \plazy never benefits exploration. Increasing \plazy decreases estimation complexity and increases visitation complexity.

\textbf{Scenario 3 and 4.}  We vary the number of states across MDPs from the same family. In scenario 4, we also let \prand $= 0.1$ to study the impact of stochasticity. In these scenarios, increasing the number of states simultaneously increases the estimation complexity and the visitation complexity.

In every scenario, hardness measures are compared with the cumulative regret of a \emph{near-optimal} agent tuned for each specific MDP (see App. \ref{app:analysis}). This regret serves as an optimistic measure of hardness. Appendix \ref{app:reg_norm} describes how these measures are normalized. %, which \emph{solely} allows comparing trends.
Note that, due to normalization, the plots should only be compared in terms of trends (growth rates) rather than absolute values.

\begin{figure}[htb]
    \centering
    \includegraphics[width=\textwidth]{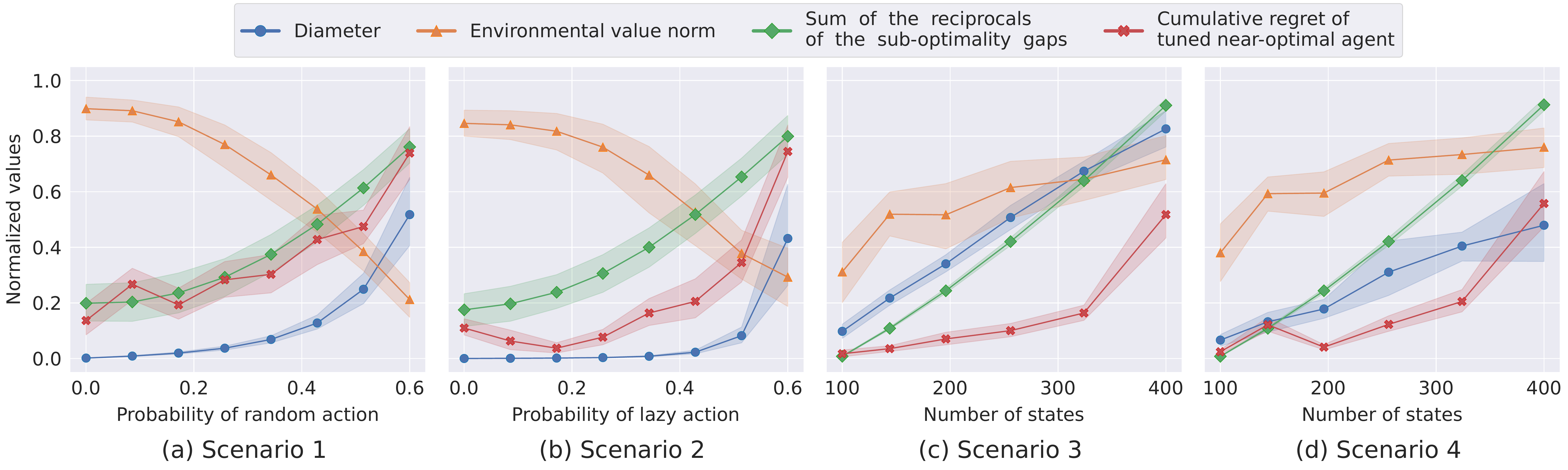}
    \caption{The \colosseum hardness analysis for the episodic \texttt{MiniGridEmpty} family.}
    \label{fig:hardness_analysis_MiniGridEmpty}
\end{figure}

\textbf{Analysis.} Figure \ref{fig:hardness_analysis_MiniGridEmpty} presents the empirical results for the episodic \texttt{MiniGridEmpty} family in the four scenarios with 95\% bootstrapped confidence intervals over twelve random seeds.

The experiments confirm our claim that the diameter captures visitation rather than estimation complexity.
This measure of hardness grows superlinearly with both \prand and \plazy (Figures \ref{fig:hardness_analysis_MiniGridEmpty}a and \ref{fig:hardness_analysis_MiniGridEmpty}b) since deliberate movement between states requires an exponentially increasing number of time steps.
Although the diameter highlights the sharply increasing visitation complexity, its trend overestimates the increase in cumulative regret of the tuned near-optimal agent, which is explained by the unaccounted decrease in estimation complexity. The diameter also increases almost linearly with the number of states (Figures \ref{fig:hardness_analysis_MiniGridEmpty}c and \ref{fig:hardness_analysis_MiniGridEmpty}d). For the small \prand (scenario 4), the relation is still approximately linear.
This linear trend underestimates the evident non-linear growth in hardness in the regret of the tuned near-optimal agent but is in line with the mild increase in visitation complexity.

The empirical evidence indicates that the environmental value norm can only capture estimation complexity.
It decreases as \plazy and \prand increase (Figures \ref{fig:hardness_analysis_MiniGridEmpty}a and \ref{fig:hardness_analysis_MiniGridEmpty}b) because the optimal value of neighboring states becomes closer, which decreases the per-step variability of the optimal value function. When the number of states increases but the transition and reward structures remain the same (Figures \ref{fig:hardness_analysis_MiniGridEmpty}c and \ref{fig:hardness_analysis_MiniGridEmpty}d), the small increase in this variability only generates a sublinear growth.

We empirically observe that the \subgaps is not particularly apt at capturing estimation complexity, due to its exclusive focus on optimal policy identification, and it also underestimates the increase in hardness induced by an increase in visitation complexity.
This measure increases weakly superlinearly in scenarios 1 and 2 (Figures \ref{fig:hardness_analysis_MiniGridEmpty}a and \ref{fig:hardness_analysis_MiniGridEmpty}b). The probability of executing the action selected by the agent decreases when \plazy and \prand increase, so the difference between the state and the state-action optimal value functions decreases sharply.
The measure increases almost linearly with the number of states (Figures \ref{fig:hardness_analysis_MiniGridEmpty}c and \ref{fig:hardness_analysis_MiniGridEmpty}d). This is explained by the fact that the average value of the additional terms in the summation is often similar to the average value of the existing terms when MDPs have the same structure of reward and transition kernels.

\subsection{\colosseum benchmarking} \label{sec:benchmark}

In this section, we benchmark five tabular agents with theoretical guarantees and four non-tabular agents.
Besides being valuable on their own, these results help to empirically validate our benchmark.

\textbf{Agents.} The tabular agents are posterior sampling for \rl (PSRL) for the episodic and continuous settings \citep{osband2013more, agrawal2017posterior}, Q-learning with UCB exploration for the episodic setting \citep{jin2018q}, Q-learning with optimism for the continuous setting \citep{wei2020model}, and UCRL2 for the continuous setting \citep{jaksch2010near}.
The non-tabular agents (from \texttt{bsuite}) are ActorCritic, ActorCriticRNN, BootDQN, and DQN.

\textbf{Experimental procedure.}
We set the total number of time steps to $500\ 000$ with a maximum training time of $10$ minutes for the tabular setting and $40$ minutes for the non-tabular setting.
If an agent does not reach the maximum number of time steps before this time limit, learning is interrupted, and the agent continues interacting using its last best policy.
This guarantees a fair comparison between agents with different computational costs.
The performance indicators are computed every $100$ time steps.
Each interaction between an agent and an MDP is repeated for $20$ seeds.
The agents' hyperparameters have been chosen by random search to minimize the average regret across MDPs with randomly sampled parameters (see Appendix \ref{app:benchmark}).
We use a deterministic emission map that assigns a uniquely identifying vector to each state (for example, a gridworld coordinate) to derive the non-tabular benchmark MDPs.
In Table~\ref{tab:benchmark_results_main}, we report the per-step normalized cumulative regrets divided by the total number of time steps (defined in Appendix~\ref{app:reg_norm}), which allows comparisons across different MDPs.
We summarize the main findings here and refer to Appendix~\ref{app:benchmark} for further details.

\begin{table}[htb]
\captionsetup{position=top}
\centering%
\caption{Normalized cumulative regrets of selected agents on the \colosseum benchmark. (a) Episodic ergodic. (b)
Episodic communicating. (c) Continuous ergodic. (d) Continuous communicating.
}%
\subfloat[%E. ergodic.%
\label{tab:episodic_ergodic_bench}]{%
\resizebox{!}{2.72cm}{%
\begin{tabular}{lrr}
\toprule
{}  &              Q-learning&                    PSRL \\%
MDP &                       &                       \\%
\midrule
DeepSea           &           $.64\pm.00$&  $\mathbf{.01}\pm.00$ \\
                 &           $.52\pm.01$&  $\mathbf{.00}\pm.00$  \\
\arrayrulecolor{black!15}\midrule%
FrozenLake        &           $.90\pm.01$&  $\mathbf{.01}\pm.00$ \\
\arrayrulecolor{black!15}\midrule%
MG-Empty    &           $1.00\pm.00$&  $\mathbf{.86}\pm.16$  \\
                 &           $1.00\pm.00$ &  $\mathbf{.94}\pm.07$ \\
                  &           $1.00\pm.00$ &  $\mathbf{.91}\pm.09$\\
                  &           $1.00\pm.00$ &  $\mathbf{.35}\pm.10$\\
                 &           $1.00\pm.00$&  $\mathbf{.44}\pm.12$  \\
                 &           $.92\pm.04$&  $\mathbf{.14}\pm.08$  \\
                  &           $.91\pm.03$ &  $\mathbf{.04}\pm.03$\\
\arrayrulecolor{black!15}\midrule%
MG-Rooms     &           $.90\pm.04$&  $\mathbf{.05}\pm.04$ \\
                  &           $1.00\pm.00$&  $\mathbf{.54}\pm.36$ \\
                 &           $.99\pm.01$&  $\mathbf{.24}\pm.29$  \\
\arrayrulecolor{black!15}\midrule%
RiverSwim         &           $.07\pm.02$&  $\mathbf{.00}\pm.00$ \\
                  &           $.91\pm.01$&  $\mathbf{.00}\pm.00$ \\
\arrayrulecolor{black!15}\midrule%
SimpleGrid      &           $.78\pm.03$ &  $\mathbf{.05}\pm.01$  \\
                  &  $\mathbf{.79}\pm.03$&  $\mathbf{.79}\pm.03$ \\
                  &  $\mathbf{.50}\pm.03$&  $\mathbf{.50}\pm.03$ \\
\arrayrulecolor{black!15}\midrule%
Taxi              &           $.84\pm.01$&  $\mathbf{.08}\pm.01$ \\
                  &           $.56\pm.02$&  $\mathbf{.05}\pm.00$ \\
\arrayrulecolor{black!30}\midrule%
\textit{Average}  &           $.81\pm.24$&  $\mathbf{.30}\pm.33$ \\
\arrayrulecolor{black!15}\midrule%
\end{tabular}

}}%
\hfill%
\subfloat[%E. communicating.%
\label{tab:episodic_communicating_bench}]{%
\resizebox{!}{2.72cm}{%
\begin{tabular}{lrr}%
\toprule
    &              Q-learning &   PSRL     \\%
MDP &                       &                       \\%
\midrule%
DeepSea           &           $.01\pm.01$&  $\mathbf{.00}\pm.00$ \\
                  &           $.83\pm.02$&  $\mathbf{.54}\pm.01$ \\
\arrayrulecolor{black!15}\midrule%
FrozenLake        &           $.78\pm.04$&  $\mathbf{.03}\pm.11$ \\
\arrayrulecolor{black!15}\midrule%
MG-Empty     &           $.59\pm.07$&  $\mathbf{.09}\pm.05$ \\
                  &           $.99\pm.00$&  $\mathbf{.24}\pm.15$ \\
                  &           $.99\pm.01$&  $\mathbf{.23}\pm.12$ \\
                  &           $1.00\pm.00$&  $\mathbf{.91}\pm.09$ \\
                  &           $1.00\pm.00$&  $\mathbf{.93}\pm.09$ \\
\arrayrulecolor{black!15}\midrule%
MG-Rooms     &           $.99\pm.01$&  $\mathbf{.21}\pm.29$ \\
                  &           $1.00\pm.00$&  $\mathbf{.44}\pm.39$ \\
                 &           $1.00\pm.00$ &  $\mathbf{.43}\pm.39$ \\
                  &           $.94\pm.05$&  $\mathbf{.04}\pm.04$ \\
\arrayrulecolor{black!15}\midrule%
RiverSwim         &           $.87\pm.00$&  $\mathbf{.00}\pm.00$ \\
                 &           $.96\pm.01$ &  $\mathbf{.80}\pm.00$ \\
\arrayrulecolor{black!15}\midrule%
SimpleGrid        &           $.78\pm.10$&  $\mathbf{.20}\pm.15$ \\
                  &           $.80\pm.00$ &  $\mathbf{.55}\pm.15$\\
                  &           $.50\pm.00$&  $\mathbf{.11}\pm.01$ \\
                  &  $\mathbf{.79}\pm.04$&  $\mathbf{.79}\pm.04$ \\
\arrayrulecolor{black!15}\midrule%
Taxi             &           $.94\pm.00$&  $\mathbf{.09}\pm.01$  \\
                  &           $.91\pm.01$ &  $\mathbf{.36}\pm.06$\\
\arrayrulecolor{black!30}\midrule%
\textit{Average}  &           $.83\pm.23$&  $\mathbf{.35}\pm.30$ \\
\arrayrulecolor{black!15}\midrule%
\end{tabular}
}}%
\hfill%
\subfloat[%Cont. ergodic.%
\label{tab:continuous_ergodic_bench}]{%
\resizebox{!}{2.72cm}{%
\begin{tabular}{lrrr}%
\toprule%
              &      Q-learning       & PSRL  &         UCRL2 \\%
MDP &                       &                        &                       \\%
\midrule%
DeepSea          &  $.94\pm.00$&  $\mathbf{.06}\pm.01$  &           $.23\pm.05$ \\
\arrayrulecolor{black!15}\midrule%
FrozenLake       &  $.83\pm.03$&  $\mathbf{.01}\pm.03$  &  $\mathbf{.01}\pm.02$ \\
\arrayrulecolor{black!15}\midrule%
MG-Empty     &  $.98\pm.02$&           $.99\pm.01$ &  $\mathbf{.05}\pm.06$ \\
                  &  $.98\pm.02$&           $.98\pm.04$ &  $\mathbf{.03}\pm.05$ \\
                  &  $.97\pm.00$&           $.95\pm.03$ &  $\mathbf{.04}\pm.01$ \\
                  &  $.98\pm.01$&           $.99\pm.01$ &  $\mathbf{.54}\pm.26$ \\
                  &  $.96\pm.01$&           $.83\pm.31$ &  $\mathbf{.01}\pm.00$ \\
                  &  $.98\pm.02$&           $.99\pm.02$ &  $\mathbf{.45}\pm.35$ \\
                  &  $.98\pm.03$&           $.99\pm.01$ &  $\mathbf{.27}\pm.33$ \\
                  &  $.98\pm.01$&           $.99\pm.01$ &  $\mathbf{.93}\pm.09$ \\
\arrayrulecolor{black!15}\midrule%
MG-Rooms     &  $.98\pm.03$&           $.99\pm.02$ &  $\mathbf{.18}\pm.29$ \\
                  &  $.98\pm.02$&           $1.00\pm.00$ &  $\mathbf{.62}\pm.36$ \\
\arrayrulecolor{black!15}\midrule%
RiverSwim         &  $.73\pm.19$&  $\mathbf{.00}\pm.00$ &  $\mathbf{.00}\pm.00$ \\
                  &  $.71\pm.22$&  $\mathbf{.00}\pm.00$ &           $.01\pm.00$ \\
                 &  $.90\pm.06$ &           $.02\pm.04$ &  $\mathbf{.01}\pm.01$ \\
                  &  $.50\pm.25$&  $\mathbf{.01}\pm.00$ &  $\mathbf{.01}\pm.01$ \\
\arrayrulecolor{black!15}\midrule%
SimpleGrid        &  $.78\pm.00$&           $.70\pm.19$ &  $\mathbf{.01}\pm.01$ \\
                  &  $.46\pm.08$&           $.01\pm.02$ &  $\mathbf{.00}\pm.00$ \\
                  &  $.49\pm.00$&           $.43\pm.16$ &  $\mathbf{.00}\pm.00$ \\
\arrayrulecolor{black!15}\midrule%
Taxi              &  $.87\pm.01$&           $.89\pm.08$ &  $\mathbf{.09}\pm.01$ \\
\arrayrulecolor{black!30}\midrule%
\textit{Average}  &  $.85\pm.18$&           $.59\pm.44$ &  $\mathbf{.17}\pm.26$ \\
\arrayrulecolor{black!15}\midrule%
\end{tabular}
}}%
\hfill%
\subfloat[%Cont. communicating.%
\label{tab:continuous_communicating_bench}]{%
\resizebox{!}{2.72cm}{%
\begin{tabular}{lrrr}%
\toprule%
                   &  Q-learning &       PSRL  &         UCRL2 \\%
MDP &                       &                        &                       \\%
\midrule%
DeepSea           &  $\mathbf{.78}\pm.00$&  $\mathbf{.78}\pm.05$ &           $.90\pm.01$ \\
                  &  $\mathbf{.99}\pm.00$&  $\mathbf{.99}\pm.00$ &  $\mathbf{.99}\pm.00$ \\
                  &  $\mathbf{.79}\pm.00$&  $\mathbf{.79}\pm.04$ &           $.92\pm.01$ \\
\arrayrulecolor{black!15}\midrule%
FrozenLake        &           $.77\pm.04$&  $\mathbf{.01}\pm.04$ &  $\mathbf{.01}\pm.01$ \\
                  &           $.84\pm.04$&  $\mathbf{.01}\pm.02$ &           $.04\pm.06$ \\
\arrayrulecolor{black!15}\midrule%
MG-Empty     &           $.51\pm.23$&           $.95\pm.22$ &  $\mathbf{.02}\pm.00$ \\
                  &  $\mathbf{.01}\pm.00$&           $1.00\pm.00$ &           $.02\pm.00$ \\
                  &  $\mathbf{.00}\pm.00$&           $.60\pm.50$ &           $.01\pm.00$ \\
                  &           $.35\pm.17$&           $1.00\pm.00$ &  $\mathbf{.01}\pm.00$ \\
                  &           $.75\pm.21$&           $1.00\pm.00$ &  $\mathbf{.08}\pm.20$ \\
\arrayrulecolor{black!15}\midrule%
MG-Rooms     &  $\mathbf{.01}\pm.01$&           $1.00\pm.00$ &           $.78\pm.40$ \\
                  &  $\mathbf{.01}\pm.01$&           $1.00\pm.00$ &           $.02\pm.01$ \\
                  &  $\mathbf{.02}\pm.02$&           $1.00\pm.00$ &           $.66\pm.47$ \\
\arrayrulecolor{black!15}\midrule%
RiverSwim         &           $.16\pm.03$&  $\mathbf{.00}\pm.01$ &  $\mathbf{.00}\pm.00$ \\
                  &           $.34\pm.14$&  $\mathbf{.01}\pm.00$ &           $.02\pm.01$ \\
\arrayrulecolor{black!15}\midrule%
SimpleGrid        &           $.11\pm.01$&           $.93\pm.00$ &  $\mathbf{.01}\pm.00$ \\
                  &  $\mathbf{.01}\pm.00$&           $.45\pm.15$ &  $\mathbf{.01}\pm.00$ \\
                  &  $\mathbf{.15}\pm.01$&           $.93\pm.00$ &           $.70\pm.40$ \\
                  &  $\mathbf{.01}\pm.00$&           $.50\pm.00$ &           $.33\pm.24$ \\
\arrayrulecolor{black!15}\midrule%
Taxi              &           $.95\pm.00$&           $.94\pm.04$ &  $\mathbf{.12}\pm.01$ \\
\arrayrulecolor{black!30}\midrule%
\textit{Average}  &           $.38\pm.37$&           $.69\pm.38$ &  $\mathbf{.28}\pm.37$ \\
\arrayrulecolor{black!15}\midrule%
\end{tabular}
}}%
\label{tab:benchmark_results_main}%
\end{table}%

\textbf{Analysis.} 
Table~\ref{tab:benchmark_results_main} often shows high variability in the performance of the same agent across MDPs of the same family.
Therefore, maximising the diversity across diameters and value norms effectively produces diverse challenges even for MDPs with similar transition and reward structures.
For example, in the continuous communicating case (Table \ref{tab:continuous_communicating_bench}), Q-learning performs well only in some MDPs of the \texttt{MiniGridEmpty} family. This also happens for UCRL2 for the \texttt{SimpleGrid} family.

The average normalized cumulative regret is lower in ergodic environments compared to communicating environments. This indicates that the ergodic setting is generally slightly easier than the communicating settings. Notably, in the continuous setting, the ergodic setting is more challenging than the communicating setting for Q-learning (Tables \ref{tab:continuous_ergodic_bench} and \ref{tab:continuous_communicating_bench}).
Designing a naturally ergodic MDP is not straightforward. In fact, the majority of MDPs in the literature are communicating. In \colosseum, ergodicity is induced by setting $\prand > 0$ in otherwise communicating MDPs. Model-free agents struggle with the resulting increase in variability of the state-action value function. 

In the episodic settings (Tables~\ref{tab:episodic_ergodic_bench} and \ref{tab:episodic_communicating_bench}), PSRL obtains excellent performances with low variability. Q-learning instead performs well in a few MDPs.
This often happens since, when the action selected by the agent is randomly substituted (due to $\prand > 0$) with one with a large sub-optimality gap, the resulting $Q$-value update introduces a critical error that requires many samples to be corrected.

In the continuous settings (Tables~\ref{tab:continuous_ergodic_bench} and \ref{tab:continuous_communicating_bench}), UCRL2 performs best in the ergodic cases when Q-learning suffers from the issue caused by $\prand > 0$ but is only slightly better than Q-learning in the communicating ones.
PSRL instead struggles with most MDPs. The reason for its weak performance in this setting is the computationally expensive optimistic sampling procedure required for its worst-case theoretical guarantees.
It often breaks the time limit before reaching the first quarter of available time steps, meaning that it lacks sufficient samples to estimate the optimal policy.

\begin{figure}[htb]
    \centering
    \includegraphics[width=\textwidth]{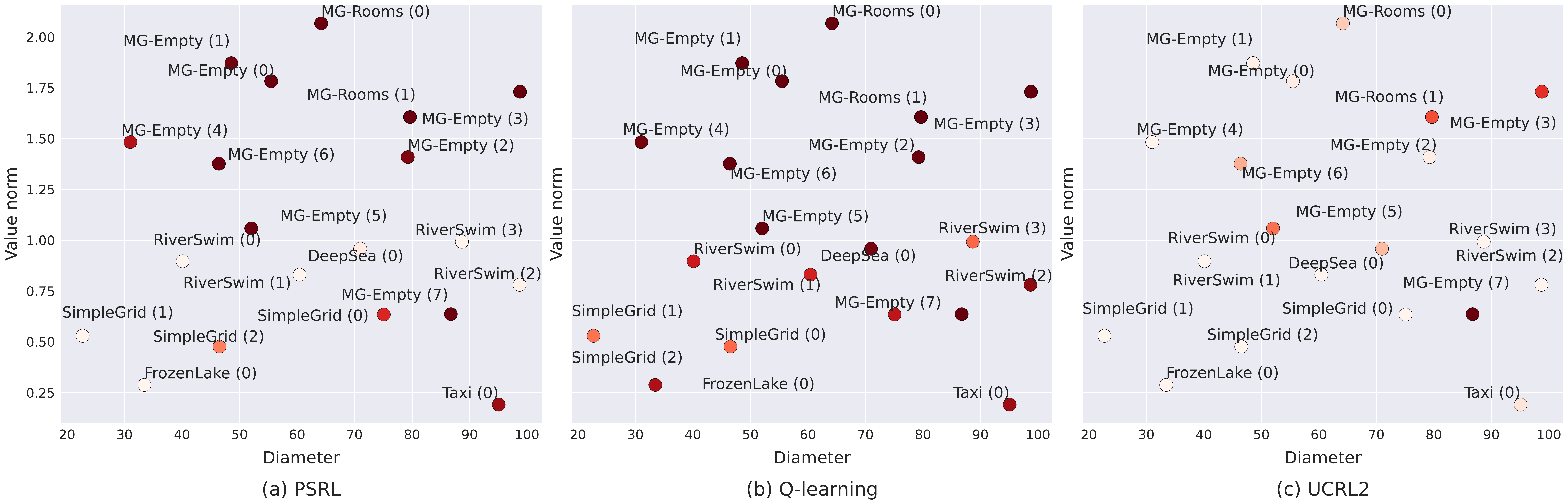}
\caption{Average cumulative regret obtained by the tabular agents with guarantees in the continuous ergodic setting placed according to the diameter and the value norm associated to the MDPs.} 
    \label{fig:hardness_and_crs_main}
\end{figure}

Figure~\ref{fig:hardness_and_crs_main} places the regret of the agents in the continuous ergodic setting (Table \ref{tab:continuous_ergodic_bench}) on a position corresponding to the diameter and value norm of the benchmark environments.
PSRL and Q-learning (Figures \ref{fig:hardness_and_crs_main}a and \ref{fig:hardness_and_crs_main}b), appear to be impacted more by the value norm than the diameter.
This is in line with the lack of sufficient samples for PSRL and the aforementioned issue related to high q estimates variability for Q-learning, which is exacerbated when the estimation complexity is higher.
In the case of UCRL2 (Figure~\ref{fig:hardness_and_crs_main}c), which provides more reliable evidence since it performs well across the MDPs, higher regret effectively corresponds to higher diameter and value norm.

\begin{figure}[htb]
    \centering
    \includegraphics[width=\textwidth]{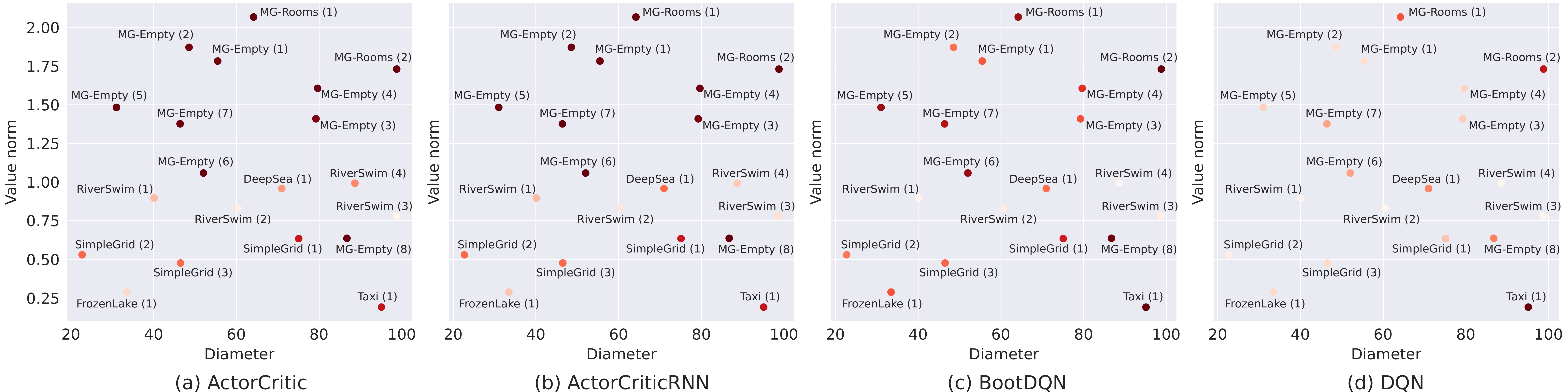}
\caption{Average cumulative regret obtained by the non-tabular baseline agents in the continuous ergodic setting placed according to the diameter and the value norm associated to the MDPs.} 
    \label{fig:nt_hardness_and_crs_main}
\end{figure}

\textbf{Non-tabular benchmarking.}
The performance of the agents is in line with the results reported by \citet{osband2020bsuite}, with the exception of BootDQN.
Being the most computationally intensive, this agent often breaks the time limit, which consequently worsens its overall performance.
Figure~\ref{fig:nt_hardness_and_crs_main} places the regret of the agents in the continuous ergodic setting on a position corresponding to the diameter and value norm of the benchmark environments.
Interestingly, and similarly to the tabular case (Figure~\ref{fig:hardness_and_crs_main}), while the best performing agent (DQN) is evidently impacted more by the diameter, the opposite holds for the other agents. Regardless of the visitation complexity, this suggests that an agent that fails to handle the estimation complexity of an environment is bound to perform badly both in the tabular and non-tabular settings.

\section{Conclusion} \label{sec:conclusion}

We established the usefulness of the theory of hardness in empirical reinforcement learning.
Prior to our work, hardness measures were limited to providing theoretical guarantees for agents.
In order to promote a wider understanding of these measures, we presented a systematic survey that newly identified two major approaches for characterizing hardness: Markov chain-based and value-based.
These approaches aim to capture complementary aspects of hardness: visitation complexity and estimation complexity.
Our survey also exposed a relative lack of measures that capture both aspects, which motivates our definition of complete measures of hardness.
Their development is important theoretically, elucidating what makes a problem hard for a specific performance criterion, and empirically, allowing the creation of principled benchmarks for a specific performance criterion.

We presented the first empirical study of (efficiently computable) hardness measures. % using several families of environments found in previous work.
This study revealed which aspects of hardness current measures capture
and clarified their relationship with the behavior of near-optimal agents. Based on these results, we proposed a benchmark for the most widely studied tabular reinforcement learning settings that contains environments that maximize diversity with respect to two highly distinct measures. Such a principled benchmark is valuable to gauge progress in the field. The new benchmark allowed conducting the most exhaustive empirical comparison between theoretically principled tabular reinforcement learning agents to date, which revealed
undocumented weaknesses of these agents and further validated our choices of environments.

As a first step towards principled non-tabular benchmarking, we argued that many commonly used environments can be encoded as BlockMDPs, which are non-tabular versions of tabular MDPs for which a partial characterization of hardness is already possible.
We observed a clear empirical relation between two tabular hardness measures and the performance of four non-tabular agents.
BlockMDPs represent a promising starting point for the future development of non-tabular hardness measures while already being useful to provide relevant insights into the performance of non-tabular agents.

Our work has led to the development of \colosseum, a pioneering tool for empirical but theoretically principled study of tabular reinforcement learning with experimental non-tabular benchmarking capabilities.
Besides implementing the aforementioned tabular benchmark, \colosseum provides valuable analysis tools: regret and hardness computations, communication class identification, logging, and visualizations.
\colosseum can also be easily extended and integrated with new agents and environments, for which we will actively seek contributions from the community.
We strongly believe that \colosseum has the potential to become a fundamental tool in \rl.

\begin{ack}
This research was financially supported by the Intelligent Games and Games Intelligence CDT (IGGI; EP/S022325/1) and used Queen Mary University of London Apocrita HPC facility.
The authors would like to thank Sjoerd van Steenkiste and Tabish Rashid for their valuable feedback.

\end{ack}

\bibliography{bibliography}

\begin{thebibliography}{35}
\providecommand{\natexlab}[1]{#1}
\providecommand{\url}[1]{\texttt{#1}}
\expandafter\ifx\csname urlstyle\endcsname\relax
  \providecommand{\doi}[1]{doi: #1}\else
  \providecommand{\doi}{doi: \begingroup \urlstyle{rm}\Url}\fi

\bibitem[Silver et~al.(2018)Silver, Hubert, Schrittwieser, Antonoglou, Lai,
  Guez, Lanctot, Sifre, Kumaran, Graepel, et~al.]{silver2018general}
David Silver, Thomas Hubert, Julian Schrittwieser, Ioannis Antonoglou, Matthew
  Lai, Arthur Guez, Marc Lanctot, Laurent Sifre, Dharshan Kumaran, Thore
  Graepel, et~al.
\newblock A general reinforcement learning algorithm that masters chess, shogi,
  and go through self-play.
\newblock \emph{Science}, 2018.

\bibitem[Vinyals et~al.(2019)Vinyals, Babuschkin, Czarnecki, Mathieu, Dudzik,
  Chung, Choi, Powell, Ewalds, Georgiev, et~al.]{vinyals2019grandmaster}
Oriol Vinyals, Igor Babuschkin, Wojciech~M Czarnecki, Micha{\"e}l Mathieu,
  Andrew Dudzik, Junyoung Chung, David~H Choi, Richard Powell, Timo Ewalds,
  Petko Georgiev, et~al.
\newblock Grandmaster level in {StarCraft} {II} using multi-agent reinforcement
  learning.
\newblock \emph{Nature}, 2019.

\bibitem[Berner et~al.(2019)Berner, Brockman, Chan, Cheung, D{\k{e}}biak,
  Dennison, Farhi, Fischer, Hashme, Hesse, et~al.]{berner2019dota}
Christopher Berner, Greg Brockman, Brooke Chan, Vicki Cheung, Przemys{\l}aw
  D{\k{e}}biak, Christy Dennison, David Farhi, Quirin Fischer, Shariq Hashme,
  Chris Hesse, et~al.
\newblock Dota 2 with large scale deep reinforcement learning.
\newblock \emph{arXiv preprint arXiv:1912.06680}, 2019.

\bibitem[Osband et~al.(2020)Osband, Doron, Hessel, Aslanides, Sezener, Saraiva,
  McKinney, Lattimore, {Sz}epesv{\'a}ri, Singh, Van~Roy, Sutton, Silver, and
  van Hasselt]{osband2020bsuite}
Ian Osband, Yotam Doron, Matteo Hessel, John Aslanides, Eren Sezener, Andre
  Saraiva, Katrina McKinney, Tor Lattimore, Csaba {Sz}epesv{\'a}ri, Satinder
  Singh, Benjamin Van~Roy, Richard Sutton, David Silver, and Hado van Hasselt.
\newblock Behaviour suite for reinforcement learning.
\newblock In \emph{International Conference on Learning Representations}, 2020.

\bibitem[Rajan et~al.(2019)Rajan, Diaz, Guttikonda, Ferreira, Biedenkapp, von
  Hartz, and Hutter]{rajan2021mdp}
Raghu Rajan, Jessica Lizeth~Borja Diaz, Suresh Guttikonda, Fabio Ferreira,
  André Biedenkapp, Jan~Ole von Hartz, and Frank Hutter.
\newblock {MDP} playground: A design and debug testbed for reinforcement
  learning.
\newblock \emph{arXiv preprint arXiv:1909.07750}, 2019.

\bibitem[Puterman(2014)]{puterman2014markov}
Martin~L Puterman.
\newblock \emph{Markov decision processes: discrete stochastic dynamic
  programming}.
\newblock John Wiley \& Sons, 2014.

\bibitem[Levin and Peres(2017)]{levin2017markov}
David~A Levin and Yuval Peres.
\newblock \emph{Markov chains and mixing times}.
\newblock American Mathematical Soc., 2017.

\bibitem[Kallenberg(2002)]{kallenberg2002classification}
LCM Kallenberg.
\newblock Classification problems in {MDP}s.
\newblock In \emph{Markov processes and controlled Markov chains}. Springer,
  2002.

\bibitem[Auer and Ortner(2006)]{auer2006logarithmic}
Peter Auer and Ronald Ortner.
\newblock Logarithmic online regret bounds for undiscounted reinforcement
  learning.
\newblock \emph{Advances in Neural Information Processing Systems}, 2006.

\bibitem[Kakade(2003)]{kakade2003sample}
Sham~Machandranath Kakade.
\newblock \emph{On the sample complexity of reinforcement learning}.
\newblock University of London, University College London (United Kingdom),
  2003.

\bibitem[Bellman(1957)]{bellman1957markovian}
Richard Bellman.
\newblock A {M}arkovian decision process.
\newblock \emph{Journal of mathematics and mechanics}, 1957.

\bibitem[Koller and Friedman(2009)]{koller2009probabilistic}
D.~Koller and N.~Friedman.
\newblock \emph{Probabilistic graphical models: principles and techniques}.
\newblock Adaptive Computation and Machine Learning series. MIT Press, 2009.

\bibitem[Kearns and Singh(2002)]{kearns2002near}
Michael Kearns and Satinder Singh.
\newblock Near-optimal reinforcement learning in polynomial time.
\newblock \emph{Machine learning}, 2002.

\bibitem[Jaksch et~al.(2010)Jaksch, Ortner, and Auer]{jaksch2010near}
Thomas Jaksch, Ronald Ortner, and Peter Auer.
\newblock Near-optimal regret bounds for reinforcement learning.
\newblock \emph{Journal of Machine Learning Research}, 2010.

\bibitem[Bartlett and Tewari(2009)]{bartlett2012regal}
Peter Bartlett and Ambuj Tewari.
\newblock {REGAL}: a regularization based algorithm for reinforcement learning
  in weakly communicating {MDP}s.
\newblock In \emph{Uncertainty in Artificial Intelligence: Proceedings of the
  25th Conference}. AUAI Press, 2009.

\bibitem[Wei et~al.(2020)Wei, Jahromi, Luo, Sharma, and Jain]{wei2020model}
Chen-Yu Wei, Mehdi~Jafarnia Jahromi, Haipeng Luo, Hiteshi Sharma, and Rahul
  Jain.
\newblock Model-free reinforcement learning in infinite-horizon average-reward
  {Markov} decision processes.
\newblock In \emph{International Conference on Machine Learning}. PMLR, 2020.

\bibitem[Agarwal et~al.(2021)Agarwal, Kakade, Lee, and
  Mahajan]{agarwal2021theory}
Alekh Agarwal, Sham~M Kakade, Jason~D Lee, and Gaurav Mahajan.
\newblock On the theory of policy gradient methods: optimality, approximation,
  and distribution shift.
\newblock \emph{Journal of Machine Learning Research}, 2021.

\bibitem[Farahmand(2011)]{farahmand2011action}
Amir~Massoud Farahmand.
\newblock Action-gap phenomenon in reinforcement learning.
\newblock \emph{Advances in Neural Information Processing Systems}, 2011.

\bibitem[Maillard et~al.(2014)Maillard, Mann, and Mannor]{maillard2014hard}
Odalric-Ambrym Maillard, Timothy~A Mann, and Shie Mannor.
\newblock How hard is my {MDP}? {T}he distribution-norm to the rescue".
\newblock \emph{Advances in Neural Information Processing Systems}, 2014.

\bibitem[Zanette and Brunskill(2019)]{zanette2019tighter}
Andrea Zanette and Emma Brunskill.
\newblock Tighter problem-dependent regret bounds in reinforcement learning
  without domain knowledge using value function bounds.
\newblock In \emph{International Conference on Machine Learning}. PMLR, 2019.

\bibitem[Simchowitz and Jamieson(2019)]{simchowitz2019non}
Max Simchowitz and Kevin~G Jamieson.
\newblock Non-asymptotic gap-dependent regret bounds for tabular {MDP}s.
\newblock \emph{Advances in Neural Information Processing Systems}, 2019.

\bibitem[Dann and Brunskill(2015)]{dann2015sample}
Christoph Dann and Emma Brunskill.
\newblock Sample complexity of episodic fixed-horizon reinforcement learning.
\newblock \emph{Advances in Neural Information Processing Systems}, 2015.

\bibitem[Wagenmaker et~al.(2022)Wagenmaker, Simchowitz, and
  Jamieson]{wagenmaker2021beyond}
Andrew~J Wagenmaker, Max Simchowitz, and Kevin Jamieson.
\newblock Beyond no regret: Instance-dependent pac reinforcement learning.
\newblock In \emph{Conference on Learning Theory}, pages 358--418. PMLR, 2022.

\bibitem[Strehl and Littman(2008)]{strehl2008analysis}
Alexander~L Strehl and Michael~L Littman.
\newblock An analysis of model-based interval estimation for {Markov} decision
  processes.
\newblock \emph{Journal of Computer and System Sciences}, 2008.

\bibitem[Dietterich(2000)]{dietterich2000hierarchical}
Thomas~G Dietterich.
\newblock Hierarchical reinforcement learning with the {MAXQ} value function
  decomposition.
\newblock \emph{Journal of Artificial Intelligence Research}, 2000.

\bibitem[Chevalier-Boisvert et~al.(2018)Chevalier-Boisvert, Willems, and
  Pal]{gym_minigrid}
Maxime Chevalier-Boisvert, Lucas Willems, and Suman Pal.
\newblock Minimalistic {G}ridworld environment for {OpenAI} gym.
\newblock \url{https://github.com/maximecb/gym-minigrid}, 2018.

\bibitem[Osband et~al.(2019)Osband, Van~Roy, Russo, Wen,
  et~al.]{osband2019deep}
Ian Osband, Benjamin Van~Roy, Daniel~J Russo, Zheng Wen, et~al.
\newblock Deep exploration via randomized value functions.
\newblock \emph{Journal of Machine Learning Research}, 2019.

\bibitem[Du et~al.(2019)Du, Krishnamurthy, Jiang, Agarwal, Dudik, and
  Langford]{du2019provably}
Simon Du, Akshay Krishnamurthy, Nan Jiang, Alekh Agarwal, Miroslav Dudik, and
  John Langford.
\newblock Provably efficient {RL} with rich observations via latent state
  decoding.
\newblock In \emph{International Conference on Machine Learning}, pages
  1665--1674. PMLR, 2019.

\bibitem[Misra et~al.(2020)Misra, Henaff, Krishnamurthy, and
  Langford]{misra2020kinematic}
Dipendra Misra, Mikael Henaff, Akshay Krishnamurthy, and John Langford.
\newblock Kinematic state abstraction and provably efficient rich-observation
  reinforcement learning.
\newblock In \emph{International Conference on Machine Learning}, pages
  6961--6971. PMLR, 2020.

\bibitem[Osband et~al.(2013)Osband, Russo, and Van~Roy]{osband2013more}
Ian Osband, Daniel Russo, and Benjamin Van~Roy.
\newblock ({M}ore) {E}fficient reinforcement learning via posterior sampling.
\newblock \emph{Advances in Neural Information Processing Systems}, 2013.

\bibitem[Agrawal and Jia(2017)]{agrawal2017posterior}
Shipra Agrawal and Randy Jia.
\newblock Posterior sampling for reinforcement learning: worst-case regret
  bounds.
\newblock \emph{Advances in Neural Information Processing Systems}, 2017.

\bibitem[Jin et~al.(2018)Jin, Allen-Zhu, Bubeck, and Jordan]{jin2018q}
Chi Jin, Zeyuan Allen-Zhu, Sebastien Bubeck, and Michael~I Jordan.
\newblock Is {Q}-learning provably efficient?
\newblock \emph{Advances in Neural Information Processing Systems}, 2018.

\bibitem[Wolfer and Kontorovich(2019)]{pmlrv99wolfer19a}
Geoffrey Wolfer and Aryeh Kontorovich.
\newblock Estimating the mixing time of ergodic {Markov} chains.
\newblock In \emph{Proceedings of the Thirty-Second Conference on Learning
  Theory}, Proceedings of Machine Learning Research, 2019.

\bibitem[Lee and Sidford(2013)]{lee2013path}
Yin~Tat Lee and Aaron Sidford.
\newblock Path finding methods for linear programming: Solving linear programs
  in {\~{o}}(sqrt(rank)) iterations and faster algorithms for maximum flow.
\newblock \emph{arXiv preprint arXiv:1312.6677}, 2013.

\bibitem[Sidford et~al.(2018)Sidford, Wang, Wu, Yang, and Ye]{sidford2018near}
Aaron Sidford, Mengdi Wang, Xian Wu, Lin Yang, and Yinyu Ye.
\newblock Near-optimal time and sample complexities for solving markov decision
  processes with a generative model.
\newblock In \emph{Advances in Neural Information Processing Systems}, 2018.

\end{thebibliography}
\bibliographystyle{unsrtnat}

\section*{Checklist}

\begin{enumerate}

\item For all authors...
\begin{enumerate}
  \item Do the main claims made in the abstract and introduction accurately reflect the paper's contributions and scope?
    \answerYes{
In the introduction, we explain where each contribution is described in the text. The same contributions are mentioned in the abstract. In the conclusion, we also present a concrete list of contributions.
    }
  \item Did you describe the limitations of your work?
    \answerYes{ We describe the assumptions behind every theoretical measure of hardness. The results of our empirical investigation are naturally limited to the environments, measures, and agents that we considered.  Nevertheless, we believe that the experimental protocol described in Appendices ~\ref{app:analysis} and ~\ref{app:benchmark} is very robust. The remaining limitations are discussed as potential future works.
}
  \item Did you discuss any potential negative societal impacts of your work?
    \answerNo{We do not envision potential negative societal impacts of our work.}
  \item Have you read the ethics review guidelines and ensured that your paper conforms to them?
    \answerYes{}
\end{enumerate}

\item If you are including theoretical results...
\begin{enumerate}
  \item Did you state the full set of assumptions of all theoretical results?
    \answerYes{
We explicitly mention the full set of assumptions involved in theoretical results and report relevant references when appropriate.}
  \item Did you include complete proofs of all theoretical results?
    \answerNA{}
\end{enumerate}

\item If you ran experiments...
\begin{enumerate}
  \item Did you include the code, data, and instructions needed to reproduce the main experimental results (either in the supplemental material or as a URL)?
    \answerYes{
We include a link to the code in Section \ref{sec:practice}.
We describe the experimental procedures for the empirical evaluation of the measures of hardness in Section~\ref{sec:analysis} and for the \colosseum benchmark evaluation in Section~\ref{sec:benchmark}. We refer to the corresponding appendices (Apps. \ref{app:analysis} and \ref{app:benchmark}) for additional details.
}
  \item Did you specify all the training details (e.g., data splits, hyperparameters, how they were chosen)?
    \answerYes{
The details for the empirical evaluation of the measures of hardness and the \colosseum benchmark evaluation are provided in Appendices \ref{app:analysis} and \ref{app:benchmark}.
}
        \item Did you report error bars (e.g., with respect to the random seed after running experiments multiple times)?
    \answerYes{
We report 95\% bootstraped confidence intervals in the plots and standard deviations in the tables.
}
        \item Did you include the total amount of compute and the type of resources used (e.g., type of GPUs, internal cluster, or cloud provider)?
    \answerYes{
These details are provided in Appendices \ref{app:analysis} and \ref{app:benchmark}.
}
\end{enumerate}

\item If you are using existing assets (e.g., code, data, models) or curating/releasing new assets...
\begin{enumerate}
  \item If your work uses existing assets, did you cite the creators?
    \answerYes{
We describe all the dependencies of \colosseum in Appendix~\ref{app:colosseum}.
}
  \item Did you mention the license of the assets?
    \answerNo{However, all these assets are open-source and freely available.}
  \item Did you include any new assets either in the supplemental material or as a URL?
    \answerYes{
The code of our newly proposed Python package (\colosseum) is included in a link in Section \ref{sec:practice}.
}
  \item Did you discuss whether and how consent was obtained from people whose data you're using/curating?
    \answerNA{}
  \item Did you discuss whether the data you are using/curating contains personally identifiable information or offensive content?
    \answerNA{}
\end{enumerate}

\item If you used crowdsourcing or conducted research with human subjects...
\begin{enumerate}
  \item Did you include the full text of instructions given to participants and screenshots, if applicable?
    \answerNA{}
  \item Did you describe any potential participant risks, with links to Institutional Review Board (IRB) approvals, if applicable?
    \answerNA{}
  \item Did you include the estimated hourly wage paid to participants and the total amount spent on participant compensation?
    \answerNA{}
\end{enumerate}

\end{enumerate}

\clearpage
\appendix

\section{\colosseum} \label{app:colosseum}

Colosseum is a pioneering Python package that creates a bridge between theory and practice in tabular reinforcement learning with an eye on the non-tabular setting.
It allows to empirically, and efficiently, investigate the hardness of MDPs, and it implements the first principled benchmark for tabular \rl algorithms.
In the following sections, we report some additional details on the capabilities of \colosseum.
However, we invite the reader to check the latest online documentation along with the tutorials that cover in detail every aspect of the package.\footnote{Available at \url{https://michelangeloconserva.github.io/Colosseum}.}

\subsection{Expected performance indicators}

Each agent in \colosseum is required to implement a function that returns its current best policy estimate $\hat \pi^*_t$ for any time step $t$.
Using an efficient implementation of the policy evaluation algorithm, \colosseum can compute the corresponding expected regret and expected average reward, which, summed across time steps, amounts to the expected cumulative reward and expected cumulative regret.
Although it is possible to perform this operation at every time step of the agent/MDP interaction, we leave the option to approximate the expected cumulative regret by calculating the expected regret every $n$ time steps and assuming that the policy of the agent in the previous $n-1$ time steps would have yielded a similar expected regret.
For instance, for $n=100$, the expected cumulative regret at time step $T=500$ would be approximated as the sum of the expected regrets calculated at time steps $t=100,200, \ldots, 500$ multiplied by $100$.

\subsection{Non-tabular capabilities} \label{app:bmdp}

\colosseum is primarily aimed at the tabular \rl setting.
However, as our ultimate goal is to develop principled non-tabular benchmarks, we offer a way to test non-tabular \rl algorithms on the \colosseum benchmark.
Although our benchmark defines a challenge that is well characterized for tabular agents, we believe that it can provide valuable insights into the performance of non-tabular algorithms.
In order to do so, we adopt the \textit{BlockMDP} formalism proposed by \citet{du2019provably}.
A BlockMDP is a tuple $\left(\cS, \cA, P, P_0, R, \cO, q\right)$, where $\cO$ and $q : \cS \to \Delta(\cO)$ are respectively the non-tabular observation space that the agent observes and the (possibly stochastic) \textit{emission map} that associates a distribution over the observation space to each state in the MDP.
Note that the agent is not provided with any information on the state space $\cS$.
\colosseum implements six deterministic emission maps with different properties and four kinds of noise to make the emission maps stochastic, which we describe below.
Examples of the emission maps with distinguishable characteristics for each MDP family will be presented in the corresponding sections.

\textbf{Emission maps:}
\begin{itemize}
    \item \textit{One-hot encoding}. This emission map assigns to each state a feature vector that is filled with zeros with the exception of an index that uniquely corresponds to the state.
    \item \textit{Linear optimal value}. This emission map assigns to each state a feature vector $\phi(s)$ that enables linear representation of the optimal value function. In other words, there is a $\theta$ such that $V^*(s) = \theta^T\phi(s)$.
    \item \textit{Linear random value}. This emission map assigns to each state a feature vector $\phi(s)$ that enables linear representation of the value function of the randomly acting policy. In other words, there is a $\theta$ such that $V^\pi(s) = \theta^T\phi(s)$, where $\pi$ is the randomly acting policy.
    \item \textit{State information.} This emission map assigns to each state a feature vector that contains uniquely identifying information about the state (e.g., coordinates for the DeepSea family).
    \item \textit{Image encoding}. This emission map assigns to each state a feature matrix that encodes the visual representation of the MDP as a grayscale image.
    \item \textit{Tensor encoding}. This emission map assigns to each state a tensor composed of the concatenation of matrices that one-hot encode the presence of a symbol in the corresponding indices. For example, for the DeepSea family, the tensor is composed of a matrix that encodes the position of the agent and a matrix that encodes the positions of white spaces.
\end{itemize}

\textbf{Noise:}
\begin{itemize}
    \item \textit{Uncorrelated light-tailed noise}. The output of the emission map is corrupted with element-wise uncorrelated Gaussian noise.
    \item \textit{Correlated light-tailed noise}. The output of the emission map is corrupted with multivariate correlated Gaussian noise with a covariance matrix sampled from a Wishart distribution with a pre-specified scale level when the MDP is created. In other words, the correlation structure of the noise remains unchanged while the agent interacts with the MDP.
    \item \textit{Uncorrelated heavy-tailed noise}. The output of the emission map is corrupted with element-wise uncorrelated Student’s t noise.
    \item \textit{Correlated heavy-tailed noise}. The output of the emission map is corrupted with multivariate correlated Student’s t noise with covariance matrix sampled from a Wishart distribution with a pre-specified scale level when the MDP is created. In other words, the correlation structure of the noise remains unchanged while the agent interacts with the MDP.
\end{itemize}

\subsection{\colosseum MDP families} \label{app:mdps}

\colosseum implements eight families of MDPs.
When selecting which families to include in Colosseum, we aimed to balance between traditional environment families (\texttt{RiverSwim} \citep{strehl2008analysis}, \texttt{Taxi} \citep{dietterich2000hierarchical}, and \texttt{FrozenLake}) and unconventional ones (\texttt{MiniGid} environments \citep{gym_minigrid}). The \texttt{DeepSea} family \citep{osband2019deep} was included since it was proposed as an example of a hard exploration problem. The \texttt{SimpleGrid} family acts as a simplified version of the \texttt{MiniGrid-Empty} environment.

Each MDP family requires a set of parameters to instantiate an MDP.
In addition to individual parameters, all MDP families share the following:
\begin{itemize}
    \item The \texttt{size} $\in \bN$ parameter controls the number of states through geometrical properties of the MDP family.
    For example, in a grid world, it controls the size of the grid.
    This parameter allows increasing the difficulty of an MDP instance without altering the fundamental structure of the MDP family.
    \item The \prand $\in [0, 1)$ parameter controls the probability $r$ that an MDP executes an action at random instead of the one selected by the agent. Concretely, the new transition kernel is given by  $P'(s_{t+1}\mid s_t, a_t) = (1-r)P(s_{t+1}\mid s_t, a_t) + \frac{r}{|\mathcal{A}|}\sum_{a}  P(s_{t+1}\mid s_t, a)$.
    Setting this parameter to a non-zero value can make a communicating MDP ergodic.
    \item The \texttt{lazy} $\in [0, 1)$ parameter controls the probability $l$ of an action not being executed. Concretely, the new transition kernel is given by $P'(s_{t+1}\mid s_t, a_t) = (1-l)P(s_{t+1}\mid s_t, a_t) + l \mathbbm{1}(s_{t+1} = s_t)$.
    This parameter can render a deterministic MDP stochastic without changing the communication class.
    \item \texttt{make\_reward\_stochastic} is a boolean parameter to render the rewards stochastic instead of deterministic (the default).
    We opted for a Beta distribution to guarantee rewards bounded in a specific range. However, it is possible to specify custom reward distributions using \texttt{scipy} random variables.
    \item \texttt{r\_min} and \texttt{r\_max} scale the rewards. The default values are $0$ and $1$, respectively.
\end{itemize}
The hardness analysis presented in Appendix~\ref{app:analysis} shows the relationships between the measures of hardness and some of these parameters for the Colosseum MDP families.
Such relationships can be easily exploited to create MDP instances with specific hardness characterization.
For example, in order to create an MDP instance with low estimation complexity and high visitation complexity, one can force the MDP instance to be deterministic by setting \prand and \plazy to zero and the size to a high value.
If instead one wants to increase the estimation complexity while keeping the visitation complexity fixed, the mean reward of a subset of states can be increased.
The scale of the increase depends on the variability of the next state distributions of the selected states.
The more variable such distributions are the higher the increase in estimation complexity.

\subsubsection{\texttt{RiverSwim}}

\begin{figure}[!ht]
    \centering
    \subfloat[MDP representation.
    \label{fig:riverswim_cont}]{\includegraphics[width=0.4\textwidth]{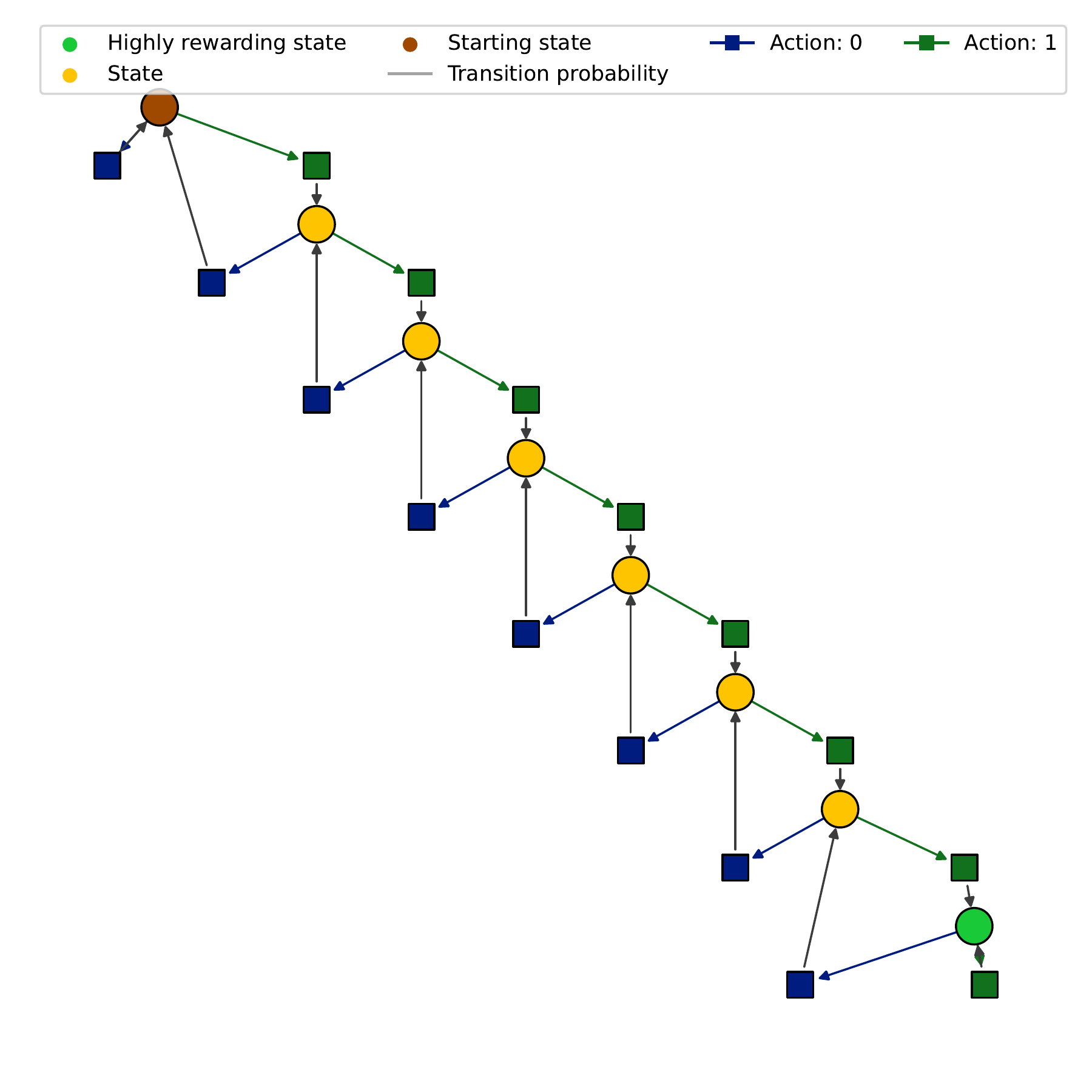}}
    \hfill
    \subfloat[Markov chain representation.
    \label{fig:riverswim_mc}]{\includegraphics[width=0.4\textwidth]{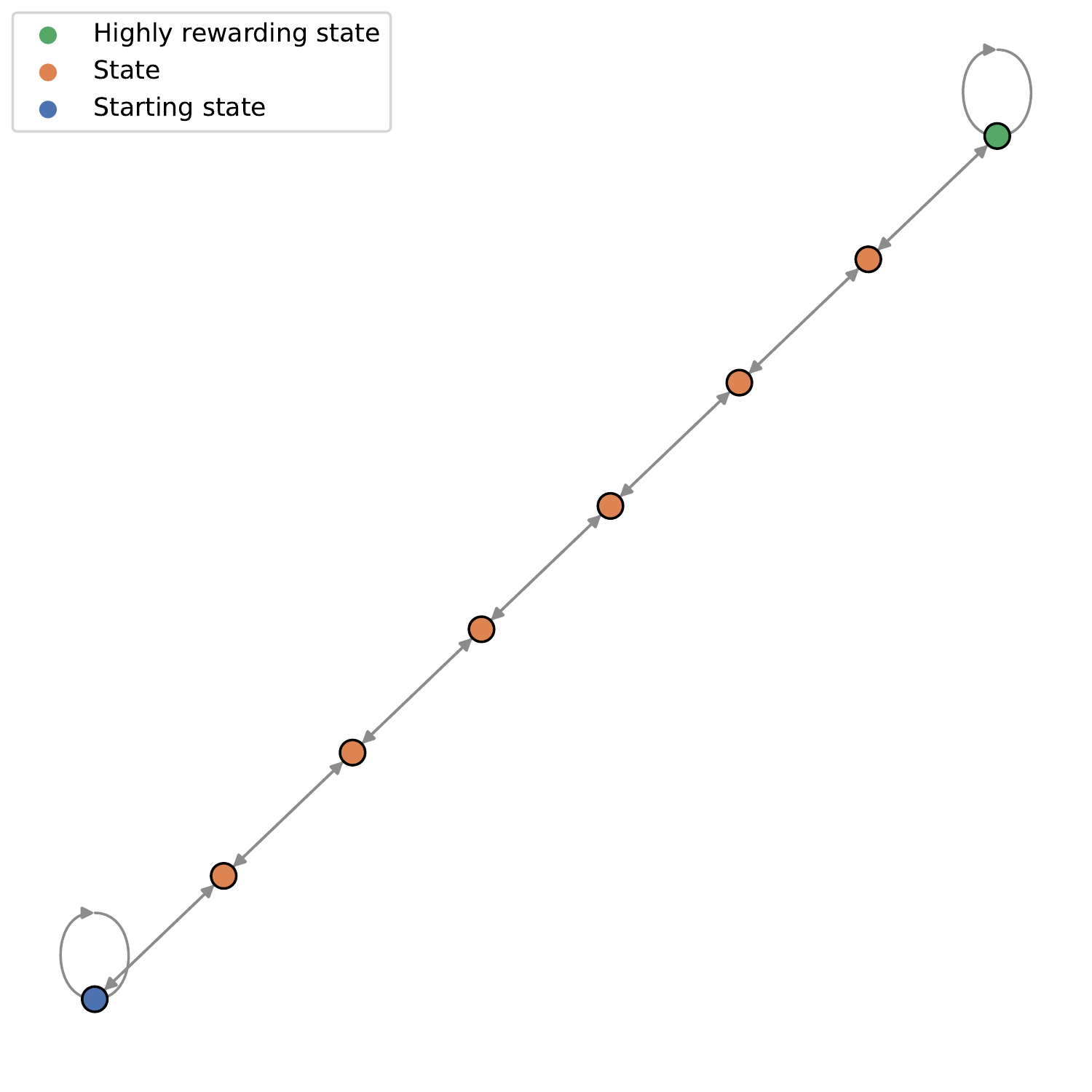}}
    \caption{\texttt{RiverSwim} MDP with length eight.}
    \label{fig:riverswim_mdp}
\end{figure}

The \texttt{RiverSwim} MDP has been introduced by \citet{strehl2008analysis} as a simple but challenging MDP.
This MDP is a chain of states where the agent can only move between adjacent states.
The agent starts in the leftmost state.
We have removed the \textit{current} mechanism proposed by \citet{strehl2008analysis}, which increases the difficulty of moving right since we provide more general controls (namely, \plazy and \prand).
The agent is given a small reward for staying in the initial state, but it can obtain a large reward in the rightmost state.
The challenge of \texttt{RiverSwim} is that the agent has to travel all the states in the chain in order to discover the highly rewarding state.
The chain structure is evident in the visual representations in Fig.~\ref{fig:riverswim_mdp}.
In the textual representation for the \texttt{RiverSwim} MDP, the letter \texttt{A} encodes the position of the agent, the letter \texttt{S} represents the starting state, and the letter \texttt{G} represents the position of the highly rewarding state.

\begin{table}[htb]
\captionsetup{position=top}
\centering%
\caption{RiverSwim emission map examples for a given state.}%
\begin{minipage}[c]{.24\textwidth}
\begin{center}
\resizebox{\textwidth}{!}{%
\begin{tabular}[b]{cccccccc}
    'S' & ' ' & ' ' & 'A' & ' ' & ' ' & ' ' & 'G'
\end{tabular}}
\end{center}
\end{minipage}%
\hfill%
\begin{minipage}[c]{.24\textwidth}
\begin{center}
$$
\begin{bmatrix}
     2.
\end{bmatrix}
$$
\end{center}
\end{minipage}%
\hfill%
\begin{minipage}[c]{.24\textwidth}
\includegraphics[width=1\textwidth]{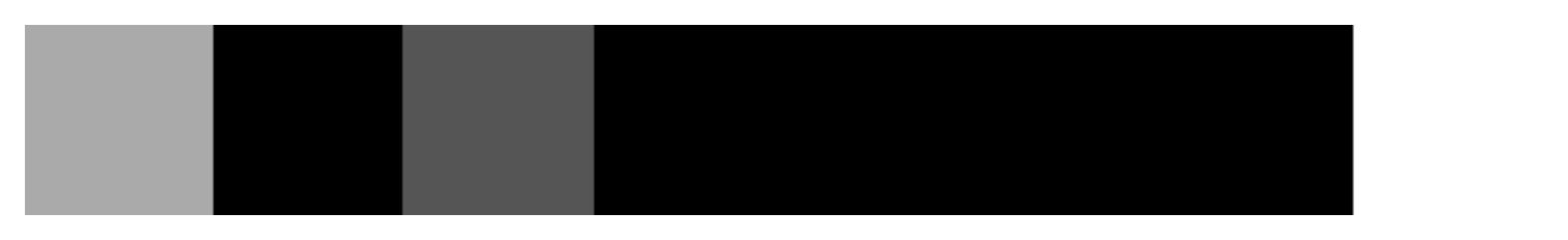}
\end{minipage}%
\hfill%
\begin{minipage}[c]{.24\textwidth}
\includegraphics[width=1\textwidth]{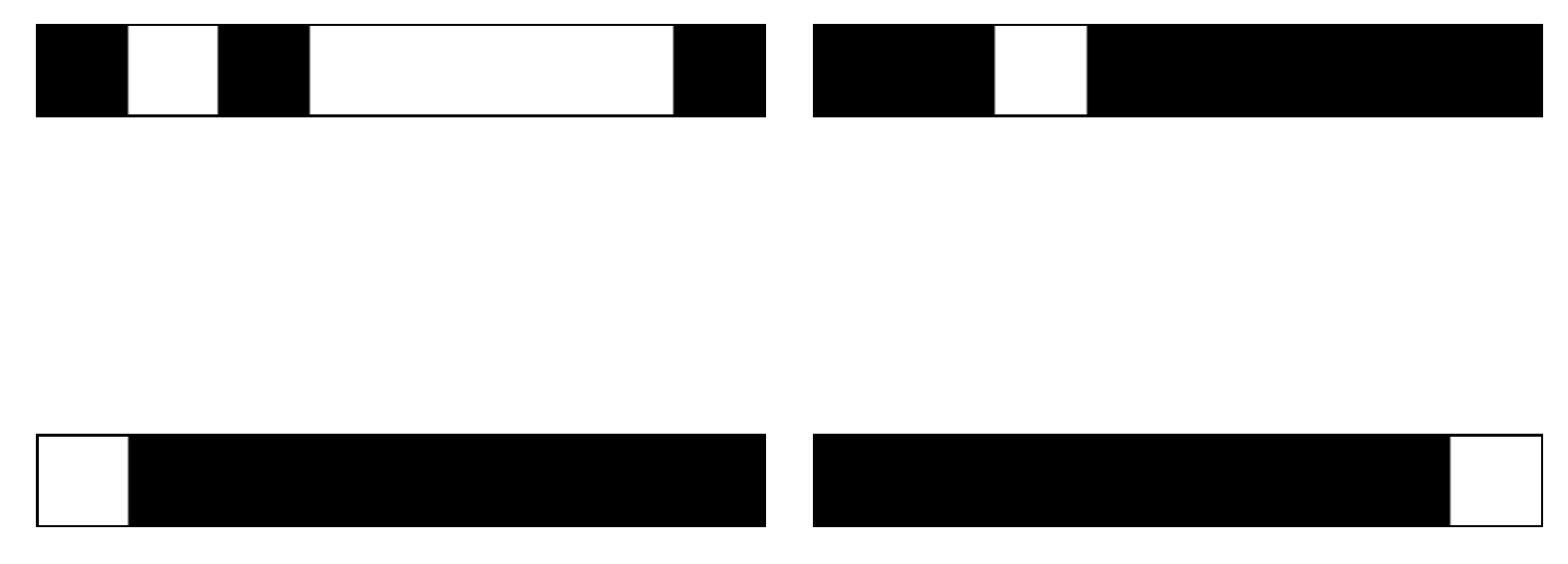}
\end{minipage}%
\vspace{0.1cm}
\begin{minipage}[c]{.03\textwidth}
(a)
\end{minipage}%
\begin{minipage}[c]{.21\textwidth}
\begin{center}
Textual state representation.
\end{center}
\end{minipage}%
\hfill%
\begin{minipage}[r]{.03\textwidth}
(b)
\end{minipage}%
\begin{minipage}[c]{.18\textwidth}
\begin{center}
State information emission map.
\end{center}
\end{minipage}%
\hfill%
\begin{minipage}[c]{.03\textwidth}
(c)
\end{minipage}%
\begin{minipage}[c]{.18\textwidth}
\begin{center}
Image encoding emission map.
\end{center}
\end{minipage}%
\hfill%
\begin{minipage}[c]{.03\textwidth}
(d)
\end{minipage}%
\begin{minipage}[c]{.18\textwidth}
\begin{center}
Tensor encoding emission map.
\end{center}
\end{minipage}%
\hfill%
\label{tab:rs_emissionmap}%
\end{table}%

\subsubsection{\texttt{DeepSea}}

\begin{figure}[!ht]
    \centering
    \subfloat[Continuous MDP representation.
    \label{fig:deepsea_cont}]{\includegraphics[width=0.4\textwidth]{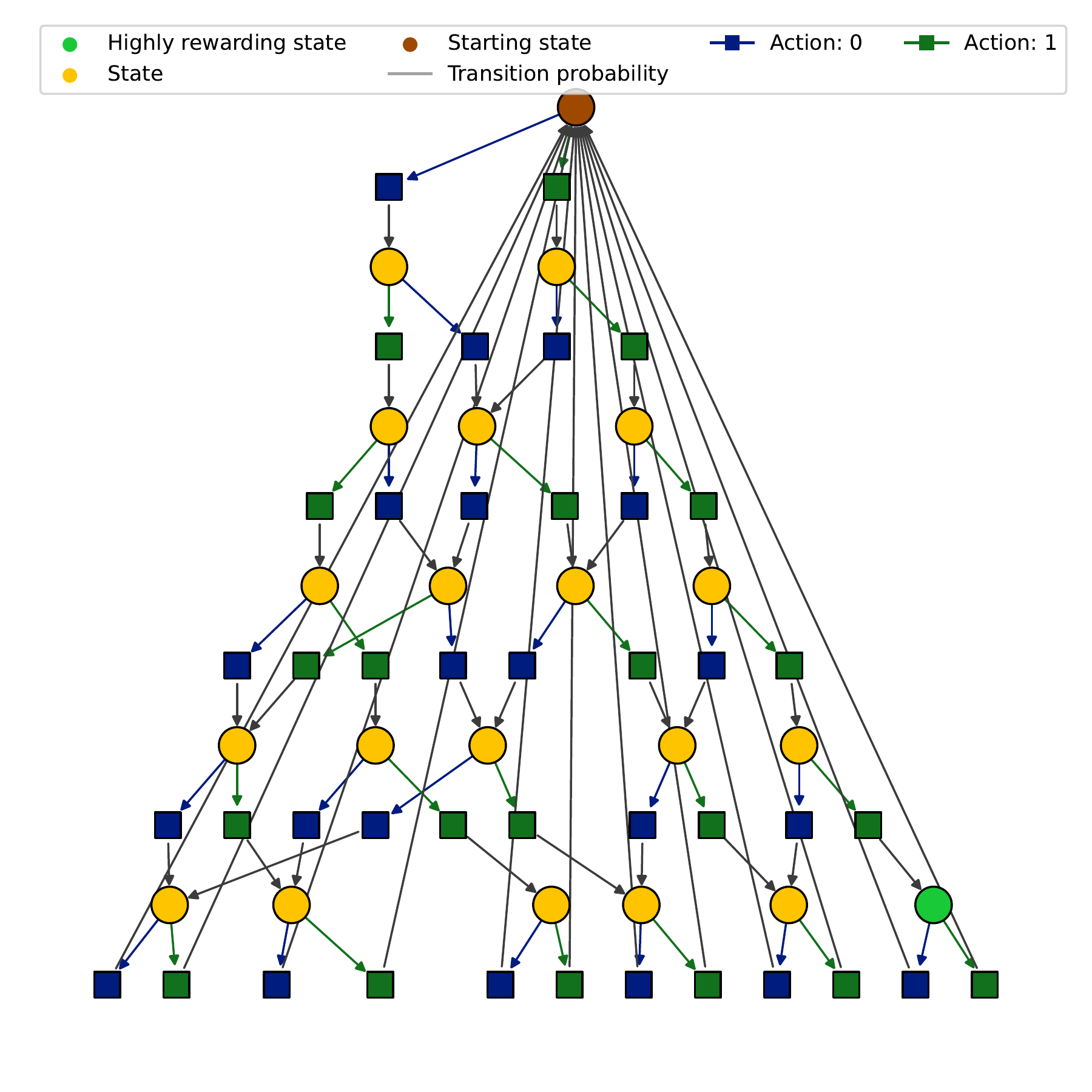}}
    \hfill
    \subfloat[Markov chain representation.
    \label{fig:deepsea_mc}]{\includegraphics[width=0.4\textwidth]{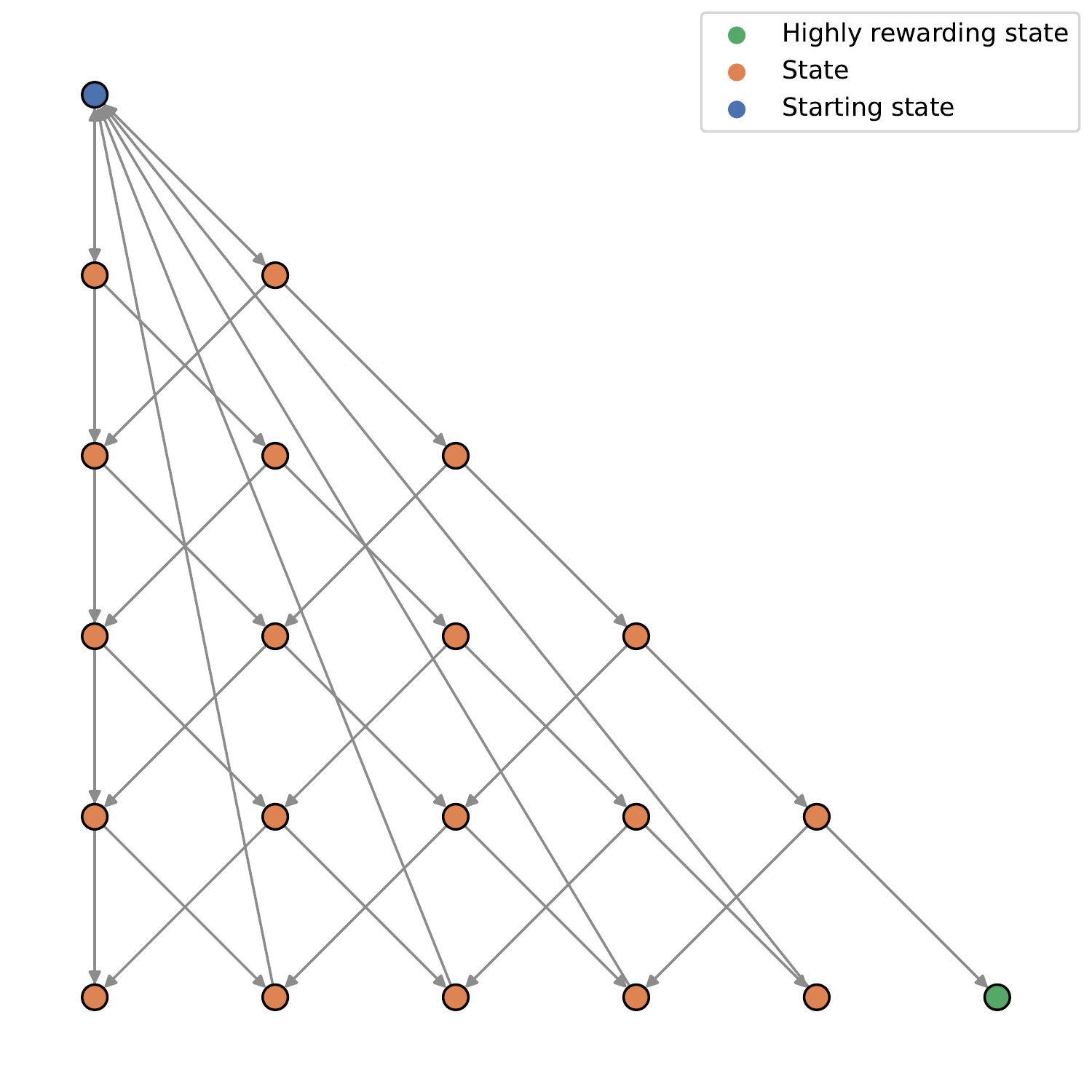}}
    \caption{\texttt{DeepSea} MDP with size six.}
    \label{fig:deepsea_mdp}
\end{figure}

The \texttt{DeepSea} MDP has been introduced by \citet{osband2019deep} as a \textit{deep exploration} challenge.
The MDP is a pyramid of states in which the agent starts at the top and dives a step down at each time step.
Depending on the action chosen, the agent can either dive to the right or to the left.
Once the agent reaches the base of the pyramid, it restarts from the top.
The agent is rewarded when diving to the left, but a large reward can be obtained by reaching the bottom rightmost state.
The main difficulty of \texttt{DeepSea} is that the agent will not be able to reach the highly rewarding state before it restarts if it chooses a single \textit{wrong} action.
We removed the \plazy parameter from \texttt{DeepSea} due to the particular structure of this MDP.
In the episodic setting, staying in the same state even once would make reaching the goal state impossible.
The pyramid structure is evident in the visual representations in Fig.~\ref{fig:deepsea_mdp}.
In the textual representation for the \texttt{DeepSea} MDP, the letter \texttt{A} corresponds the to position of the agent.

\begin{table}[htb]
\captionsetup{position=top}
\centering%
\caption{DeepSea emission map examples for a given state.}%
\begin{minipage}[c]{.24\textwidth}
\begin{center}
\resizebox{\textwidth}{!}{%
\begin{tabular}[b]{cccccc}
     ' ' & ' ' & ' ' & ' ' & ' ' & ' '\\
     ' ' & ' ' & ' ' & ' ' & ' ' & ' '\\
     ' ' & ' ' & ' ' & ' ' & ' ' & ' '\\
     ' ' & ' ' & 'A' & ' ' & ' ' & ' '\\
     ' ' & ' ' & ' ' & ' ' & ' ' & ' '\\
     ' ' & ' ' & ' ' & ' ' & ' ' & ' '\\
\end{tabular}}
\end{center}
\end{minipage}%
\hfill%
\begin{minipage}[c]{.24\textwidth}
\begin{center}
$$
\begin{bmatrix}
     2. \\ 2.
\end{bmatrix}
$$
\end{center}
\end{minipage}%
\hfill%
\begin{minipage}[c]{.24\textwidth}
\includegraphics[width=1\textwidth]{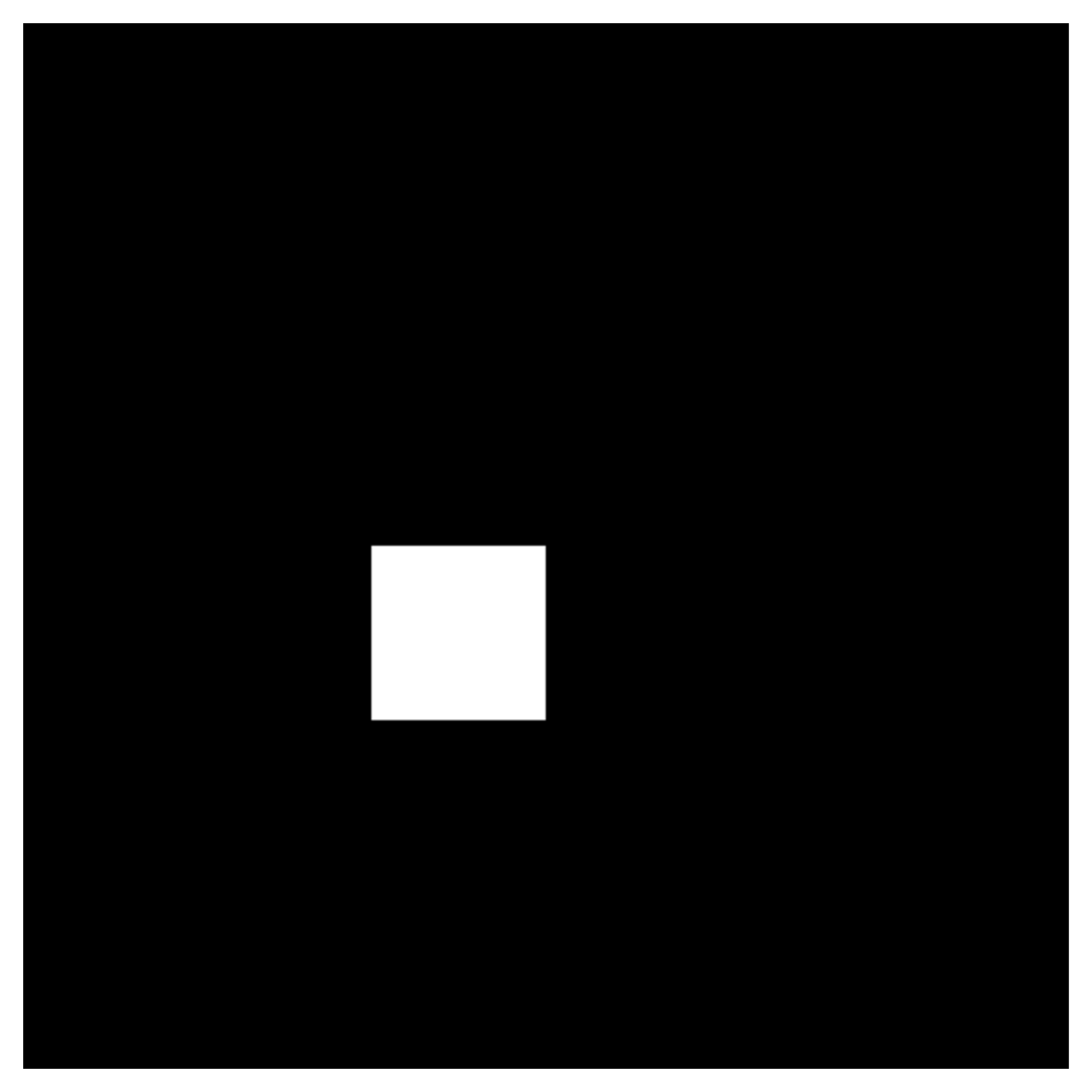}
\end{minipage}%
\hfill%
\begin{minipage}[c]{.24\textwidth}
\includegraphics[width=1\textwidth]{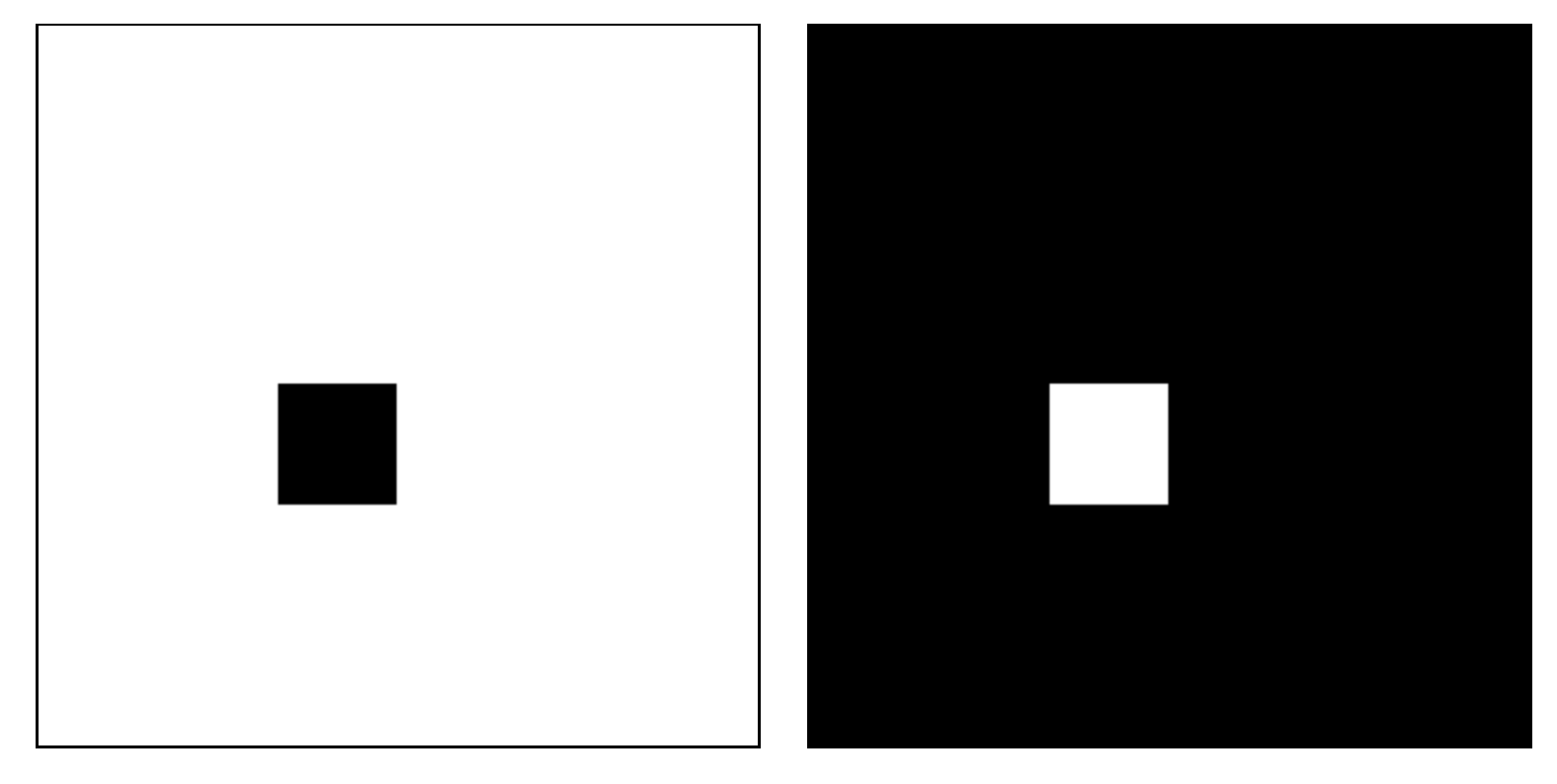}
\end{minipage}%
\vspace{0.1cm}
\begin{minipage}[c]{.03\textwidth}
(a)
\end{minipage}%
\begin{minipage}[c]{.21\textwidth}
\begin{center}
Textual state representation.
\end{center}
\end{minipage}%
\hfill%
\begin{minipage}[r]{.03\textwidth}
(b)
\end{minipage}%
\begin{minipage}[c]{.18\textwidth}
\begin{center}
State information emission map.
\end{center}
\end{minipage}%
\hfill%
\begin{minipage}[c]{.03\textwidth}
(c)
\end{minipage}%
\begin{minipage}[c]{.18\textwidth}
\begin{center}
Image encoding emission map.
\end{center}
\end{minipage}%
\hfill%
\begin{minipage}[c]{.03\textwidth}
(d)
\end{minipage}%
\begin{minipage}[c]{.18\textwidth}
\begin{center}
Tensor encoding emission map.
\end{center}
\end{minipage}%
\hfill%
\label{tab:ds_emission_map}%
\end{table}%

\subsubsection{\texttt{SimpleGrid}}

\begin{figure}[!ht]
    \centering
    \subfloat[Continuous MDP representation.
    \label{fig:simplegrid_cont}]{\includegraphics[width=0.4\textwidth]{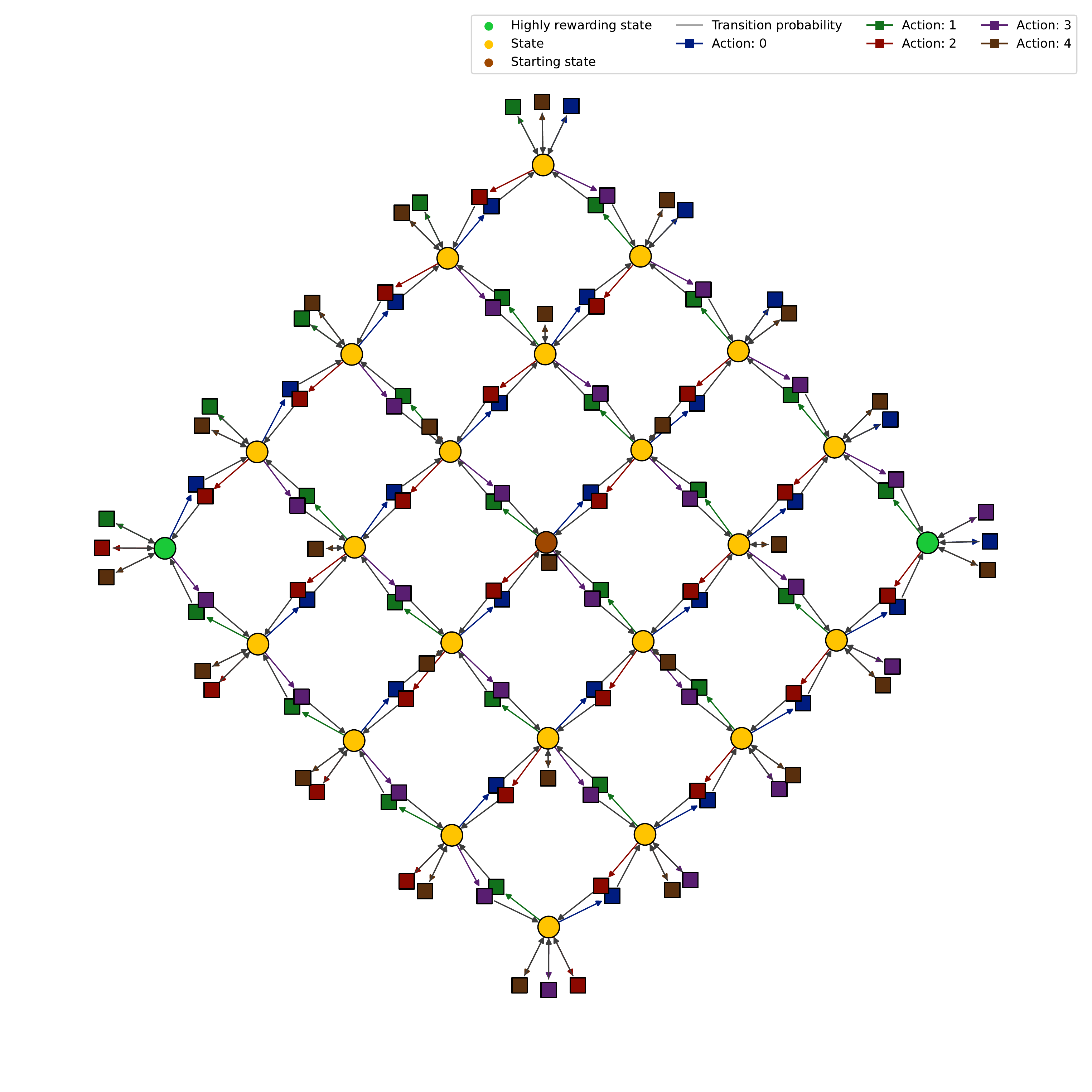}}
    \hfill
    \subfloat[Markov chain representation.
    \label{fig:simplegrid_mc}]{\includegraphics[width=0.4\textwidth]{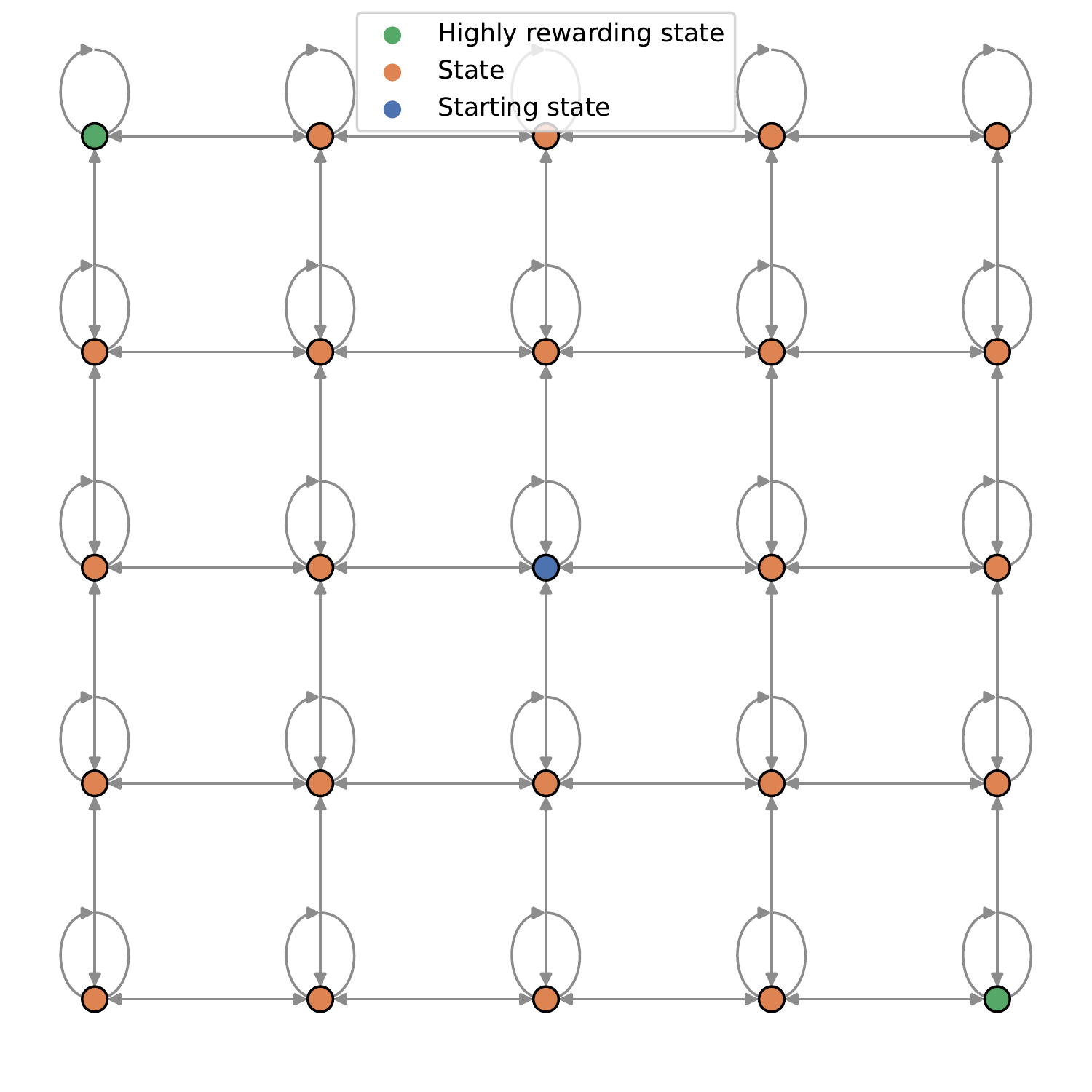}}
    \caption{\texttt{SimpleGrid} MDP with size five.}
    \label{fig:simple_mdp}
\end{figure}

The \texttt{SimpleGrid} MDP is a grid world in which the agent can either choose to move in one of the cardinal directions or stay in the same position.
At each time step, the agent is given a small reward.
Depending on the parameters of the MDP, some corner states yield a high reward whereas the other ones produce close to zero rewards.
The starting states are selected from the states in the center of the grid in order to be as far as possible from the corners.
The grid structure is clearly distinguishable in the visual representation in Fig.~\ref{fig:simple_mdp}.
In the textual representation for the \texttt{SimpleGrid} MDP, the letter \texttt{A} encodes the position of the agent and the symbols \texttt{$+$} and \texttt{$-$} represent the states that yield large reward and zero reward, respectively.
\begin{table}[htb]
\captionsetup{position=top}
\centering%
\caption{SimpleGrid emission map examples for a given state.}%
\begin{minipage}[c]{.24\textwidth}
\begin{center}
\resizebox{\textwidth}{!}{%
\begin{tabular}[b]{cccc}
         '+' & ' ' & ' ' & '-'\\
         ' ' & ' ' & ' ' & 'A'\\
         ' ' & ' ' & ' ' & ' '\\
         '-' & ' ' & ' ' & '+',\\
\end{tabular}
}
\end{center}
\end{minipage}%
\hfill%
\begin{minipage}[c]{.24\textwidth}
\begin{center}
$$
\begin{bmatrix}
     3. \\ 2.
\end{bmatrix}
$$
\end{center}
\end{minipage}%
\hfill%
\begin{minipage}[c]{.24\textwidth}
\includegraphics[width=1\textwidth]{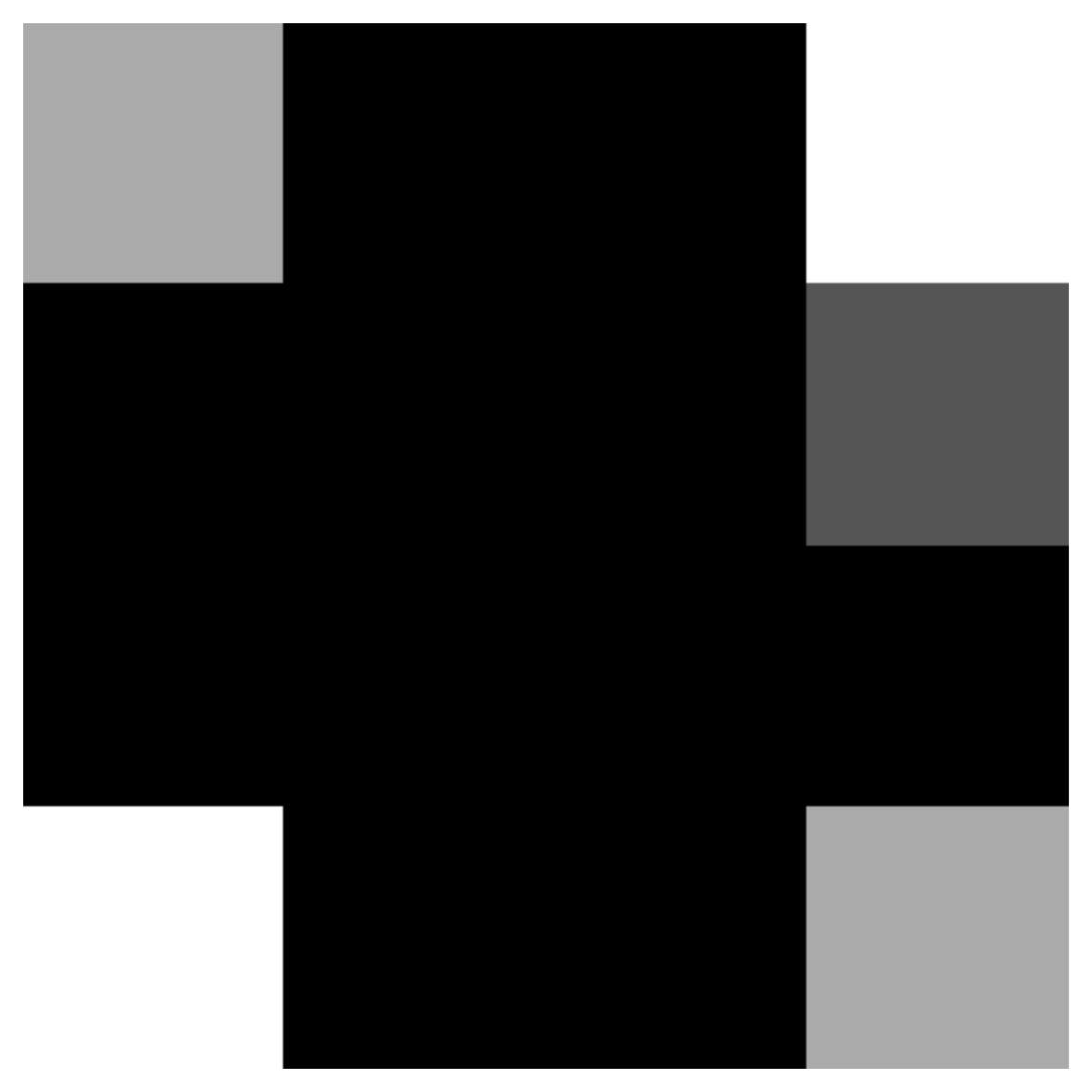}
\end{minipage}%
\hfill%
\begin{minipage}[c]{.24\textwidth}
\includegraphics[width=1\textwidth]{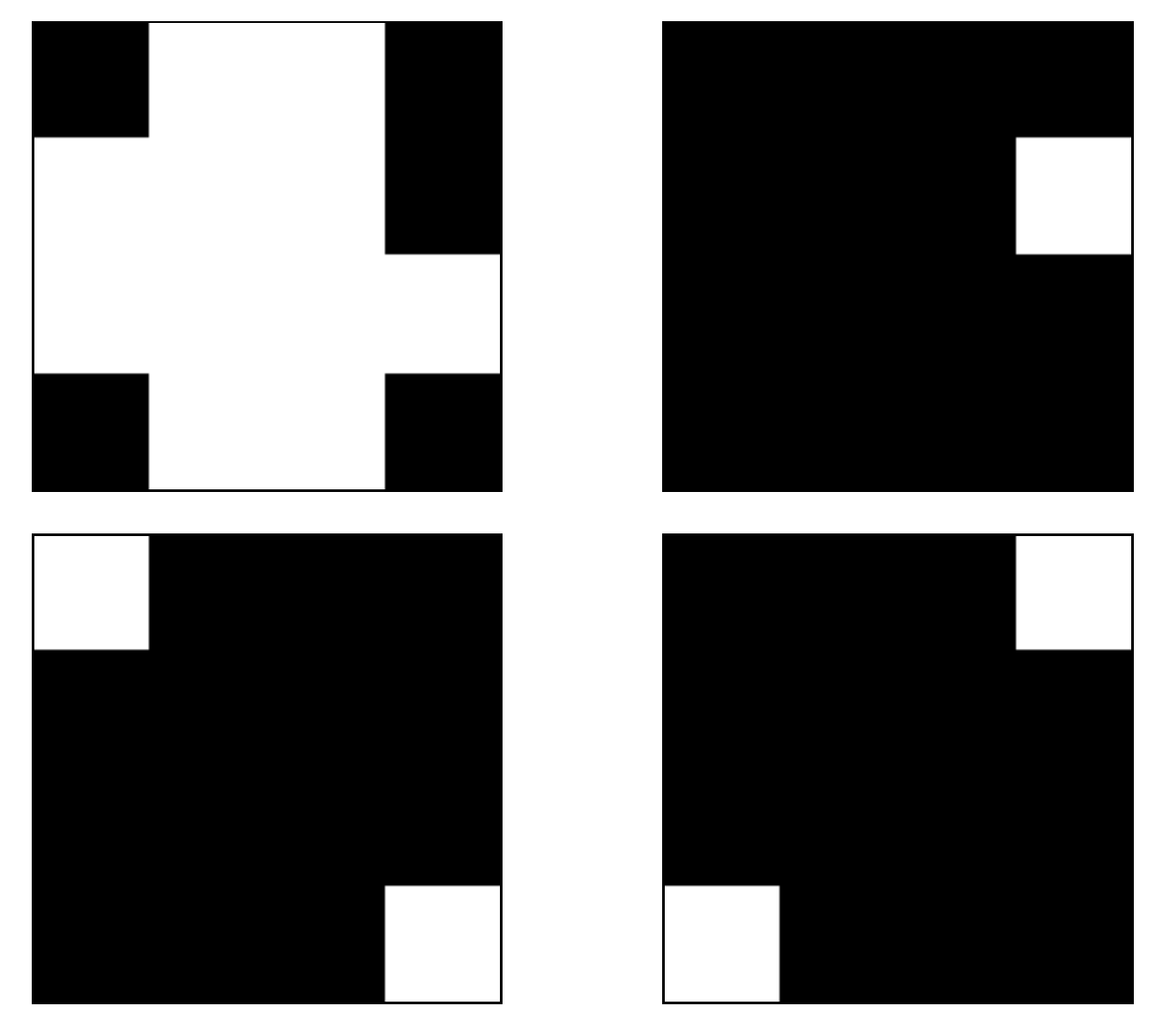}
\end{minipage}%
\vspace{0.1cm}
\begin{minipage}[c]{.03\textwidth}
(a)
\end{minipage}%
\begin{minipage}[c]{.21\textwidth}
\begin{center}
Textual state representation.
\end{center}
\end{minipage}%
\hfill%
\begin{minipage}[r]{.03\textwidth}
(b)
\end{minipage}%
\begin{minipage}[c]{.18\textwidth}
\begin{center}
State information emission map.
\end{center}
\end{minipage}%
\hfill%
\begin{minipage}[c]{.03\textwidth}
(c)
\end{minipage}%
\begin{minipage}[c]{.18\textwidth}
\begin{center}
Image encoding emission map.
\end{center}
\end{minipage}%
\hfill%
\begin{minipage}[c]{.03\textwidth}
(d)
\end{minipage}%
\begin{minipage}[c]{.18\textwidth}
\begin{center}
Tensor encoding emission map.
\end{center}
\end{minipage}%
\hfill%
\label{tab:sg_emissionmap}%
\end{table}%

\subsubsection{MiniGrid MDPs} \label{app:minigrid_mdps}

Gym MiniGrid (MG) \citep{gym_minigrid} is an important testbed for non-tabular \rl agents.
It presents several families of MDPs that produce different challenges.
The base structure is a grid world where an agent can move by going forward, rotating left, and rotating right.
Depending on the MDP family, the agent can have further access to actions such as pick, drop, and interact.
The goal is always to reach a highly rewarding state.
\colosseum implements three families of MiniGrid MDPs: \texttt{MG-Empty}, \texttt{MG-Rooms}, and \texttt{MG-DoorKey}.

\begin{figure}[!ht]
    \centering
    \subfloat[Continuous MDP representation.
    \label{fig:minigridempty_cont}]{\includegraphics[width=0.4\textwidth]{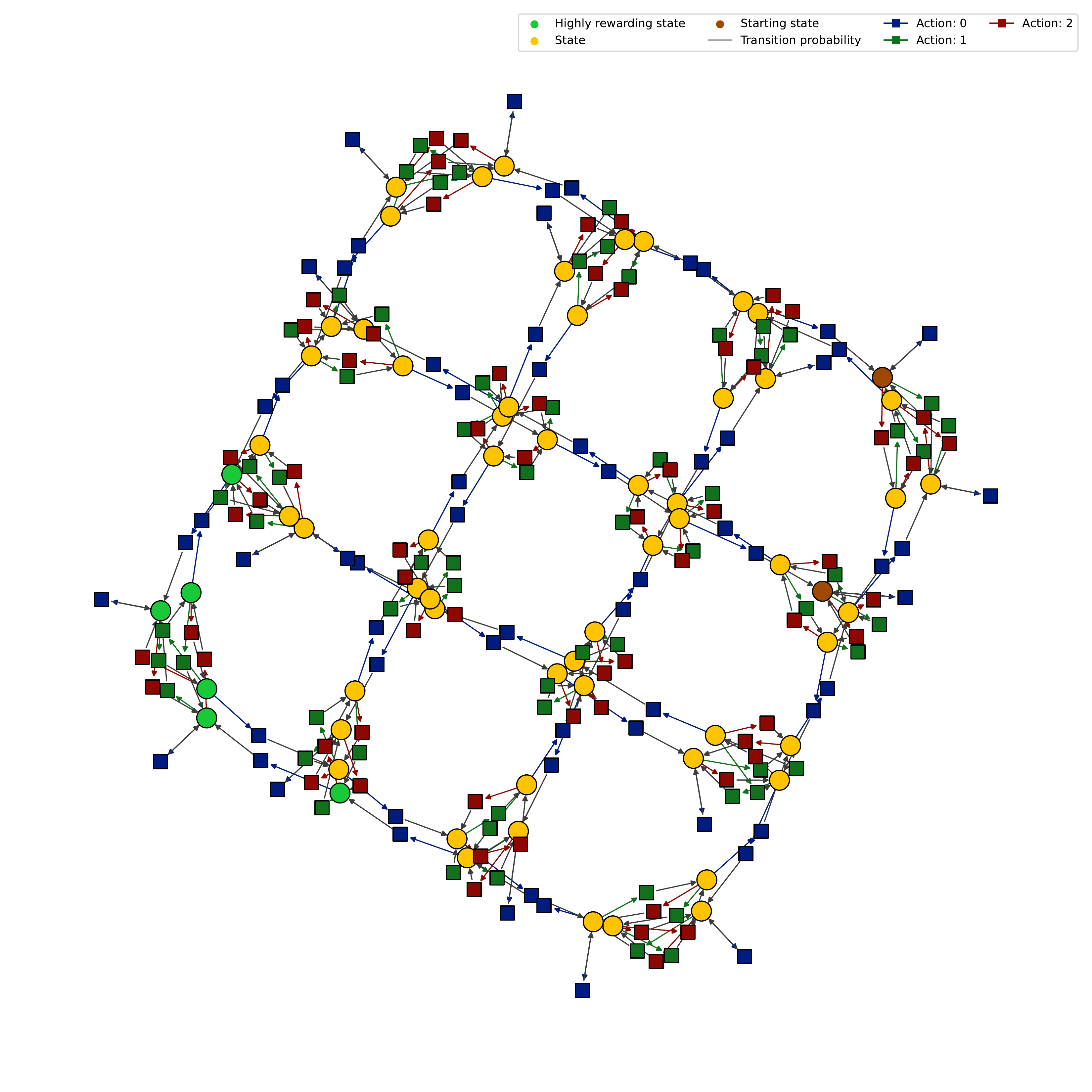}}
    \hfill
    \subfloat[Markov chain representation.
    \label{fig:minigridempty_mc}]{\includegraphics[width=0.4\textwidth]{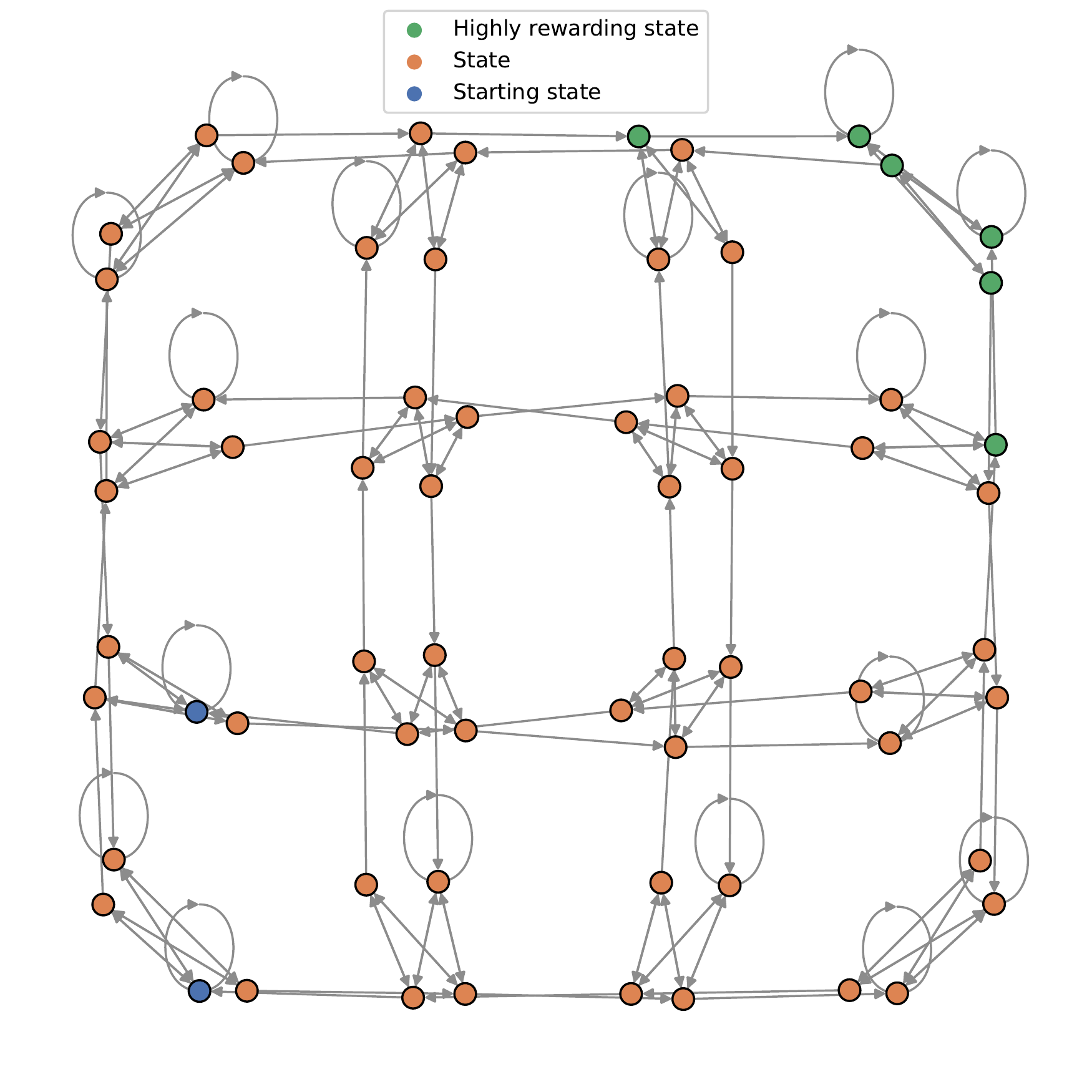}}
    \caption{\texttt{MG-Empty} MDP with size four.}
    \label{fig:empty_mdp}
\end{figure}

The \texttt{MG-Empty} MDP contains only the basic structure of an MG environment.
The starting states are selected from a randomly selected border of the grid and the highly rewarding state is located on the border at the opposite side of the grid.
Although \texttt{MG-Empty} appears similar to \texttt{SimpleGrid}, it implements a more complex mechanism to move between states that results in a completely different transition structure, which is evident in the visual representations in Fig.~\ref{fig:empty_mdp}.
Each group of four states corresponds to the agent rotating in the same position.
In the textual representation for the \texttt{MG-Empty} MDP, the symbols \texttt{>,v,<}, and \texttt{$\wedge$} encode the position and the rotation of the agent, whereas the letter \texttt{G} represents the position of the goal.

\begin{table}[htb]
\captionsetup{position=top}
\centering%
\caption{MiniGrid-Empty emission map examples for a given state.}%
\begin{minipage}[c]{.24\textwidth}
\begin{center}
\resizebox{\textwidth}{!}{%
\begin{tabular}[b]{ccccc}
     ' ' & ' ' & ' ' & ' '\\
     '>' & ' ' & ' ' & ' '\\
     ' ' & ' ' & ' ' & ' '\\
     ' ' & ' ' & ' ' & 'G'\\
\end{tabular}}
\end{center}
\end{minipage}%
\hfill%
\begin{minipage}[c]{.24\textwidth}
\begin{center}
$$
\begin{bmatrix}
     0. \\ 1. \\ 3.
\end{bmatrix}
$$
\end{center}
\end{minipage}%
\hfill%
\begin{minipage}[c]{.24\textwidth}
\includegraphics[width=1\textwidth]{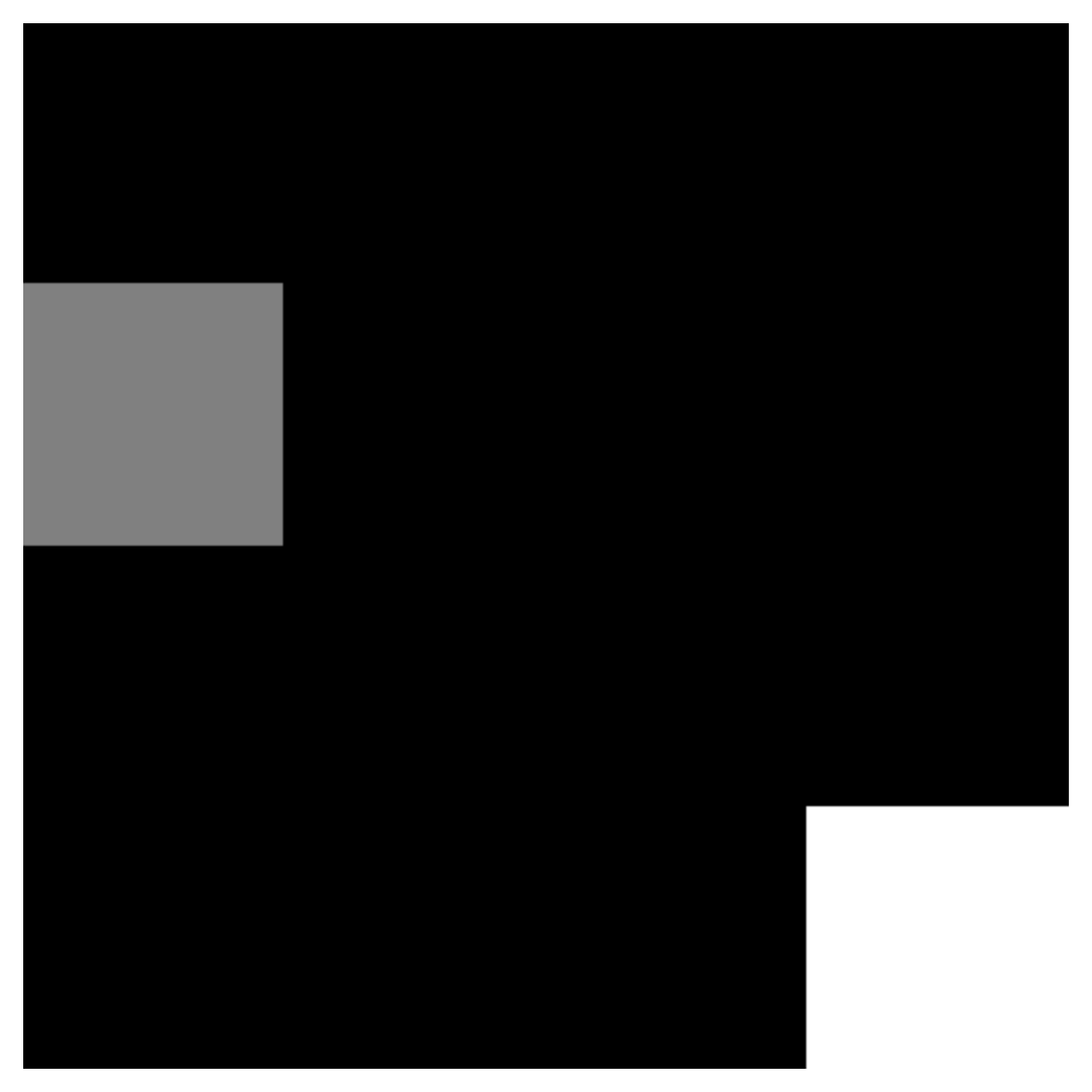}
\end{minipage}%
\hfill%
\begin{minipage}[c]{.24\textwidth}
\includegraphics[width=1\textwidth]{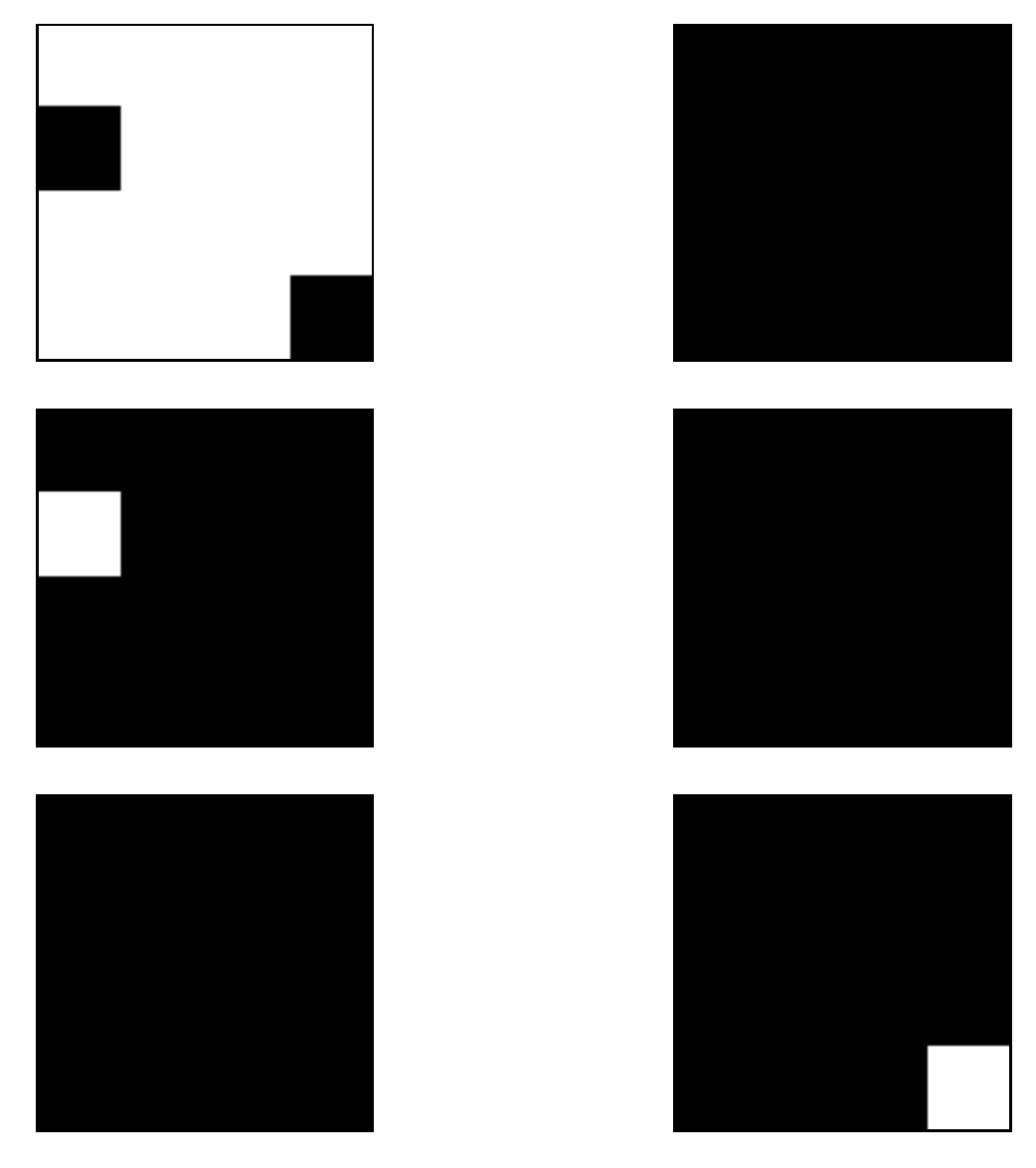}
\end{minipage}%
\vspace{0.1cm}
\begin{minipage}[c]{.03\textwidth}
(a)
\end{minipage}%
\begin{minipage}[c]{.21\textwidth}
\begin{center}
Textual state representation.
\end{center}
\end{minipage}%
\hfill%
\begin{minipage}[r]{.03\textwidth}
(b)
\end{minipage}%
\begin{minipage}[c]{.18\textwidth}
\begin{center}
State information emission map.
\end{center}
\end{minipage}%
\hfill%
\begin{minipage}[c]{.03\textwidth}
(c)
\end{minipage}%
\begin{minipage}[c]{.18\textwidth}
\begin{center}
Image encoding emission map.
\end{center}
\end{minipage}%
\hfill%
\begin{minipage}[c]{.03\textwidth}
(d)
\end{minipage}%
\begin{minipage}[c]{.18\textwidth}
\begin{center}
Tensor encoding emission map.
\end{center}
\end{minipage}%
\hfill%
\label{tab:mge_emissionmap}%
\end{table}%

\begin{figure}[!ht]
    \centering
    \subfloat[Continuous MDP representation.
    \label{fig:minigridrooms_cont}]{\includegraphics[width=0.4\textwidth]{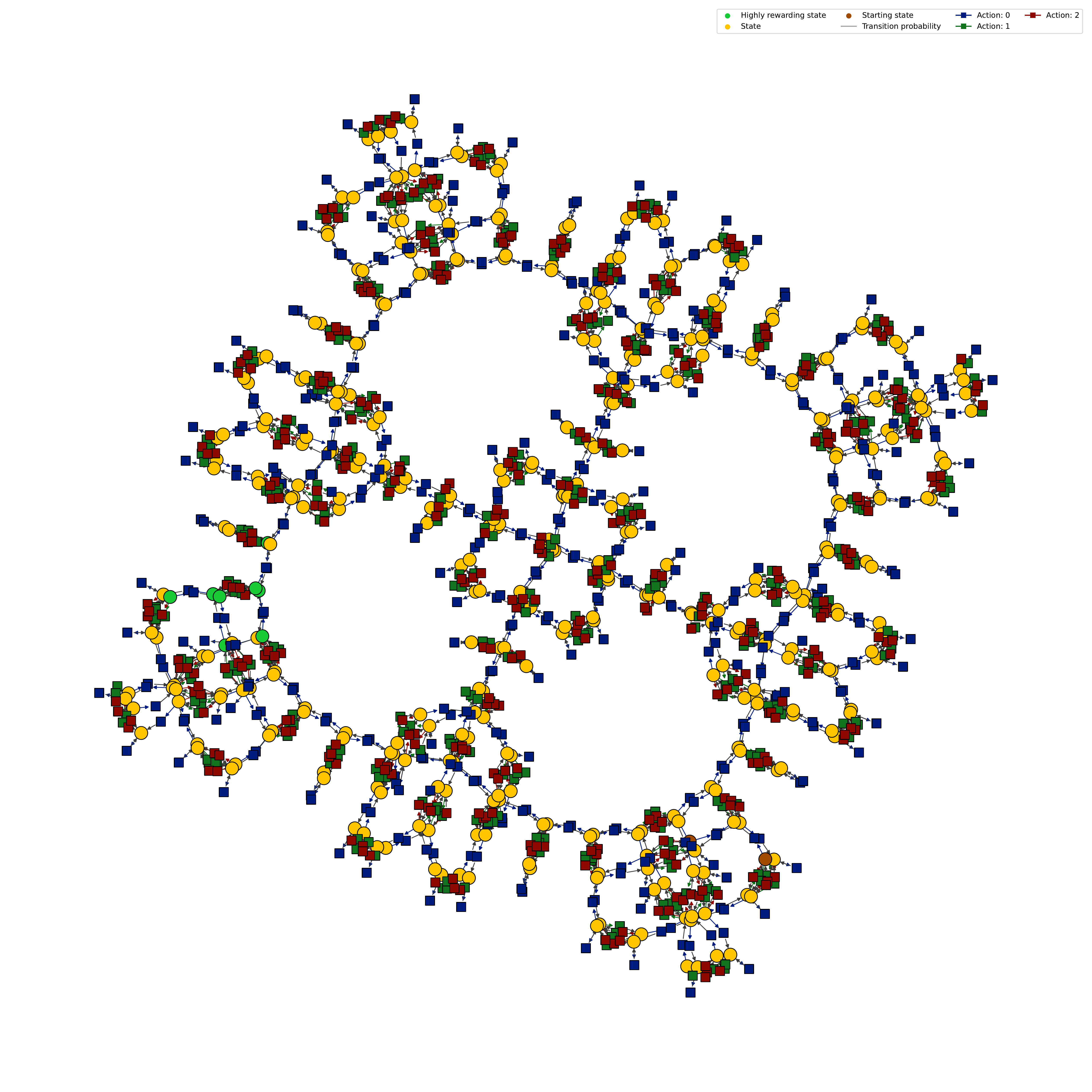}}
    \hfill
    \subfloat[Markov chain representation.
    \label{fig:minigridrooms_mc}]{\includegraphics[width=0.4\textwidth]{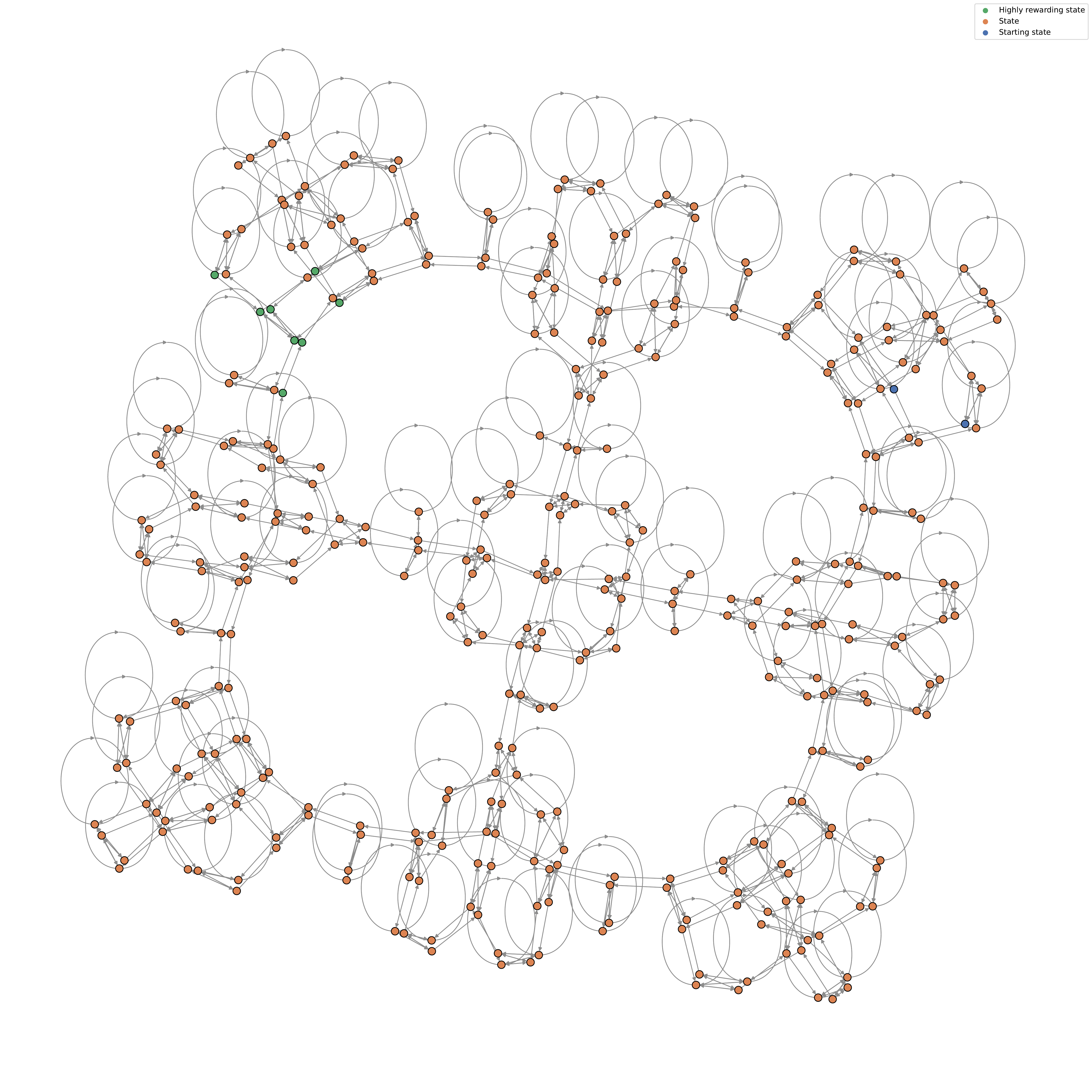}}
    \caption{\texttt{MG-Rooms} MDP with nine rooms of size three.}
    \label{fig:doors_mdp}
\end{figure}

The \texttt{MG-Rooms} is a collection of grids connected with narrow passages.
The presence of such bottlenecks produces a significantly higher challenge for exploration when compared to open grids, especially when \prand is non-zero.
The rooms are evident in the visual representations in Fig.~\ref{fig:doors_mdp}.
In the textual representation for the \texttt{MG-Rooms} MDP, the symbols \texttt{>,v,<}, and \texttt{$\wedge$} encode the position and the rotation of the agent, the letter \texttt{W} represents a wall, and the letter \texttt{G} represents the goal.
\begin{table}[htb]
\captionsetup{position=top}
\centering%
\caption{MiniGrid-Rooms emission map examples for a given state.}%
\begin{minipage}[c]{.24\textwidth}
\begin{center}
\resizebox{\textwidth}{!}{%
\begin{tabular}[b]{ccccccccc}
' ' & 'G' & ' ' & ' ' & 'W' & ' ' & ' ' & ' ' & ' '\\
' ' & ' ' & ' ' & ' ' & ' ' & ' ' & ' ' & ' ' & ' '\\
' ' & ' ' & ' ' & ' ' & 'W' & ' ' & ' ' & ' ' & ' '\\
' ' & ' ' & ' ' & ' ' & 'W' & ' ' & '>' & ' ' & ' '\\
'W' & 'W' & ' ' & 'W' & 'W' & 'W' & 'W' & ' ' & 'W'\\
' ' & ' ' & ' ' & ' ' & 'W' & ' ' & ' ' & ' ' & ' '\\
' ' & ' ' & ' ' & ' ' & ' ' & ' ' & ' ' & ' ' & ' '\\
' ' & ' ' & ' ' & ' ' & 'W' & ' ' & ' ' & ' ' & ' '\\
' ' & ' ' & ' ' & ' ' & 'W' & ' ' & ' ' & ' ' & ' '\\
\end{tabular}}
\end{center}
\end{minipage}%
\hfill%
\begin{minipage}[c]{.24\textwidth}
\begin{center}
$$
\begin{bmatrix}
     6. \\ 5. \\ 2.
\end{bmatrix}
$$
\end{center}
\end{minipage}%
\hfill%
\begin{minipage}[c]{.24\textwidth}
\includegraphics[width=1\textwidth]{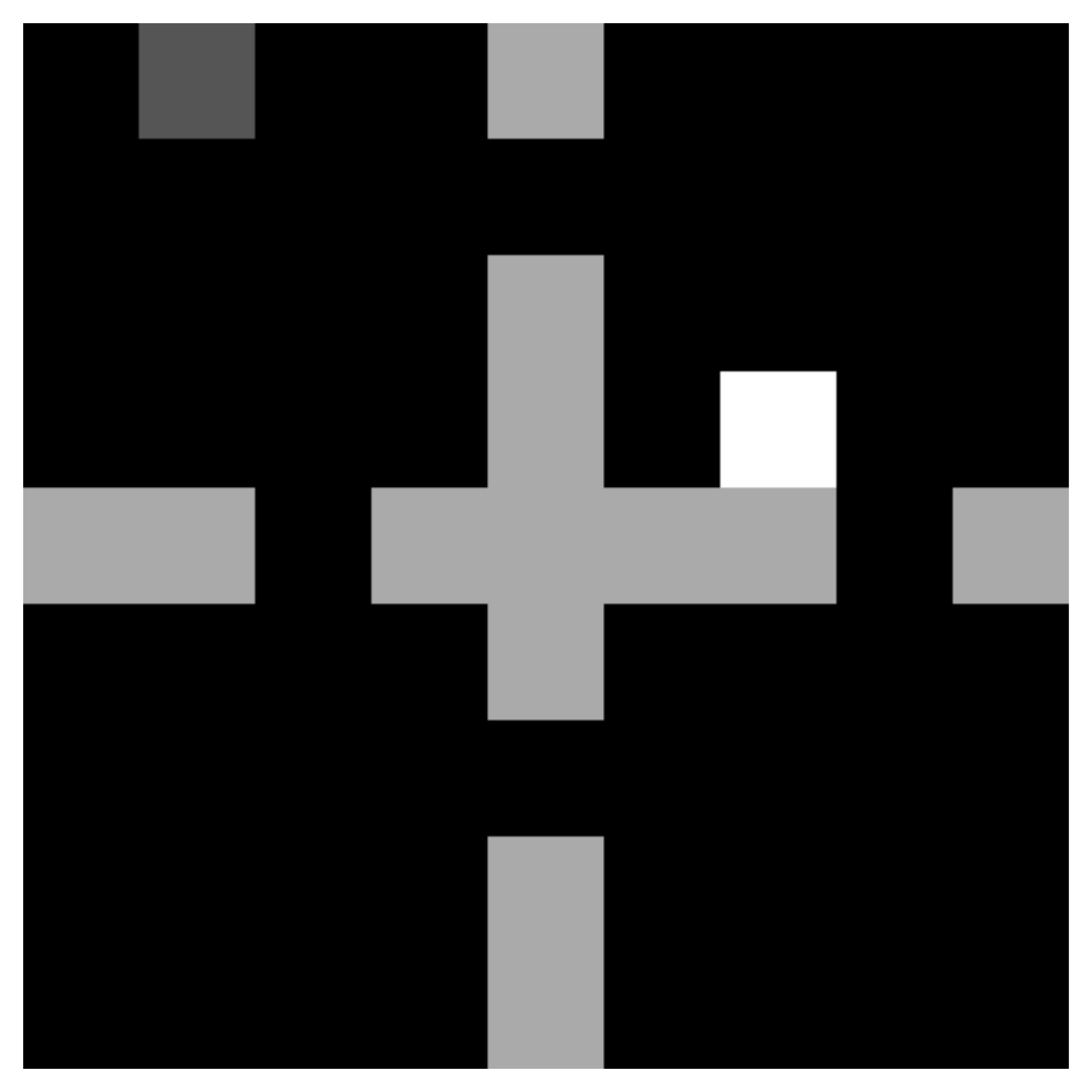}
\end{minipage}%
\hfill%
\begin{minipage}[c]{.24\textwidth}
\includegraphics[width=1\textwidth]{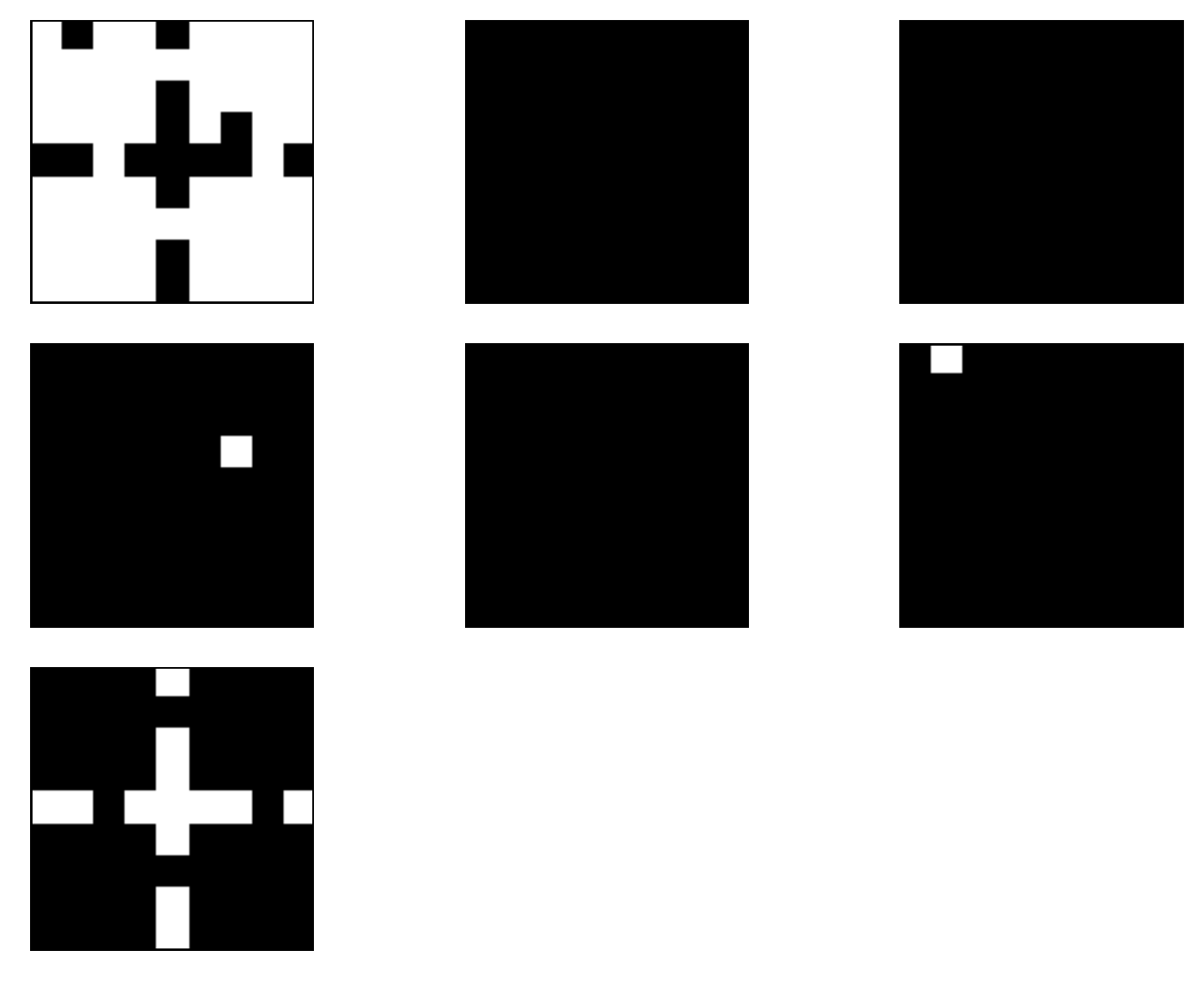}
\end{minipage}%
\vspace{0.1cm}
\begin{minipage}[c]{.03\textwidth}
(a)
\end{minipage}%
\begin{minipage}[c]{.21\textwidth}
\begin{center}
Textual state representation.
\end{center}
\end{minipage}%
\hfill%
\begin{minipage}[r]{.03\textwidth}
(b)
\end{minipage}%
\begin{minipage}[c]{.18\textwidth}
\begin{center}
State information emission map.
\end{center}
\end{minipage}%
\hfill%
\begin{minipage}[c]{.03\textwidth}
(c)
\end{minipage}%
\begin{minipage}[c]{.18\textwidth}
\begin{center}
Image encoding emission map.
\end{center}
\end{minipage}%
\hfill%
\begin{minipage}[c]{.03\textwidth}
(d)
\end{minipage}%
\begin{minipage}[c]{.18\textwidth}
\begin{center}
Tensor encoding emission map.
\end{center}
\end{minipage}%
\hfill%
\label{tab:mgt_emissionmap}%
\end{table}%

\begin{figure}[!ht]
    \centering
    \subfloat[Continuous MDP representation.
    \label{fig:minigriddoorkey_cont}]{\includegraphics[width=0.4\textwidth]{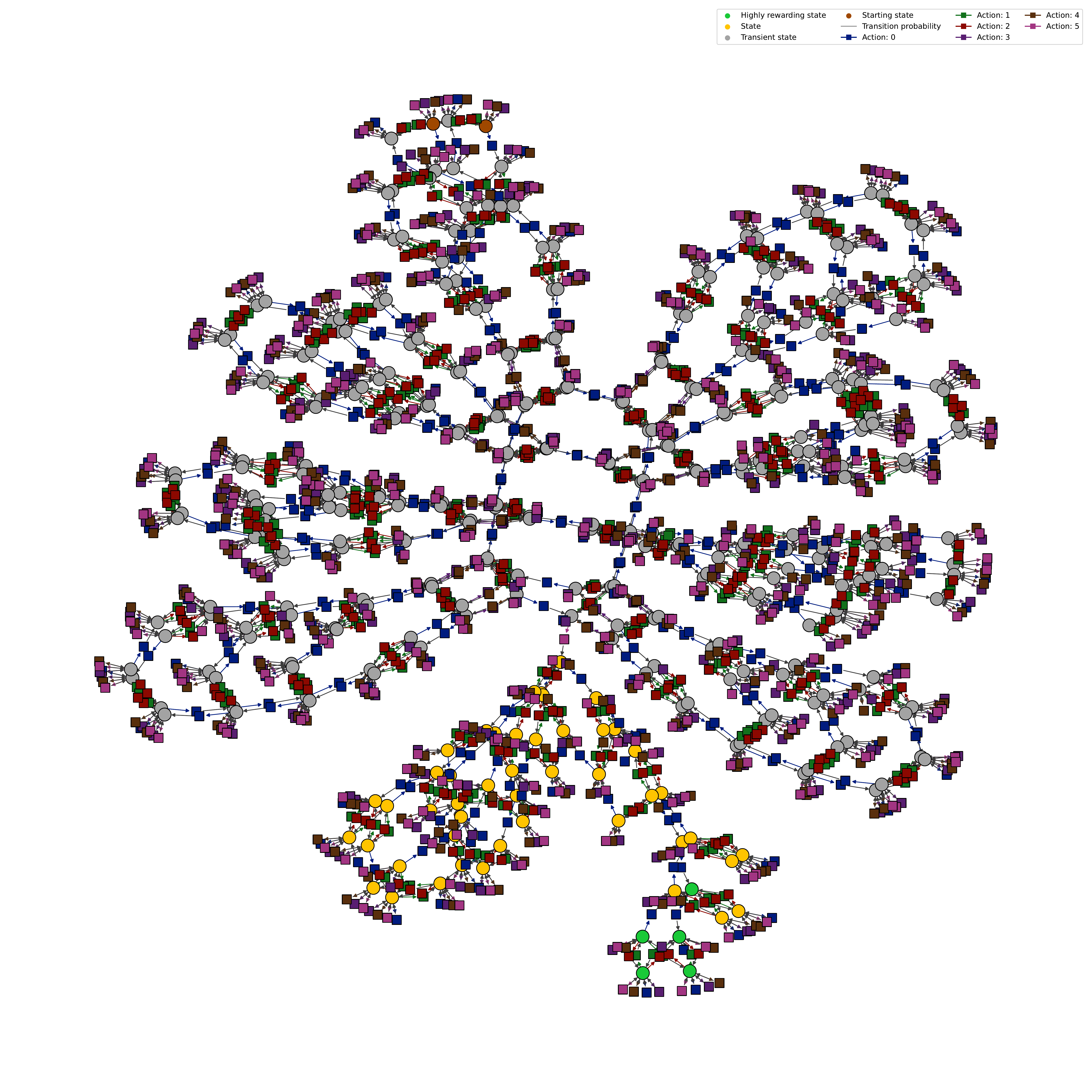}}
    \hfill
    \subfloat[Markov chain representation.
    \label{fig:minigriddoorkey_mc}]{\includegraphics[width=0.4\textwidth]{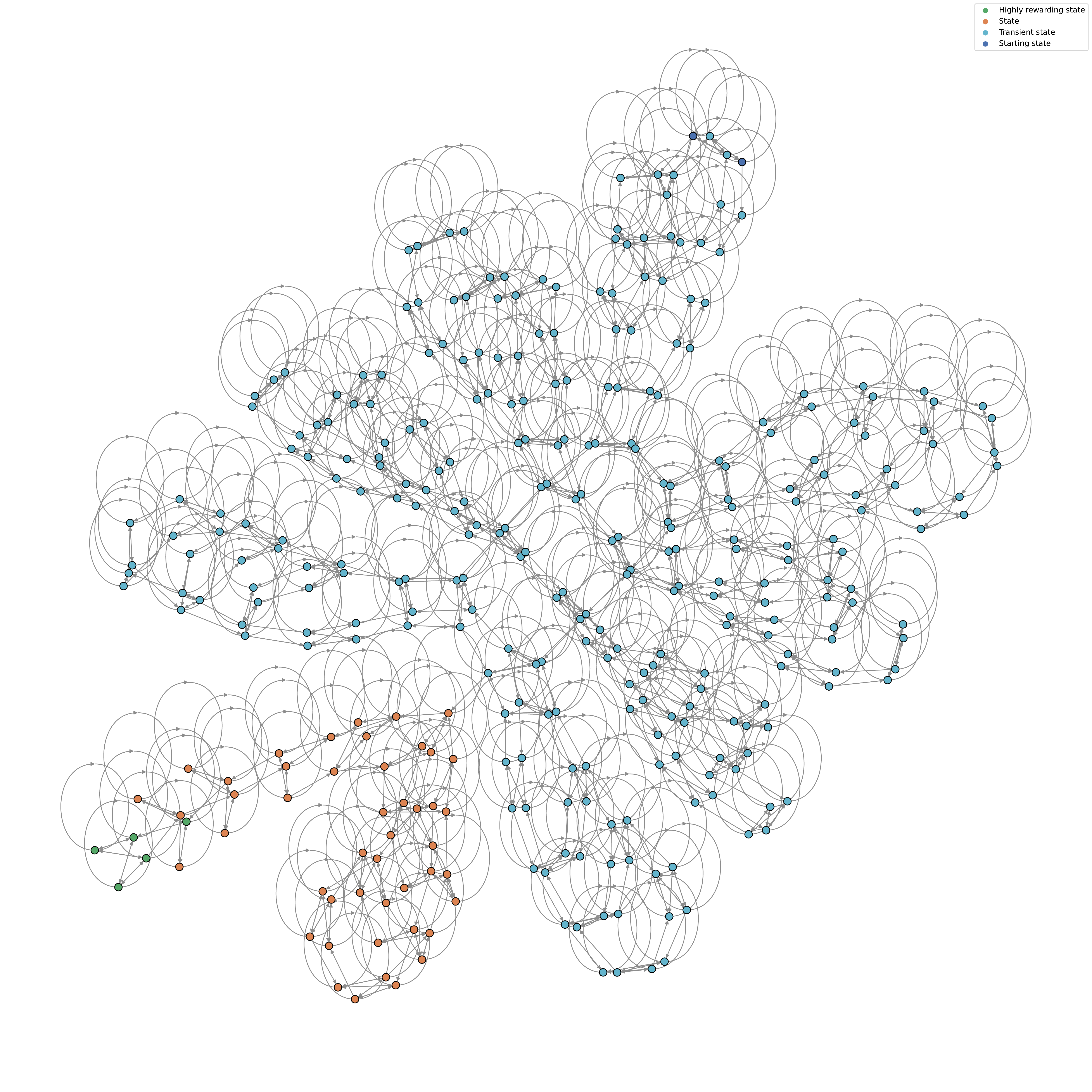}}
    \caption{\texttt{MG-DoorKey} MDP with size four.}
    \label{fig:doorkey_mdp}
\end{figure}

In the \texttt{MG-DoorKey} environment, the grid world is divided into two rooms separated by a wall with a door that can be opened using a key.
The key is positioned in the same room where the agent starts.
Differently from \texttt{MG-Empty} and \texttt{MG-Rooms}, in this case, the agent has all the five actions available to allow it to interact with the key and the door.
\texttt{MG-DoorKey} is a particularly challenging MDP for several reasons.
First, the agent has to take a very long sequence of actions before reaching the highly rewarding state.
Further, the action that picks the key produces effect only in the very few states in which the agent is correctly positioned in front of the key, and the action that opens the door only has effect in the single state in which the agent has the key and is correctly positioned in front of the door.
In this case, it is not possible to clearly identify the structure of the grid in the visual representations in Fig.~\ref{fig:doorkey_mdp}.
In the continuous setting, every MDP instance from this class is weakly-communicating.
Once the door has been opened, it is not possible to close it.
In the textual representation for the \texttt{MG-DoorKey} MDP, the symbols \texttt{>,v,<}, and \texttt{$\wedge$} encode the position and the rotation of the agent, the letter \texttt{K} represents the key, the letters \texttt{C} and \texttt{O} respectively stand for closed door and opened door, and the letter \texttt{G} represents the goal.

\subsubsection{\texttt{FrozenLake}}

\begin{figure}[!ht]
    \centering
    \subfloat[Continuous MDP representation.
    \label{fig:frozenlake_cont}]{\includegraphics[width=0.4\textwidth]{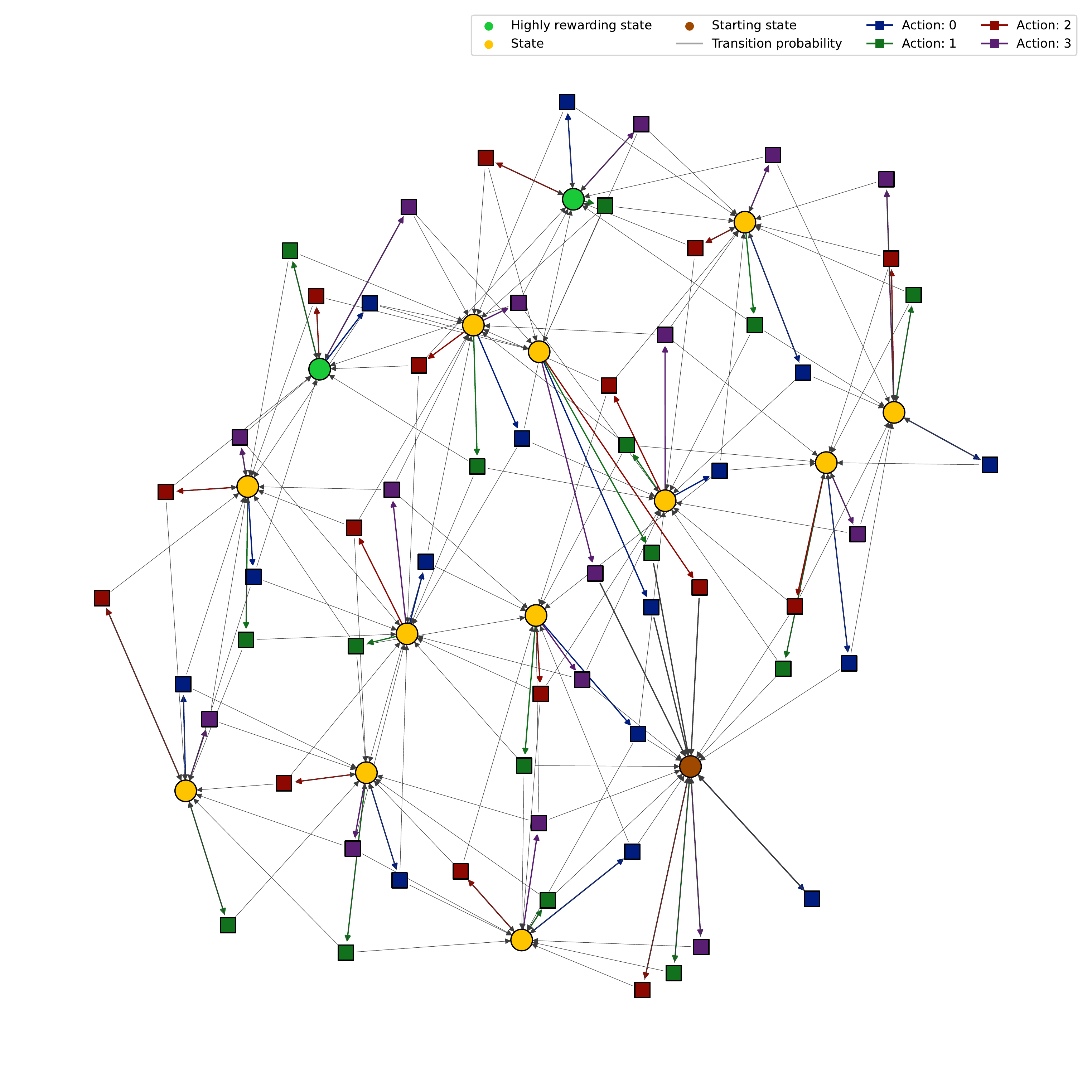}}
    \hfill
    \subfloat[Markov chain representation.
    \label{fig:frozenlake_mc}]{\includegraphics[width=0.4\textwidth]{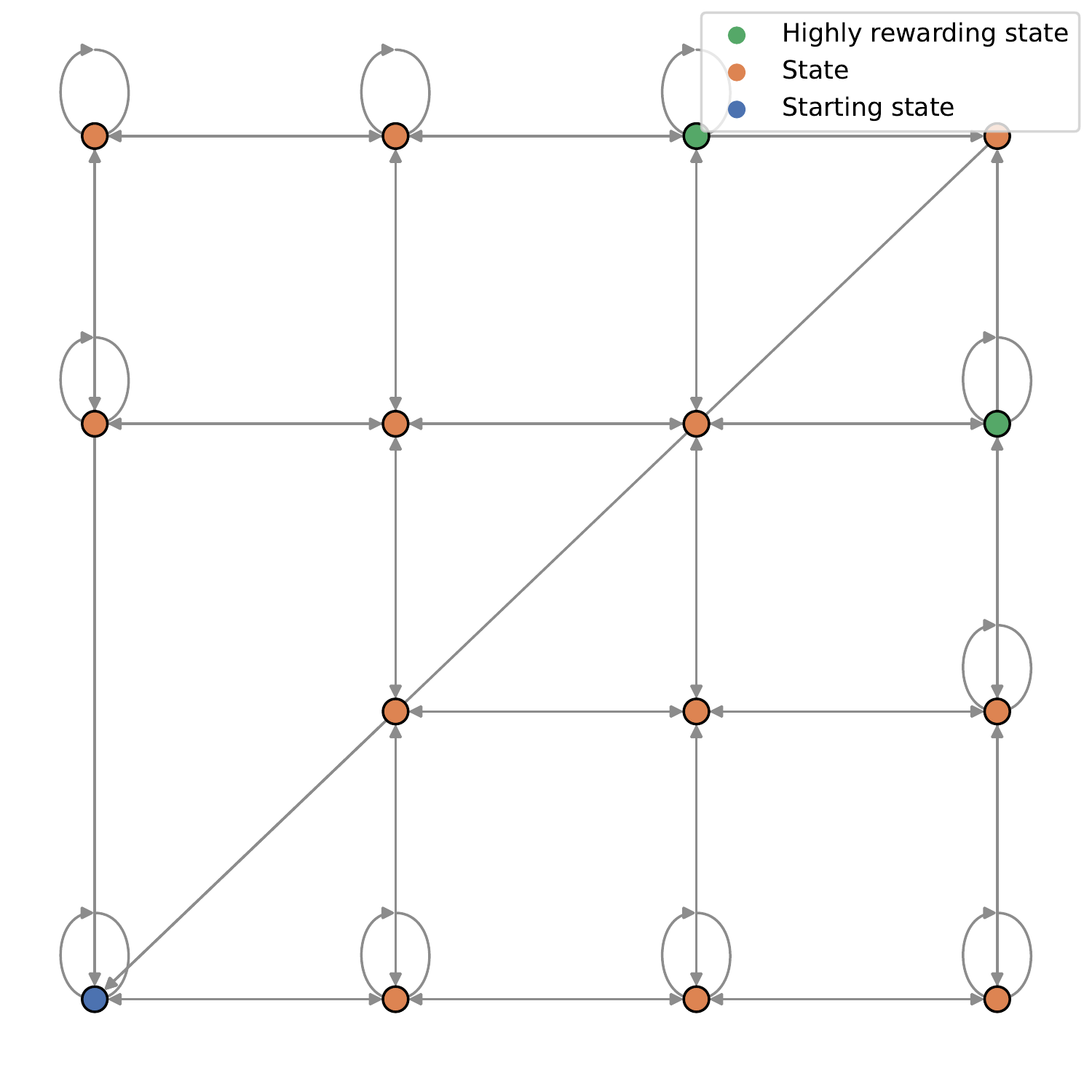}}
    \caption{\texttt{FrozenLake} MDP with size five.}
    \label{fig:frozenlake_mdp}
\end{figure}

The \texttt{FrozenLake} MDP is a grid world where the agent has to walk over a frozen lake to reach a highly rewarding state.
Some tiles of the grid are walkable, whereas others represent holes in which the agent may fall (leading back to the starting point).
The agent can move in the cardinal directions. However, movement on the walkable tiles is not entirely deterministic, and so the agent risks falling into holes if it walks too close to them.
The agent receives a small reward at each time step and zero reward when it falls into a hole.
The starting position of the agent is the bottom left state, which is the one farthest away from the goal position.
The challenge presented by \texttt{FrozenLake} lies in the high stochasticity of the movement.
A successful agent has to learn to balance the risk of falling into holes with reaching the goal quickly.
The structure of the grid is evident in the Markov chain representation but not in the MDP representation, as shown in Fig.~\ref{fig:frozenlake_mdp}.
In the textual representation for the \texttt{FrozenLake} MDP, the letter \texttt{A} corresponds to the position of the agent, the letter \texttt{F} represents a frozen tile over which the agent can safely walk, the letter \texttt{H} stands for a hole, and the letter \texttt{G} represents the goal.

\begin{table}[htb]
\captionsetup{position=top}
\centering%
\caption{FrozenLake emission map examples for a given state.}%
\begin{minipage}[c]{.24\textwidth}
\begin{center}
\resizebox{\textwidth}{!}{%
\begin{tabular}[b]{ccccc}
     'F' & 'F' & 'F' & 'H' & 'G'\\
     'F' & 'H' & 'H' & 'F' & 'H'\\
     'F' & 'H' & 'F' & 'H' & 'F'\\
     'F' & 'F' & 'F' & 'F' & 'F'\\
     'A' & 'F' & 'F' & 'F' & 'H'\\
\end{tabular}}
\end{center}
\end{minipage}%
\hfill%
\begin{minipage}[c]{.24\textwidth}
\begin{center}
$$
\begin{bmatrix}
     0. \\ 0.
\end{bmatrix}
$$
\end{center}
\end{minipage}%
\hfill%
\begin{minipage}[c]{.24\textwidth}
\includegraphics[width=1\textwidth]{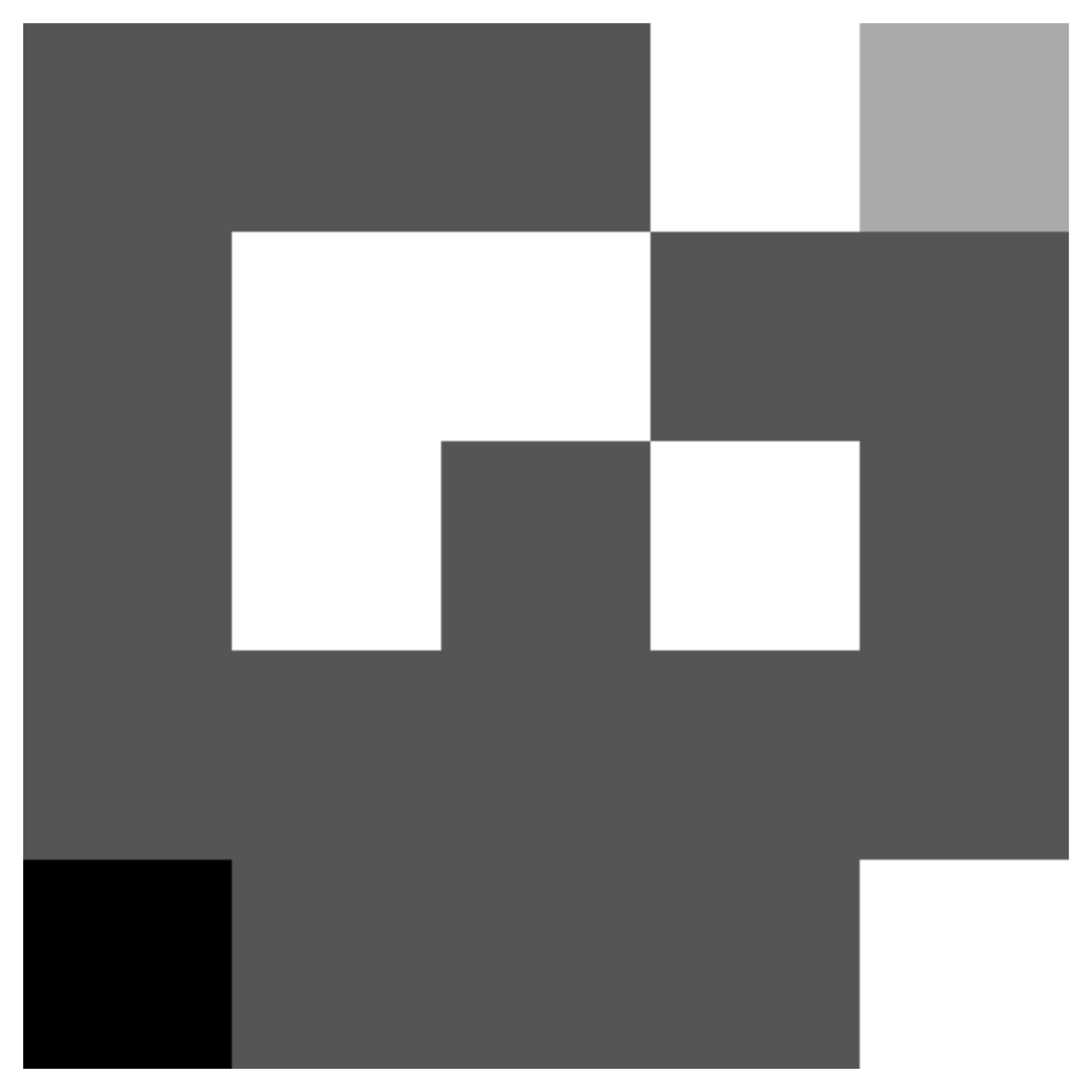}
\end{minipage}%
\hfill%
\begin{minipage}[c]{.24\textwidth}
\includegraphics[width=1\textwidth]{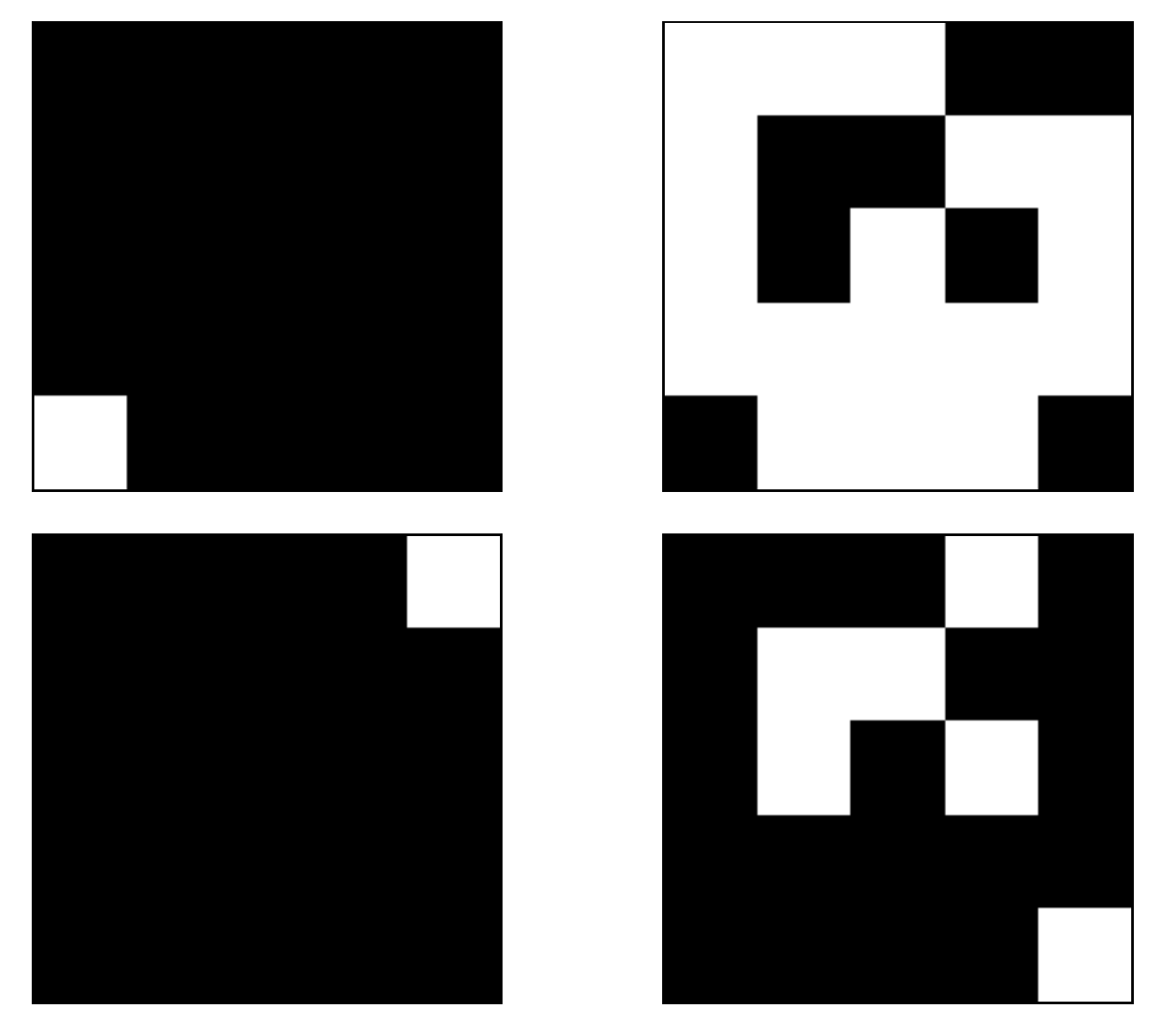}
\end{minipage}%
\vspace{0.1cm}
\begin{minipage}[c]{.03\textwidth}
(a)
\end{minipage}%
\begin{minipage}[c]{.21\textwidth}
\begin{center}
Textual state representation.
\end{center}
\end{minipage}%
\hfill%
\begin{minipage}[r]{.03\textwidth}
(b)
\end{minipage}%
\begin{minipage}[c]{.18\textwidth}
\begin{center}
State information emission map.
\end{center}
\end{minipage}%
\hfill%
\begin{minipage}[c]{.03\textwidth}
(c)
\end{minipage}%
\begin{minipage}[c]{.18\textwidth}
\begin{center}
Image encoding emission map.
\end{center}
\end{minipage}%
\hfill%
\begin{minipage}[c]{.03\textwidth}
(d)
\end{minipage}%
\begin{minipage}[c]{.18\textwidth}
\begin{center}
Tensor encoding emission map.
\end{center}
\end{minipage}%
\hfill%
\label{tab:fl_emissionmap}%
\end{table}%

\subsubsection{\texttt{Taxi}}

\begin{figure}[!ht]
    \centering
    \subfloat[Continuous MDP representation.
    \label{fig:taxi_cont}]{\includegraphics[width=0.4\textwidth]{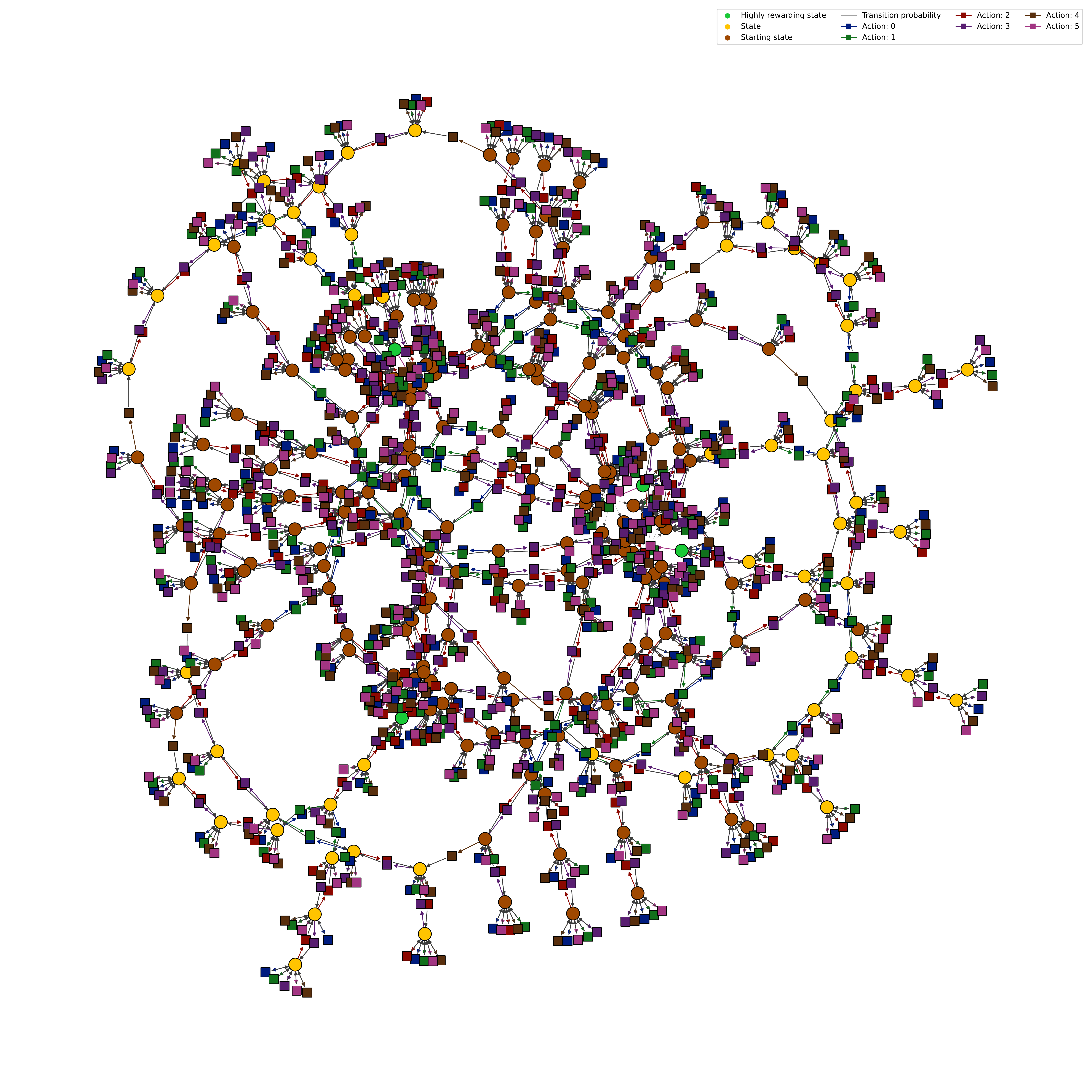}}
    \hfill
    \subfloat[Markov chain representation.
    \label{fig:taxi_mc}]{\includegraphics[width=0.4\textwidth]{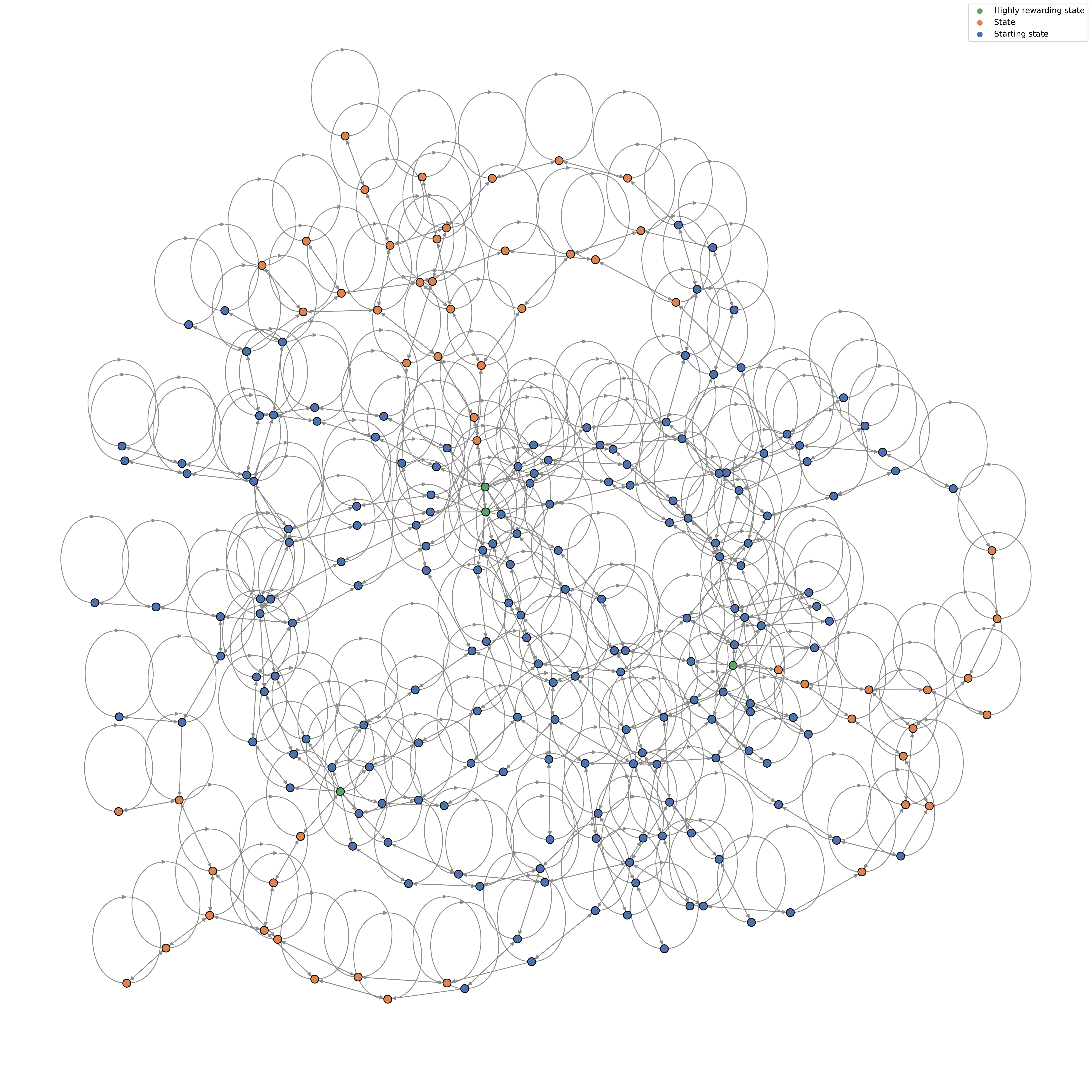}}
    \caption{\texttt{Taxi} MDP with size four.}
    \label{fig:taxi_mdp}
\end{figure}

The \texttt{Taxi} MDP is a grid world where the agent has to pick up and drop off passengers \citep{dietterich2000hierarchical}.
Each time a passenger is taken to the correct location a new passenger and destination appear.
The agent has six actions available, which correspond to the cardinal directions, picking a passenger, and dropping off a passenger.
The agent receives a large reward when it drops off a passenger at the correct destination and zero reward when it tries to drop off a passenger at an incorrect destination.
At every other time step, it receives a small reward.
The transition and reward structures of the \texttt{Taxi} MDP are particularly challenging to visualize due to the complexity of the task, as can be seen in Fig.~\ref{fig:taxi_mdp}.
In the textual representation for the \texttt{Taxi} MDP, the letter \texttt{A} encodes the position of the agent, the letter \texttt{W} represents a wall, the letter \texttt{P} represents the position of the passenger, and the letter \texttt{D} represents the final destination.

\begin{table}[t]
\captionsetup{position=top}
\centering%
\caption{Taxi emission map examples for a given state.}%
\begin{minipage}[c]{.24\textwidth}
\begin{center}
\resizebox{\textwidth}{!}{%
\begin{tabular}[b]{cccc}
     'X' & ' ' & 'P' & 'X'\\
     'X' & ' ' & ' ' & 'X'\\
     ' ' & ' ' & ' ' & ' '\\
     ' ' & 'X' & 'X' & ' '\\
     'D' & 'X' & 'X' & 'A'\\
\end{tabular}
}
\end{center}
\end{minipage}%
\hfill%
\begin{minipage}[c]{.24\textwidth}
\begin{center}
$$
\begin{bmatrix}
     0.\\ 4.\\ 4.\\ 3.\\ 0.\\ 0.
\end{bmatrix}
$$
\end{center}
\end{minipage}%
\hfill%
\begin{minipage}[c]{.24\textwidth}
\includegraphics[width=1\textwidth]{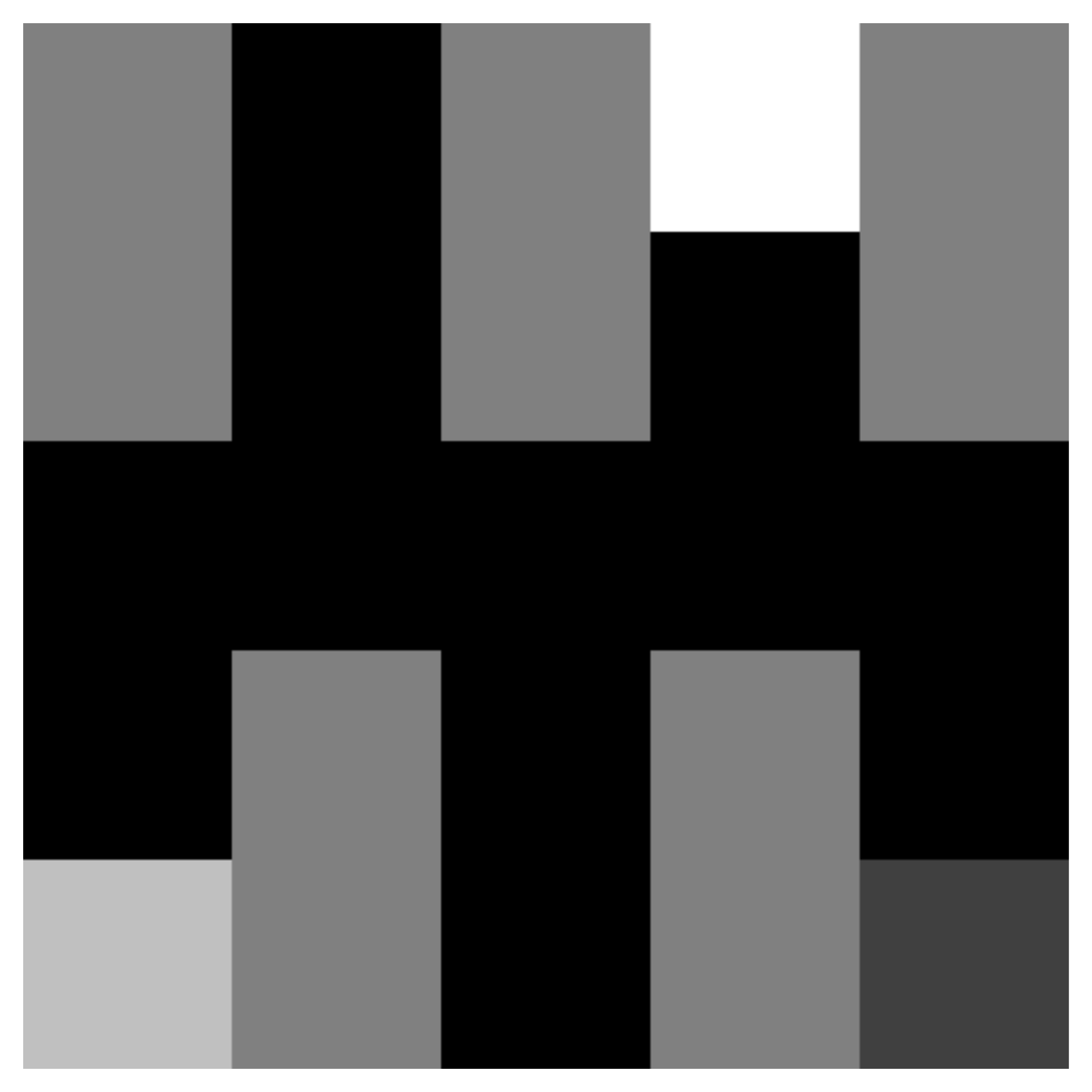}
\end{minipage}%
\hfill%
\begin{minipage}[c]{.24\textwidth}
\includegraphics[width=1\textwidth]{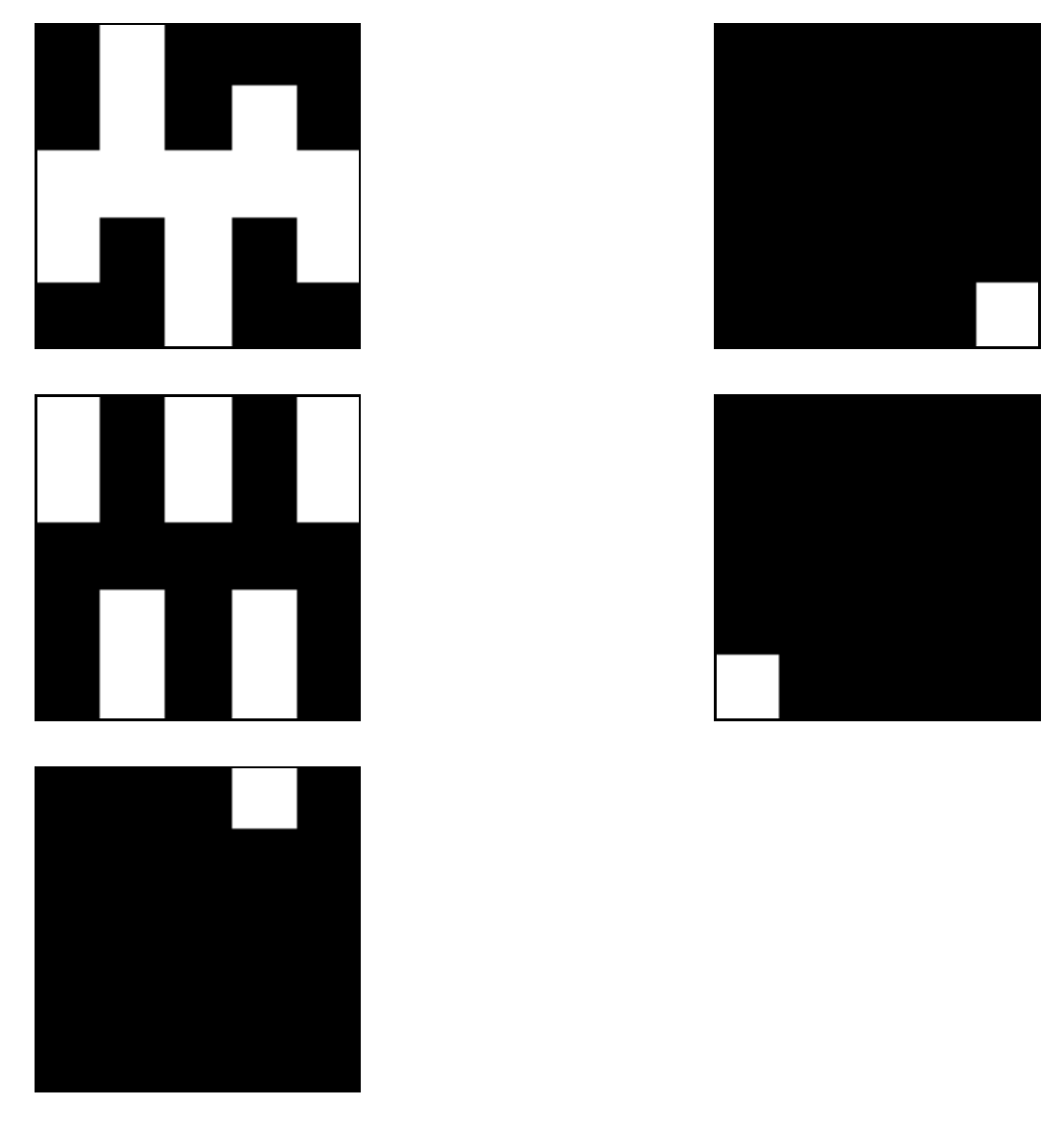}
\end{minipage}%
\vspace{0.1cm}
\begin{minipage}[c]{.03\textwidth}
(a)
\end{minipage}%
\begin{minipage}[c]{.21\textwidth}
\begin{center}
Textual state representation.
\end{center}
\end{minipage}%
\hfill%
\begin{minipage}[r]{.03\textwidth}
(b)
\end{minipage}%
\begin{minipage}[c]{.18\textwidth}
\begin{center}
State information emission map.
\end{center}
\end{minipage}%
\hfill%
\begin{minipage}[c]{.03\textwidth}
(c)
\end{minipage}%
\begin{minipage}[c]{.18\textwidth}
\begin{center}
Image encoding emission map.
\end{center}
\end{minipage}%
\hfill%
\begin{minipage}[c]{.03\textwidth}
(d)
\end{minipage}%
\begin{minipage}[c]{.18\textwidth}
\begin{center}
Tensor encoding emission map.
\end{center}
\end{minipage}%
\hfill%
\label{tab:t_emissionmap}%
\end{table}%

\newpage

\subsection{Diameter calculation} \label{app:diameter}

Recall that the diameter $D$ is defined as
$$
        D := \sup\limits_{s_1 \neq s_2} \inf_\pi T^\pi_{s_1 \to s_2}.
$$
From this definition, it is evident that the main challenge to computing the diameter is the infimum over policy space, which grows exponentially with the number of states and actions.

For Markov chains, the expected time to arrival $T_{s \to e}$ from state $s$ to state $e$ can be expressed recursively as
$$
    T_{s \to e} = \text{P}(e\mid s) + \sum_{s'} \text{P}(s'\mid s) (1 + T_{s' \to e}).
$$
In the case of an MDP M, we may derive a Bellman optimality equation for a policy that minimizes the expected arrival time to a given state. Concretely, the expected time to arrival $T^{*}_{s \to e}$ when following the best policy to transition from state $s$ to state $e$ is given by
\begin{equation} \label{eq:diam}
        T^{*}_{s \to e} = \min_{a \in \mathcal{A}} \left(P(e\mid s, a) + \sum_{s'} P(s'\mid s, a) (1 + T^{*}_{s' \to e})\right).
\end{equation}
In other words, consider an MDP $\text{M}'_e$ that is identical to M except for the reward kernel, which is given by $R(s,a) = -\mathbbm{1}(s \neq e)$. By calculating the optimal value function for $\text{M}'_e$ and taking the complement of its minimum across states, we may find $\max_s T^*_{s \to e}$. The diameter is obtained by computing this quantity for each state $e$ and selecting the maximum. This operation can be implemented efficiently through parallelization.
Note that, in order to preserve the exact value of the diameter, we do not include a discount factor in Equation~\ref{eq:diam}. As such, the resulting procedure is not yet guaranteed to converge. 
In practice, this does not represent an issue. We believe that a formal proof of convergence for the Bellman update corresponding to Equation~\ref{eq:diam} is possible.

\clearpage

\section{Measures of hardness computational complexity} \label{app:cc}

Previous literature has never been concerned with the computational complexity of hardness measures as they were not designed to be practically computed.
For this reason, there are no available efficient algorithms to compute most measures of hardness.
In the following, we provide a discussion of the challenges in the computational complexity of the measures of hardness from Section~\ref{sec:theory}.

\subsection{Mixing time}
Recall that the mixing time of an MDP is defined as the maximum of the mixing time of the Markov chain yielded by a policy. Considering that the development of the efficient computation of the mixing time of a Markov chain is in itself an active area of research \citep{pmlrv99wolfer19a} and that the policy space grows exponentially in the cardinality of the state and action spaces, it becomes clear that the development of an efficient algorithm to compute the mixing time is extremely challenging, if not impossible.

\subsection{Diameter}
Recall that the diameter is defined as the expected number of time steps required to transition between two states when following the best policy for doing so.
Similarly to the mixing time, computing the diameter involves a minimization in policy space.
However, we were able to remove the dependency on the exponentially growing policy space by noting that the diameter of an MDP M can be related to solving $|S|$ MDPs that are closely related to M with a discount factor of $1$ (App.~\ref{app:diameter}).
\citet{lee2013path} propose a linear program to solve the non-linear Bellman equation that has $\tilde O(|S|^{2.5}|A|)$ computational complexity and no $\gamma$ dependency.
Since computing the diameter requires solving $|S|$ MDPs, the resulting computational complexity is $\tilde O(|S|^{3.5}|A|)$.

\subsection{Distribution mismatch coefficient}
Recall the the computation of the distribution mismatch coefficient involves a maximization in policy space and the stationary distribution.
Although the stationary distribution of a Markov chain can be efficiently computed with several algorithms including linear programming, an algorithm that allows to remove the dependency on the exponentially growing policy space is not currently known for this measure of hardness.

\subsection{Action-gap regularity}
Since the action-gap regularity is defined as a constant in the upper bound of an integral, a closed form equation for this measure of hardness is not available.
Further, it has not been extended to MDPs with more than two actions.

\subsection{Environmental value norm}
When the environmental value norm is to be interpreted as a measure of hardness for an MDP, \citet{maillard2014hard} suggest to use the environmental value norm of the optimal policy, which can be efficiently computed as the norm of the optimal state value function with respect to the transition probabilities of the Markov chain yielded by the optimal policy.
Since the norm is computed as a simple matrix product, the dominating term in the computational complexity is the calculation of the optimal state value function.
See Table 2 in \citet{sidford2018near} for the computational complexity of several available algorithms.
In Table~\ref{tab:hmcc}, we report the computational complexity of value iteration since it is the most widely known of such algorithms.

\subsection{Sum of the reciprocals of the sub-optimality gaps}
Similarly to the environmental value norm, the computational complexity of the \subgaps is dominated by the calculation of the optimal state value function.

\section{Normalization procedures} \label{app:reg_norm}

\paragraph{Measures of hardness.}
The normalization of the theoretical measures of hardness (diameter, environmental value norm, and the \subgaps) is carried out by scaling the values to the range $[0,1]$ given their maximum and minimum values.
For example, in the empirical investigation of the measures of hardness (see Sec.~\ref{sec:analysis}), the maximum and minimum values are taken across all the different seeds and parameters.
For the cumulative regret of the near-optimal agent (when used as an optimistic measure of hardness), we leverage the per-step normalization procedure that is described in the next paragraph.
This procedure takes into account the bounded range of the optimal return of an MDP.

\paragraph{Expected cumulative regret.}
In the episodic case, the per-step normalized version of the episodic regret is obtained by dividing the regret by the difference between the value of the optimal policy and the value of the policy with the least value. For a policy $\pi$, this is given by 
\begin{equation*}
    \frac{V_{0, \texttt{epi}}^*(s_0) - V_{0, \texttt{epi}}^\pi(s_0)}{V_{0, \texttt{epi}}^*(s_0) - V_{0, \texttt{epi}}^{-}(s_0)} \in [0, 1],
\end{equation*}
where $\pi^- = \argmin\limits_{\pi \in \Pi} V_{0, \texttt{epi}}^\pi(s_0)$.
In the continuous case, the per-step normalized average instantaneous regret is obtained similarly as
\begin{equation*}
    \frac{\rho^* - \frac{1}{t} \sum_{k=0}^t r_t}{\rho^* - \rho^-} \in [0,1],
\end{equation*}
where $r_t$ is the reward obtained by the policy at time step $t$ and $\rho^- = \min\limits_{\pi \in \Pi} \rho^\pi$.

\clearpage
\section{Empirical investigation of hardness measures } \label{app:analysis}

\begin{figure}[htb]
    \centering
    \includegraphics[width=0.75\textwidth]{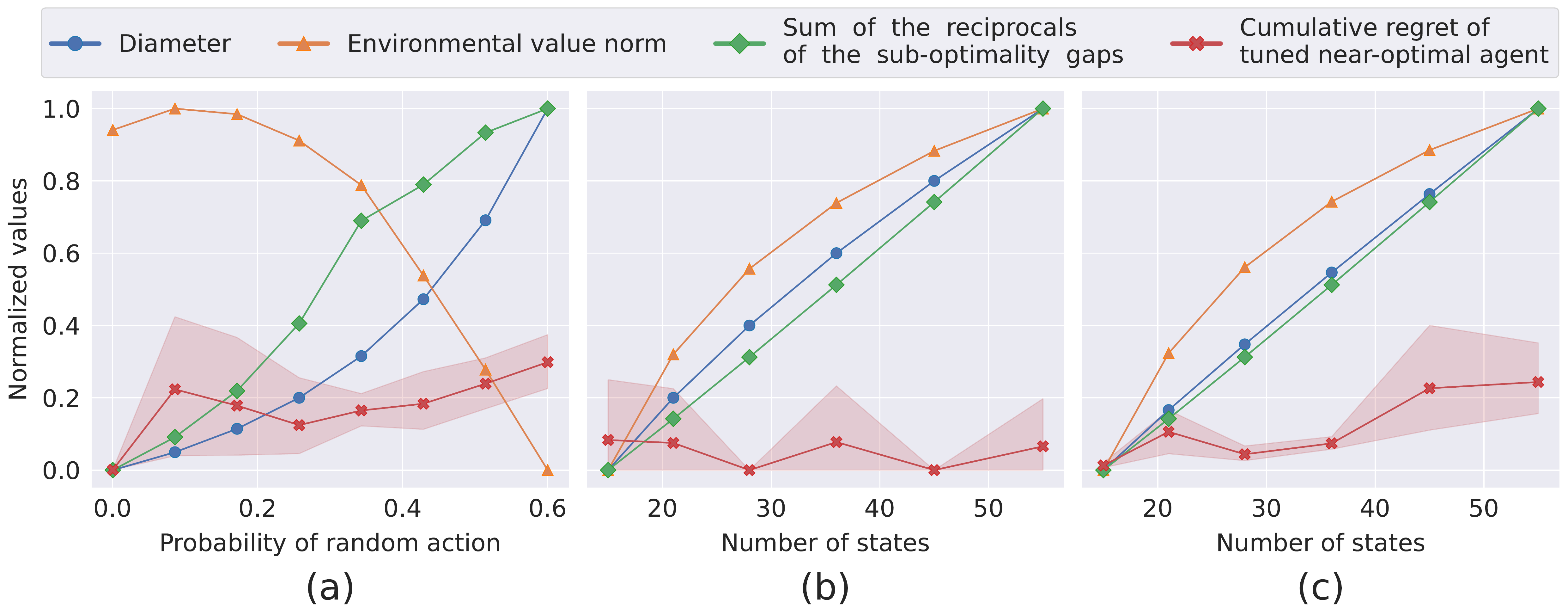}
    \includegraphics[width=0.75\textwidth]{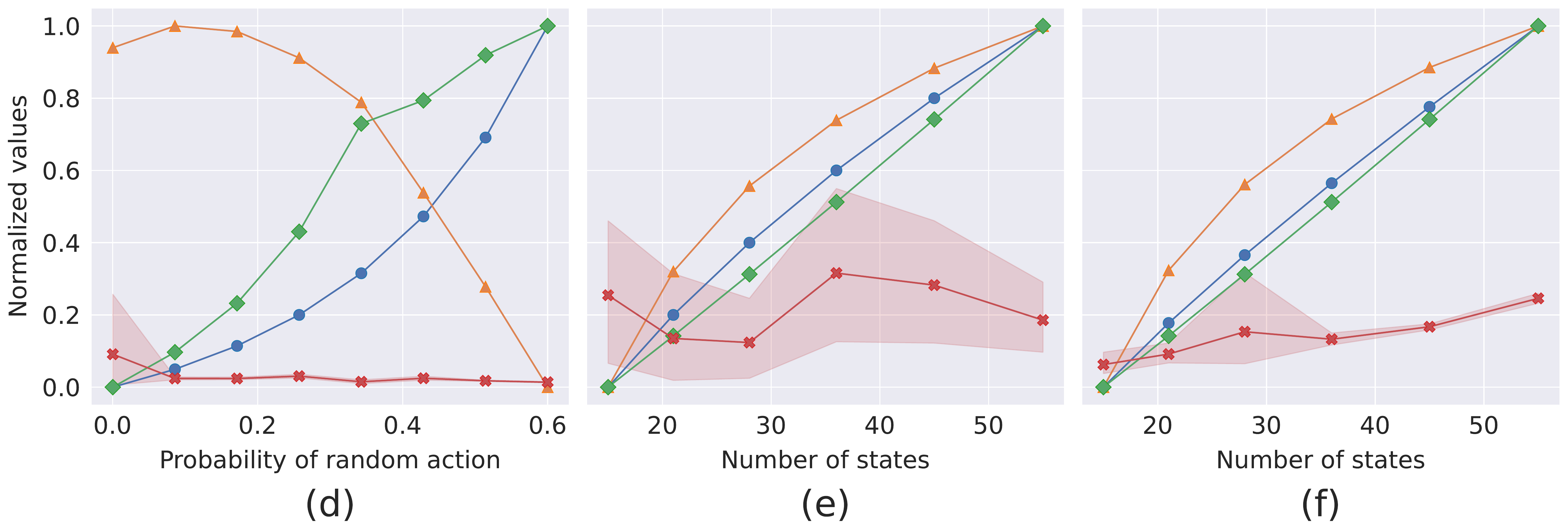}
    \caption{\texttt{DeepSea} results in the episodic (top) and continuous (bottom) settings, scenarios 1, 3, and 4 correspond to the columns from left to right}
    \label{fig:hardness_analysis_DeepSea_app}
\end{figure}

In our empirical investigation of the measures of hardness, we consider five MDP families (\texttt{MG-Empty}, \texttt{SimpleGrid}, \texttt{FrozenLake}, \texttt{RiverSwim}, and \texttt{DeepSea}) that include different levels of stochasticity and challenge.
Each MDP family is tested in four scenarios that highlight different aspects of hardness.
Note that each measure has been normalized (as described in App.~\ref{app:reg_norm}), which \emph{solely} allows comparing trends (growth rates).
Figures \ref{fig:hardness_analysis_MiniGridEmpty_app}, \ref{fig:hardness_analysis_SimpleGrid_app}, \ref{fig:hardness_analysis_FrozenLake_app}, \ref{fig:hardness_analysis_RiverSwim_app} and \ref{fig:hardness_analysis_DeepSea_app} (pg. \pageref{fig:hardness_analysis_MiniGridEmpty_app}) report the results of our investigation along with the 95\% bootstrapped confidence intervals over twelve seeds.

\paragraph{Scenario 1.} We vary the probability \prand that an MDP executes a random action instead of the action selected by an agent.
As \prand increases, estimating the optimal value function becomes easier since every policy yields increasingly similar value functions.
This produces a decrease in the estimation complexity.
However, intentionally visiting states becomes harder.

\paragraph{Scenario 2.} We vary the probability \plazy that an MDP stays in the same state instead of executing the action selected by an agent. Contrary to increasing \prand, increasing \plazy never benefits exploration through the execution of random actions.
Increasing \plazy decreases estimation complexity and increases visitation complexity.

\paragraph{Scenario 3 and 4.}  We vary the number of states across MDPs from the same family. In scenario 4, we also let \prand $= 0.1$ to study the impact of stochasticity. In these scenarios, increments in the number of states increase both estimation complexity and visitation complexity.

\paragraph{Cumulative regret of a \emph{near-optimal} agent.}
In every scenario, the measures of hardness are compared with the cumulative regret of a \emph{near-optimal} agent that serves as an optimistic approximation of a complete measure of hardness.
The near-optimal agents have been chosen between the ones available in \colosseum with the lowest average cumulative regret in the benchmarking results (such as PSRL in the episodic setting and UCRL2 in the continuous setting).
In order to optimistically approximate a complete measure of hardness, we tune the hyperparameters of the agents for each MDP in every scenario.
Concretely, we perform a random search with the objective of minimizing the average cumulative regret resulting from an interaction of the agent with the MDP that lasts for $200\,000$ time steps with a maximum time limit of two minutes across three seeds.
The budget for the random search is $120$ samples.

\paragraph{Computational power.}
The empirical investigation has been carried out on a desktop PC equipped with an \textit{AMD Ryzen 9 5950X 16-Core Processor} and required less than $24$ hours for all the MDP families and scenarios.
The most computationally intensive part of the procedure is the hyperparameter search.

\paragraph{Limitations.}
The main limitation of our empirical investigation is the selection of the MDP families, near-optimal agents, and scenarios.
Although we believe to have proposed a solid methodology, we are open to discussing the inclusion of additional experiments to further enhance \colosseum.

\subsection{Analysis of results}

\begin{figure}[htb]
    \centering
    \includegraphics[width=1\textwidth]{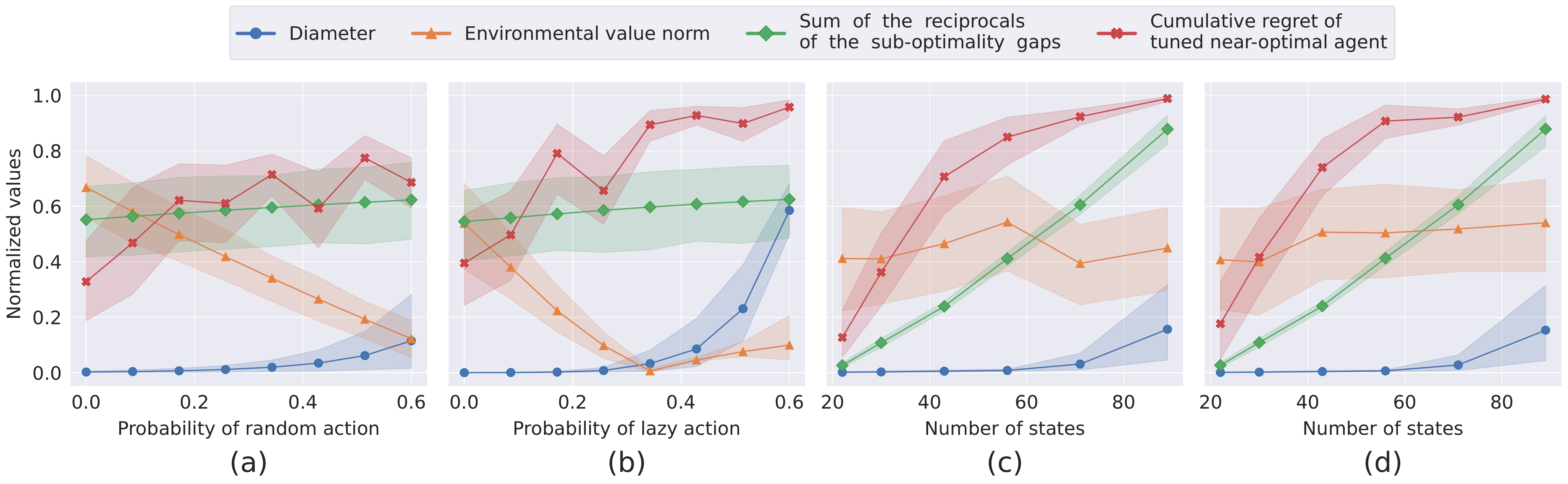}
    \includegraphics[width=1\textwidth]{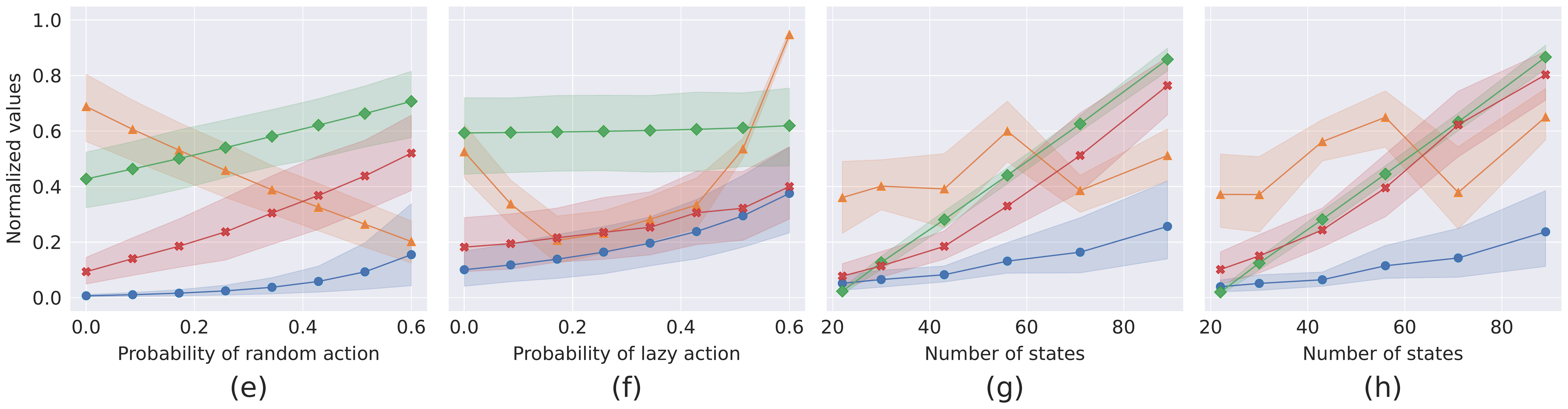}
    \caption{\texttt{FrozenLake} results in the episodic (top) and continuous (bottom) settings, scenarios 1-4 correspond to the columns from left to right}
    \label{fig:hardness_analysis_FrozenLake_app}
\end{figure}

\paragraph{Diameter.} The diameter grows superlinearly with both \prand and \plazy since deliberate movement between states requires an exponentially increasing number of time steps.
As clearly shown in the figures, this phenomenon is exacerbated in the episodic setting.
If the agent is forced to take a random action or to stay in the same state, it can miss the opportunity to reach the target state in the current episode and has to try again in the next episode.
Although the diameter highlights this sharply increasing visitation complexity, its trend overestimates the increase in cumulative regret of the tuned near-optimal agent, which is explained by the unaccounted decrease in estimation complexity. The diameter also increases almost linearly with the number of states. When \prand is relatively small, an approximately linear relationship can still be observed.
This linear trend underestimates the non-linear growth in hardness clearly shown in the cumulative regret of the tuned near-optimal agent but is in line with the mild increase in visitation complexity.
\texttt{FrozenLake} in the episodic setting (Figures \ref{fig:hardness_analysis_FrozenLake_app}c and \ref{fig:hardness_analysis_FrozenLake_app}d) represents the only exception. Given the extremely high level of stochasticity of the MDP, increasing the number of states drastically increases the visitation complexity while making it easier for the agent to act near-optimally.

\begin{figure}[htb]
    \centering
    \includegraphics[width=1\textwidth]{figures/hardness_analysis/hardness_analysis_MiniGridEmptyEpisodic.pdf}
    \includegraphics[width=1\textwidth]{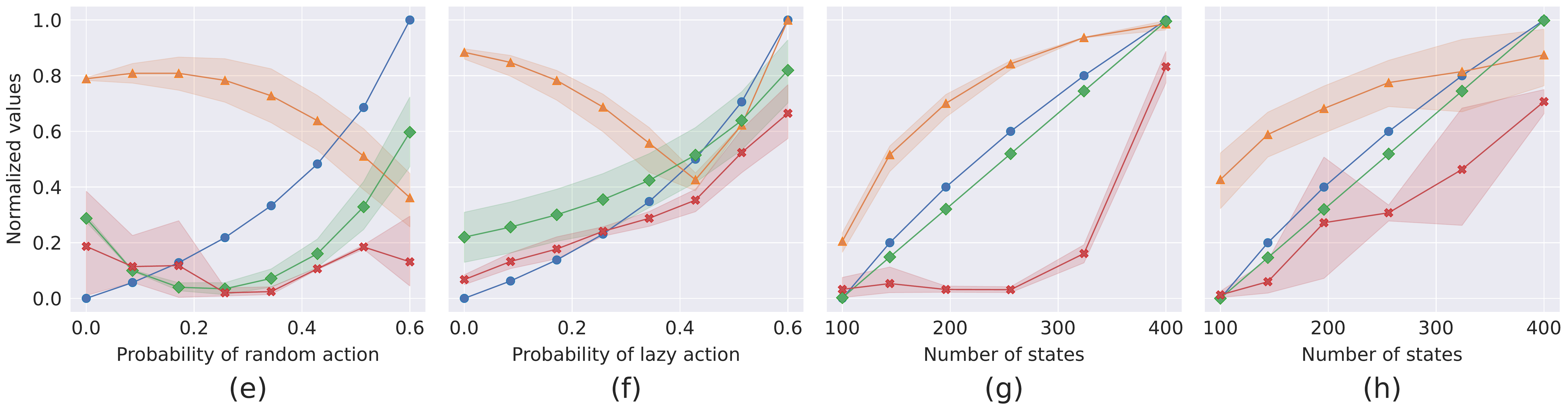}
    \caption{\texttt{MiniGridEmpty} results in the episodic (top) and continuous (bottom) settings, scenarios 1-4 correspond to the columns from left to right.}
    \label{fig:hardness_analysis_MiniGridEmpty_app}
\end{figure}

\paragraph{Environmental value norm.} The environmental value norm decreases as \plazy and \prand increase because the optimal value of neighboring states becomes closer, which decreases the per-step variability of the optimal state value function.
However, we note that for the \texttt{MG-Empty} and the \texttt{FrozenLake} MDP families in the continuous cases (see Figures \ref{fig:hardness_analysis_MiniGridEmpty_app}f and \ref{fig:hardness_analysis_FrozenLake_app}f)
as \plazy increases, the environmental value norm first decreases and later increases.
From a certain value of \plazy onward, there is a significant probability of the agent remaining in the same state.
This provokes large changes in the value of states that are distant from the highly rewarding states and no changes at all for highly rewarding states since the \textit{lazy} transition is comparable to taking the optimal action.
Due to the large changes in the suboptimal region of the state space and the absence of changes in the optimal region of the state space, the overall one-step variability of the state value function increases.
Note that this does not happen in the episodic case due to the restarting mechanism and whether it happens or not in the continuous case depends on the transition and reward structure of an MDP.
When the number of states increases but the transition and reward structures remain the same, the small increase in measured variability only causes the environmental value norm to grow sublinearly. These findings are strong evidence that this measure is only suited to capture estimation complexity.

\paragraph{Sum of the reciprocals of the sub-optimality gaps.} The \subgaps increases weakly superlinearly in scenarios 1 and 2. The probability of executing the action selected by the agent decreases when \plazy and \prand increase, and so the difference between the optimal value function and the optimal state-action value function decreases sharply.
\texttt{FrozenLake} (Figures \ref{fig:hardness_analysis_FrozenLake_app}a, \ref{fig:hardness_analysis_FrozenLake_app}b, \ref{fig:hardness_analysis_FrozenLake_app}e and \ref{fig:hardness_analysis_FrozenLake_app}f) represents an exception as the \subgaps is almost constant.
\texttt{FrozenLake} naturally incorporates an exceptionally high level of stochasticity and so varying \plazy and \prand does not significantly affect the value functions.
The \subgaps increases almost linearly with the number of states. This is explained by the fact that the average value of the additional terms in the summation is often similar to the average value of the existing terms given the same structure of reward and transition kernels.
This measure of hardness is not particularly apt at capturing estimation complexity, since it focuses solely on optimal policy identification. It also underestimates the increase in hardness induced by an increase in visitation complexity.

\begin{figure}[htb]
    \centering
    \includegraphics[width=1\textwidth]{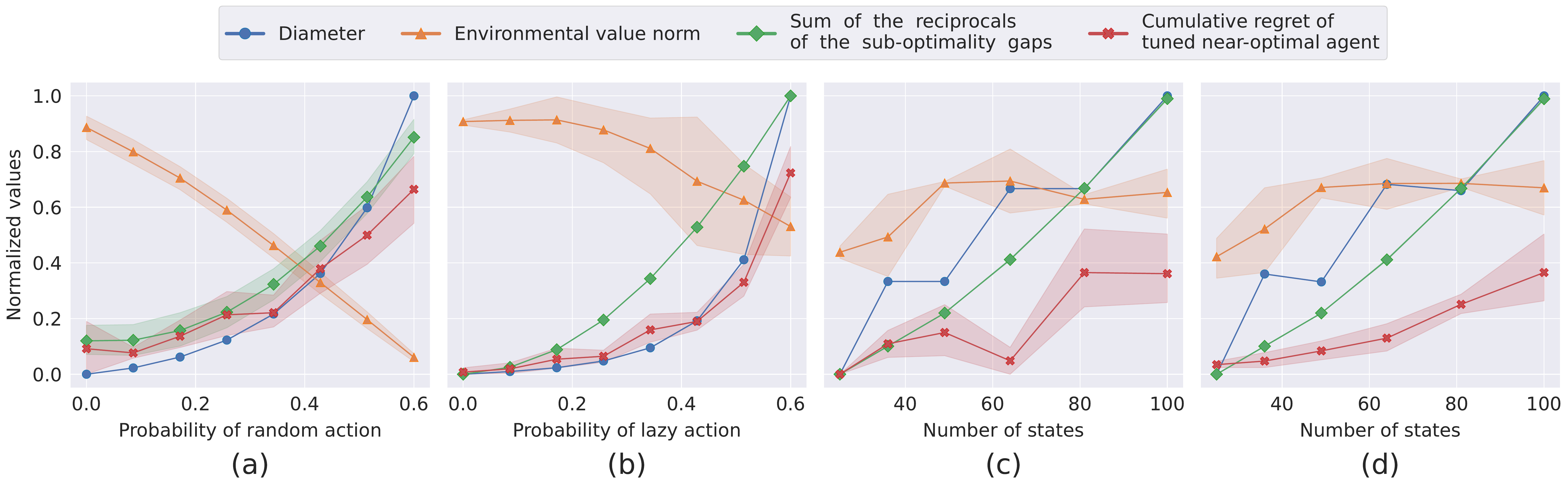}
    \includegraphics[width=1\textwidth]{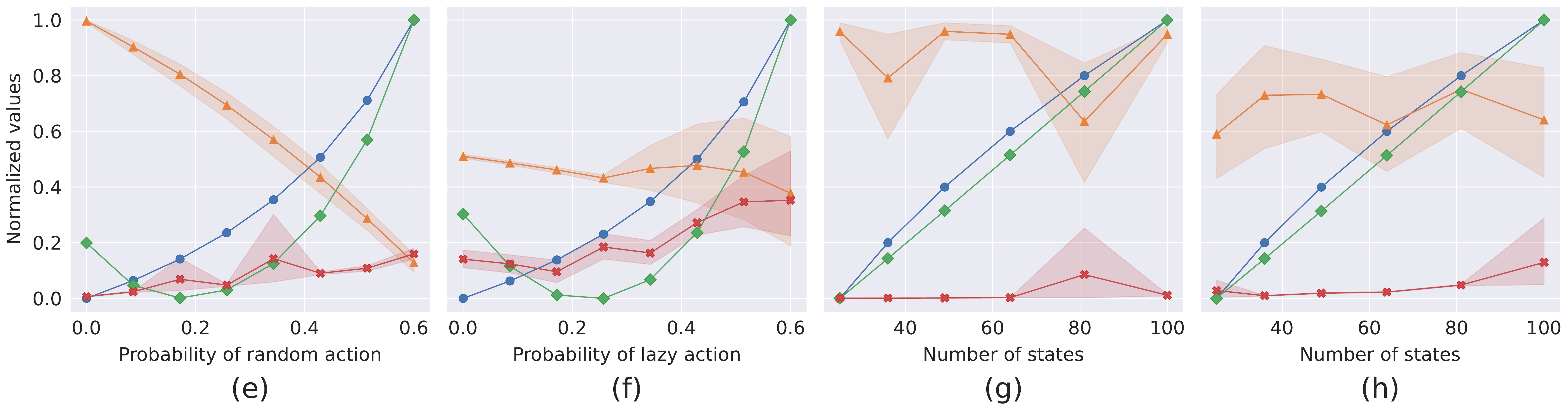}
    \caption{\texttt{SimpleGrid} results in the episodic (top) and continuous (bottom) settings, scenarios 1-4 correspond to the columns from left to right}
    \label{fig:hardness_analysis_SimpleGrid_app}
\end{figure}

\begin{figure}[htb]
    \centering
    \includegraphics[width=1\textwidth]{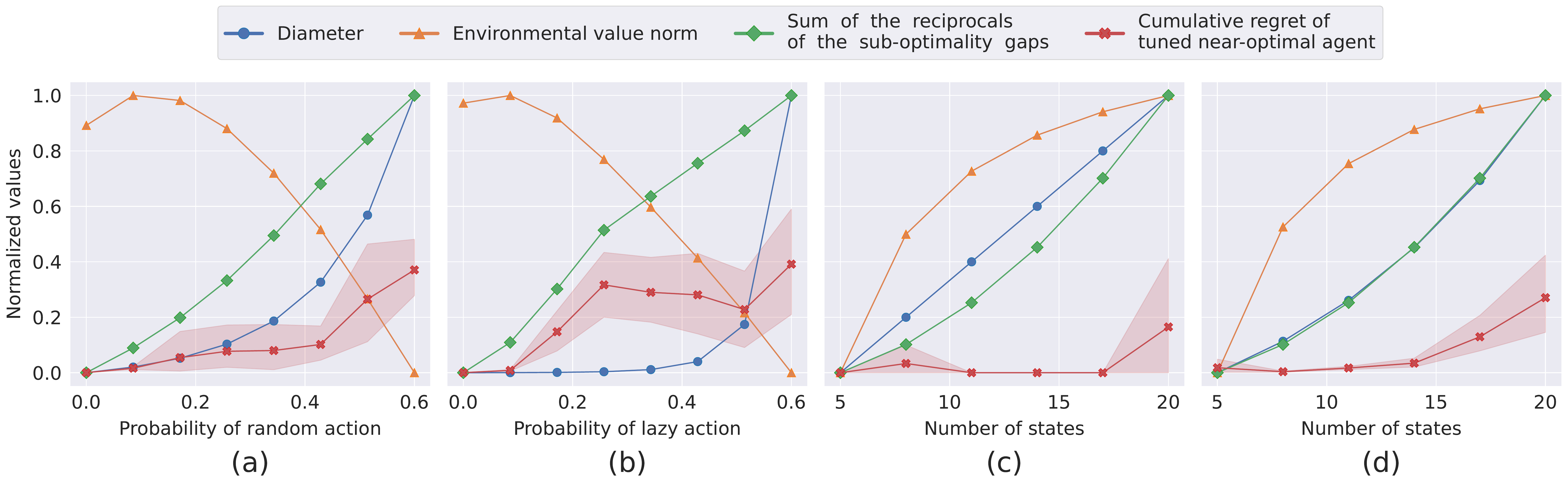}
    \includegraphics[width=1\textwidth]{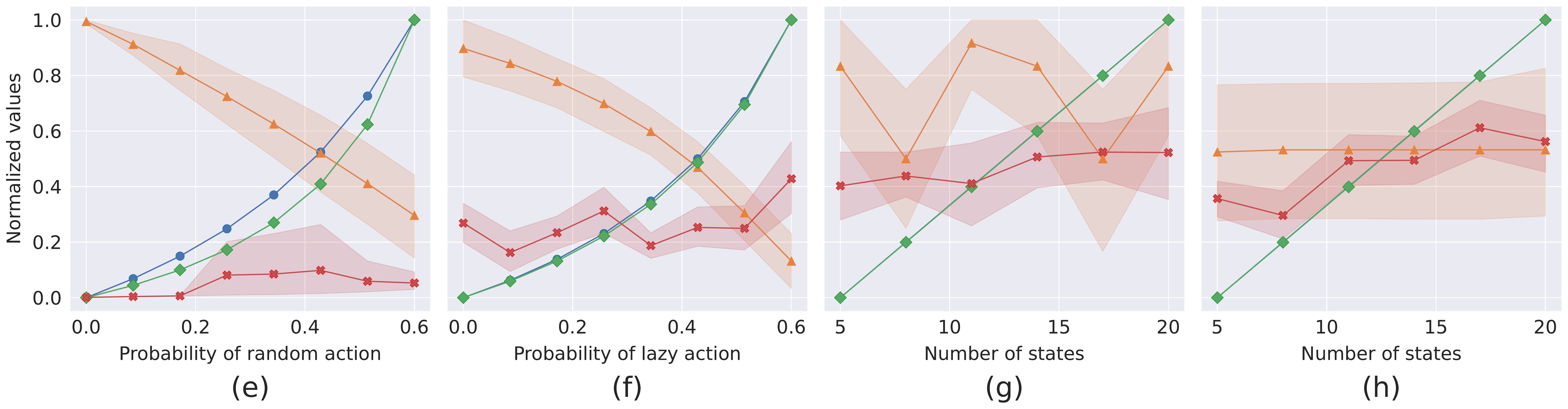}
    \caption{\texttt{RiverSwim} results in the episodic (top) and continuous (bottom) settings, scenarios 1-4 correspond to the columns from left to right}
    \label{fig:hardness_analysis_RiverSwim_app}
\end{figure}

\paragraph{Cumulative regret of the tuned near-optimal agent.}
The trends of the cumulative regret of the tuned near-optimal agent present more variability when compared to the theoretical measures of hardness.
This reflects the fact that this is an approximation of a complete measure of hardness based on agents that have specific strengths and weaknesses.
Overall, we note a tendency of superlinear growth in scenarios 1 and 2.
Such tendency is specifically marked for the grid worlds, such as \texttt{MG-Empty} (see Fig.~\ref{fig:hardness_analysis_MiniGridEmpty_app}) and \texttt{SimpleGrid} (see Fig.~\ref{fig:hardness_analysis_SimpleGrid_app}).
In these MDP families, the highly rewarding states are located far from the starting states and therefore the visitation complexity plays a fundamental role.
In the \texttt{FrokenLake} family (see Fig.~\ref{fig:hardness_analysis_FrozenLake_app}), the trend is linear (episodic setting) or sub-linear (continuous setting), which is caused by the relatively low impact of the parameters \prand and \plazy in the already highly stochastic MDPs.
The cumulative regret of the tuned near-optimal agent presents a moderately superlinear growth in the episodic case and remains almost constant for the \texttt{RiverSwim} family (see Fig.~\ref{fig:hardness_analysis_RiverSwim_app}).
This results from the fact that, in the continuous case, the absence of the restarting mechanism in combination with the chain structure of the MDP allows the agent to suffer only minimal impact from the increasing values of the parameters \prand and \plazy.
Finally, for the \texttt{DeepSea} family, the regret is constant.
Increases in \prand dramatically reduce the possibility of visiting the highly rewarding state due to the pyramid structure of the MDP.
In scenarios 3 and 4, the overall tendency is still superlinear but less marked compared to scenarios 1 and 2.
The superlinear growth is most evident for the \texttt{MG-Empty} family (see Fig.~\ref{fig:hardness_analysis_MiniGridEmpty_app}) and for the \texttt{RiverSwim} family in the episodic case (see Figures~\ref{fig:hardness_analysis_RiverSwim_app}c and \ref{fig:hardness_analysis_RiverSwim_app}d).
For the \texttt{MG-Empty}, when the grid size is increased, and with it the number of states, the MDP becomes increasingly challenging to navigate since the agent has to coordinate its rotation with its forward movement in order to effectively transition between states.
For the \texttt{RiverSwim} family, the challenge comes from the restarting mechanism on a chain structure. The agent is required to take a perfect sequence of actions in order to visit the last state of the chain, otherwise it will be reset to the start.
In the continuous setting (see Figures~\ref{fig:hardness_analysis_RiverSwim_app}g and \ref{fig:hardness_analysis_RiverSwim_app}h), instead, the trends are mostly linear, similarly to what happens in scenarios 1 and 2.
The less challenging structure of the \texttt{SimpleGrid} family (see Fig.~\ref{fig:hardness_analysis_SimpleGrid_app}) induces weakly superlinear trends of the cumulative regret of the tuned near-optimal agent.
We note that, contrary to the theoretical measures of hardness, the episodic setting does not appear to be harder (which would be suggested by steeper trends).
This discrepancy is particularly noticeable in the \texttt{FrozenLake} family (see Fig.~\ref{fig:hardness_analysis_FrozenLake_app}) which yields a mostly linear trend in the continuous settings and clearly sublinear trends in the episodic settings.
In the \texttt{DeepSea} family, the cumulative regret of the tuned near-optimal agent is almost constant in scenario 3 and almost linear in scenario 4.
The main challenge for this family lies in the pyramidal structure of the MDP rather than the number of states.
However, setting $\prand = 0.1$ creates a more challenging task for the agent as more time steps are required to find the highly rewarding state.
We also note that the difference in results between the episodic and continuous settings is minimal, which is unsurprising given the MDP structure.

\clearpage
\section{Benchmarking} \label{app:benchmark}

In the following sections, we provide details on the selection methodology for the environments in the benchmark, and we explain the full benchmarking procedure from the hyperparameters selection to the benchmark evaluation.

\subsection{Benchmark environments selection}
The environments in the benchmark have been selected to be as diverse as possible with respect to the diameter and the environmental value norm.
Based on the theoretical properties of these measures and the results of the empirical comparison in Section \ref{app:analysis}, 
we believe that they represent valid proxies for the visitation complexity and the estimation complexity.
The candidate environments have been sampled from a set of parameters such that their diameter is less than $100$ and the environmental value norm is less than $3.5$.
This guarantees a sufficient challenge for the \rl agents while limiting the scale of the environments.

\begin{figure}[h!]
    \centering
    \includegraphics[width=1\textwidth]{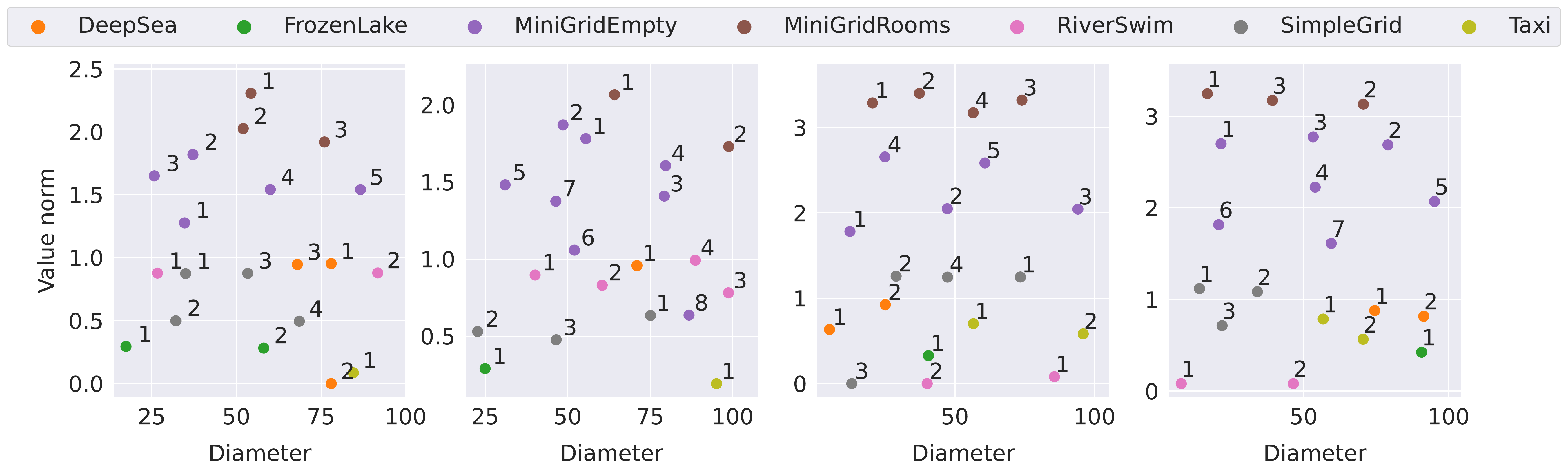}
    \caption{Positions in measure of hardness space of the set of MDPs in the benchmark.}
    \label{fig:benchmark}
\end{figure}

Figure~\ref{fig:benchmark} represents the benchmark MDPs placed according to their diameter and environmental value norm.
The selection features MDPs with varying combinations of values of diameter in the interval $[20, 100]$ and environmental value norm in the interval $[0, 3.5]$

\subsection{Hyperparameter selection}
The hyperparameter selection procedure is to be considered an integral component of the \colosseum benchmarking procedure to ensure fair hyperparameter tuning.

Each \colosseum agent is required to define a sampling space for all its parameters.
These sampling spaces are used by the package to conduct a random search optimization procedure with the objective of minimizing the cumulative regret across a set of randomly sampled environments.
The random sampling procedure for environments is defined for each \colosseum MDP family and aims to provide a varied set of MDPs of up to moderate scale.
Note that the \textit{MG-DoorKey} family is excluded from the hyperparameter selection as it is weakly communicating in the continuous case.
Tutorials on how to implement the aforementioned functions for novel agents and environments are available online.

The hyperparameters for the agents employed in the paper have been obtained with $50$ samples from the hyperparameter spaces, which have been evaluated on $12$ MDPs from each family, for a total of $84$ MDPs with a training time of $20$ minutes and a maximum number of total time steps of $200\ 000$.

\subsection{Computational resources}
The experiments for the benchmarking procedure have been carried out using CPUs from the Queen Mary University of London Apocrita HPC facility.
Note that, due to the time constraint imposed by \colosseum, the computational resources required to run the benchmark are bounded, and the benchmarking procedure is easily parallelizable.

\subsection{Tabular setting}

\paragraph{Experimental procedure.}
We set the total number of time steps to $500\ 000$ with a maximum training time of $10$ minutes for the tabular setting and $40$ minutes for the non-tabular case.
If an agent does not reach the maximum number of time steps before this time limit, learning is interrupted, and the agent keeps using its last best policy.
This guarantees a fair comparison between agents with different computational costs.
The performance indicators are computed every $100$ time steps.
Each interaction between an agent and an MDP is repeated for $20$ seeds.
The per-step normalized cumulative regret (defined in App.~\ref{app:reg_norm}) is employed as a performance measure since it provides a unified scale across different MDPs.

\begin{figure}[htb!]
    \captionsetup[subfloat]{position=bottom}
    \centering
    \subfloat[][Episodic communicating setting.]{
    \includegraphics[width=0.66\linewidth]{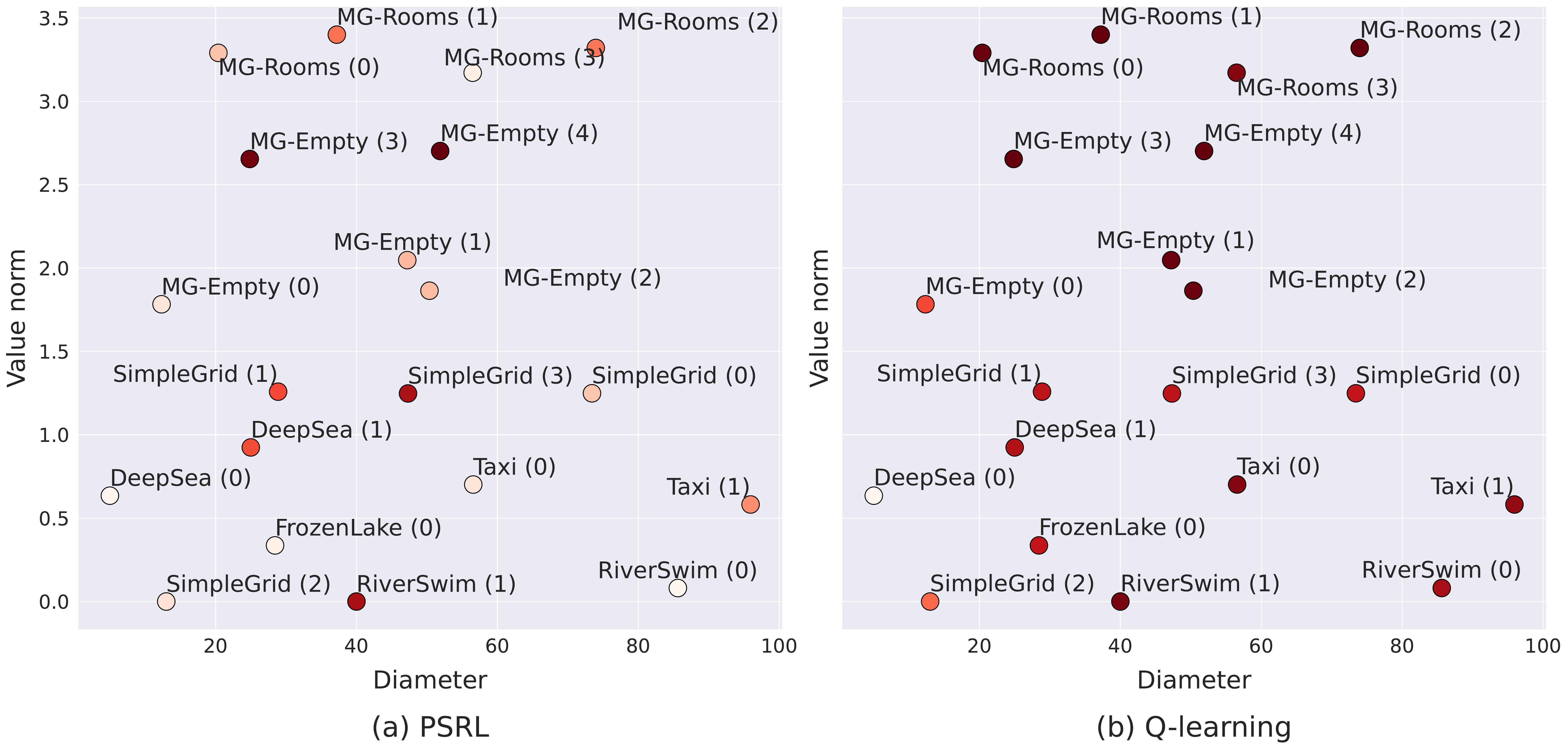}}\\
    \subfloat[][Episodic ergodic setting.]{
    \includegraphics[width=0.66\linewidth]{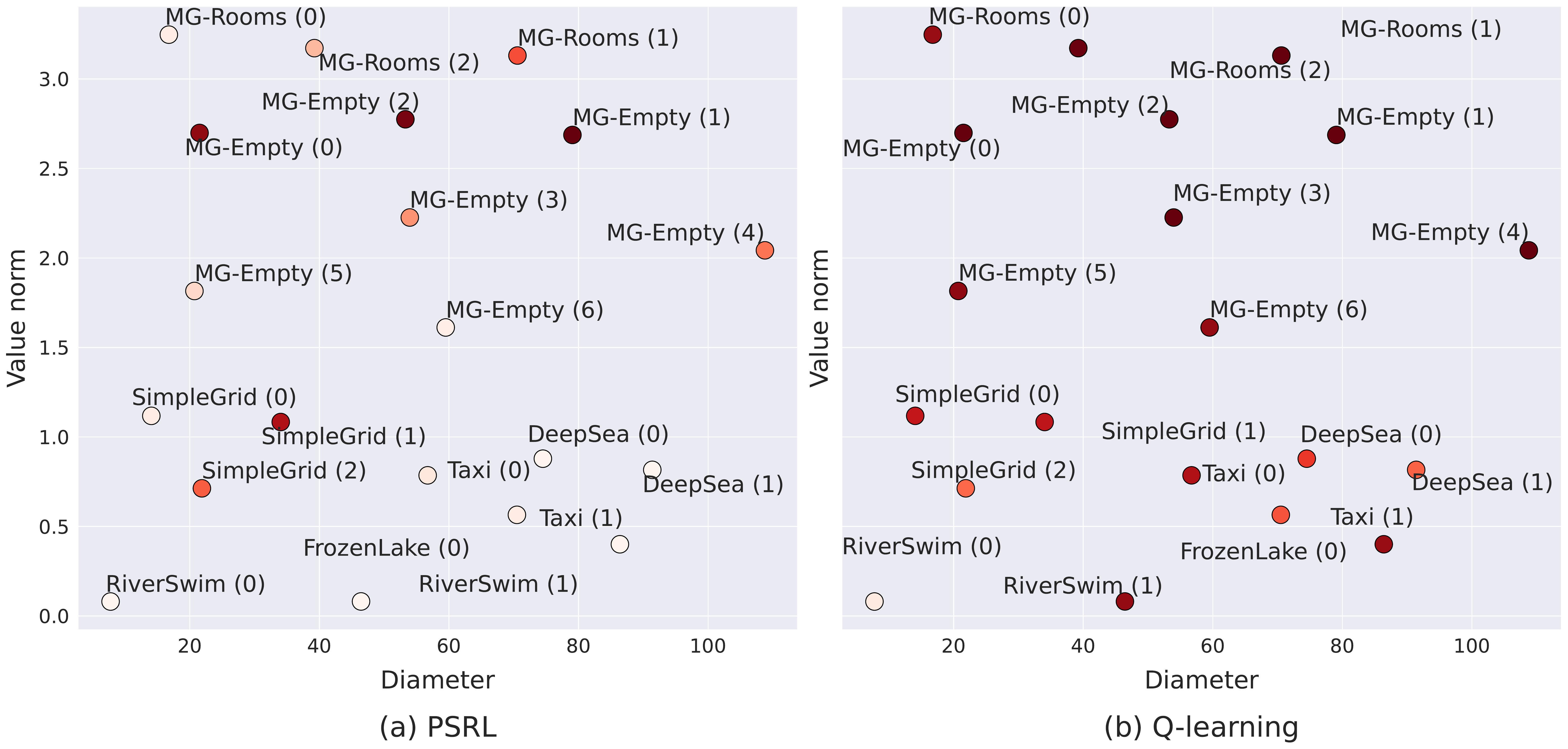}}\\
    \subfloat[][Continuous communicating setting.]{
    \includegraphics[width=\linewidth]{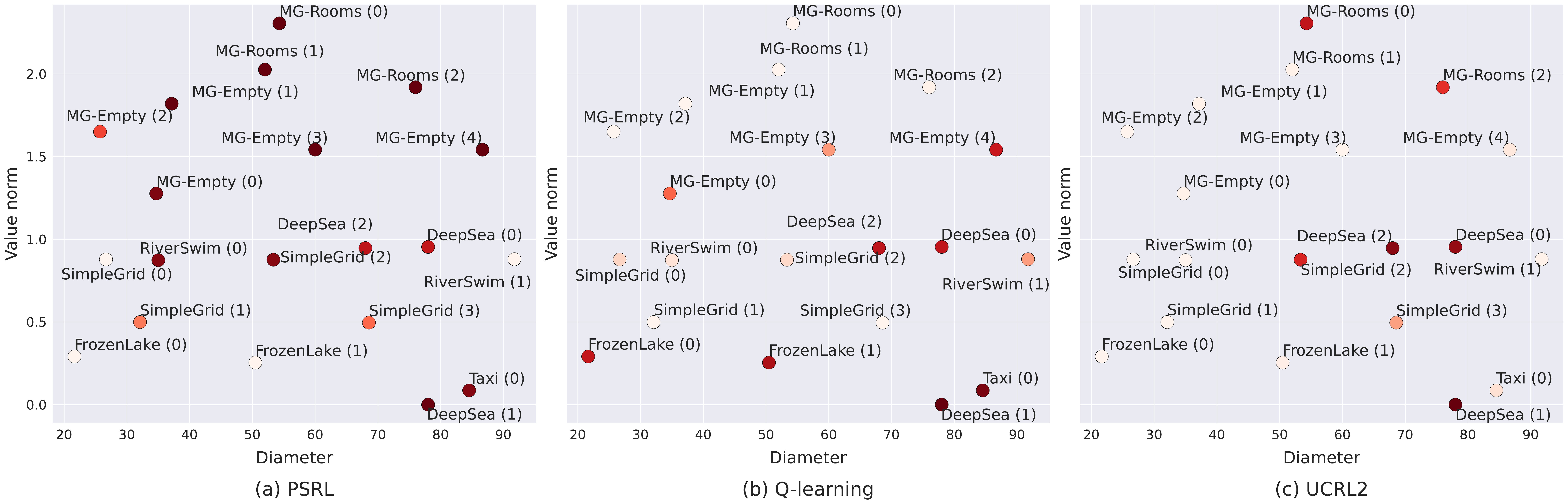}}\\
    \subfloat[][Continuous ergodic setting.]{
    \includegraphics[width=\linewidth]{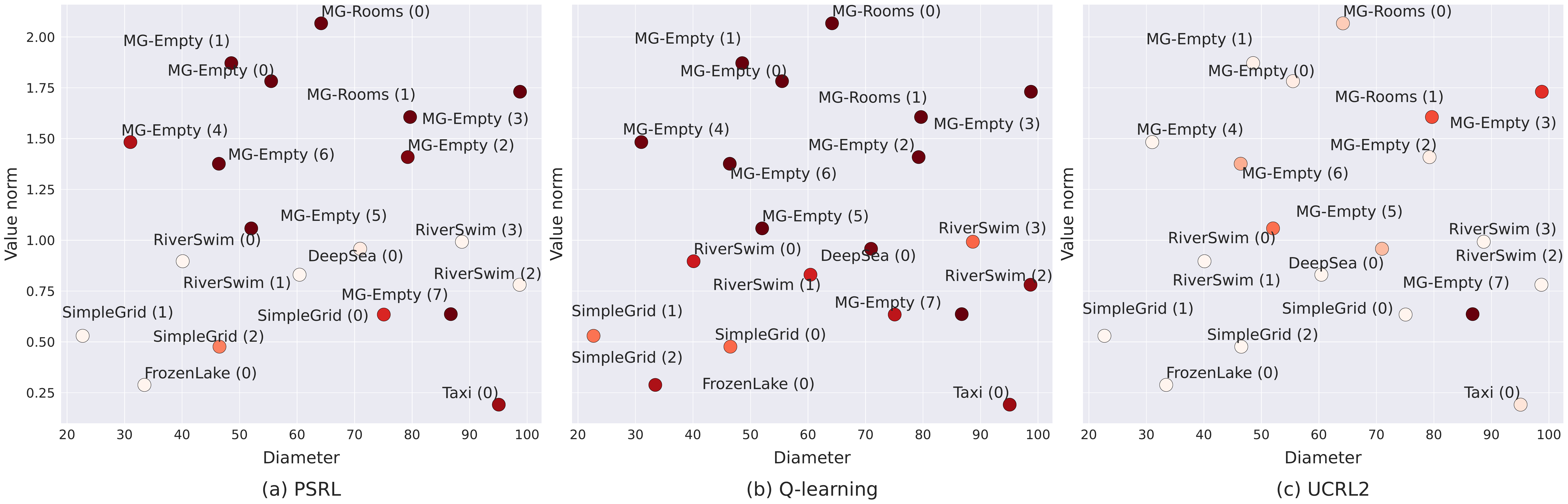}}
\caption{Average cumulative regret obtained by the agents in the continuous ergodic setting placed according to the diameter and the value norm values of the benchmark MDPs.} 
\label{fig:hardness_and_crs}
\end{figure}

\paragraph{Benchmark hardness.}
In order to illustrate how hardness measures relate to cumulative regret in the benchmark,
Figures
\ref{fig:hardness_and_crs}a,
\ref{fig:hardness_and_crs}b,
\ref{fig:hardness_and_crs}c, and
\ref{fig:hardness_and_crs}d
place the average cumulative regret obtained by each agent in each benchmark MDP in a coordinate that corresponds to the diameter and the environmental value norm of that MDP.
In the episodic setting (Figures \ref{fig:hardness_and_crs}a and \ref{fig:hardness_and_crs}b), we note that the environmental value norm has an evident impact on the Q-learning agent, whereas the effect of the diameter is most noticeable in the communicating case. Still, in the episodic setting, the diameter has a small influence compared to the environmental value norm for PSRL.
In the continuous setting (Figures \ref{fig:hardness_and_crs}c and \ref{fig:hardness_and_crs}d), there is generally a positive relationship between both of these hardness measures and the average cumulative regret for UCRL2.
For Q-learning and PSRL, the diameter seems to have a generally smaller influence on the average cumulative regret.

\begin{figure}[htb!]
    \centering
    \includegraphics[width=\linewidth]{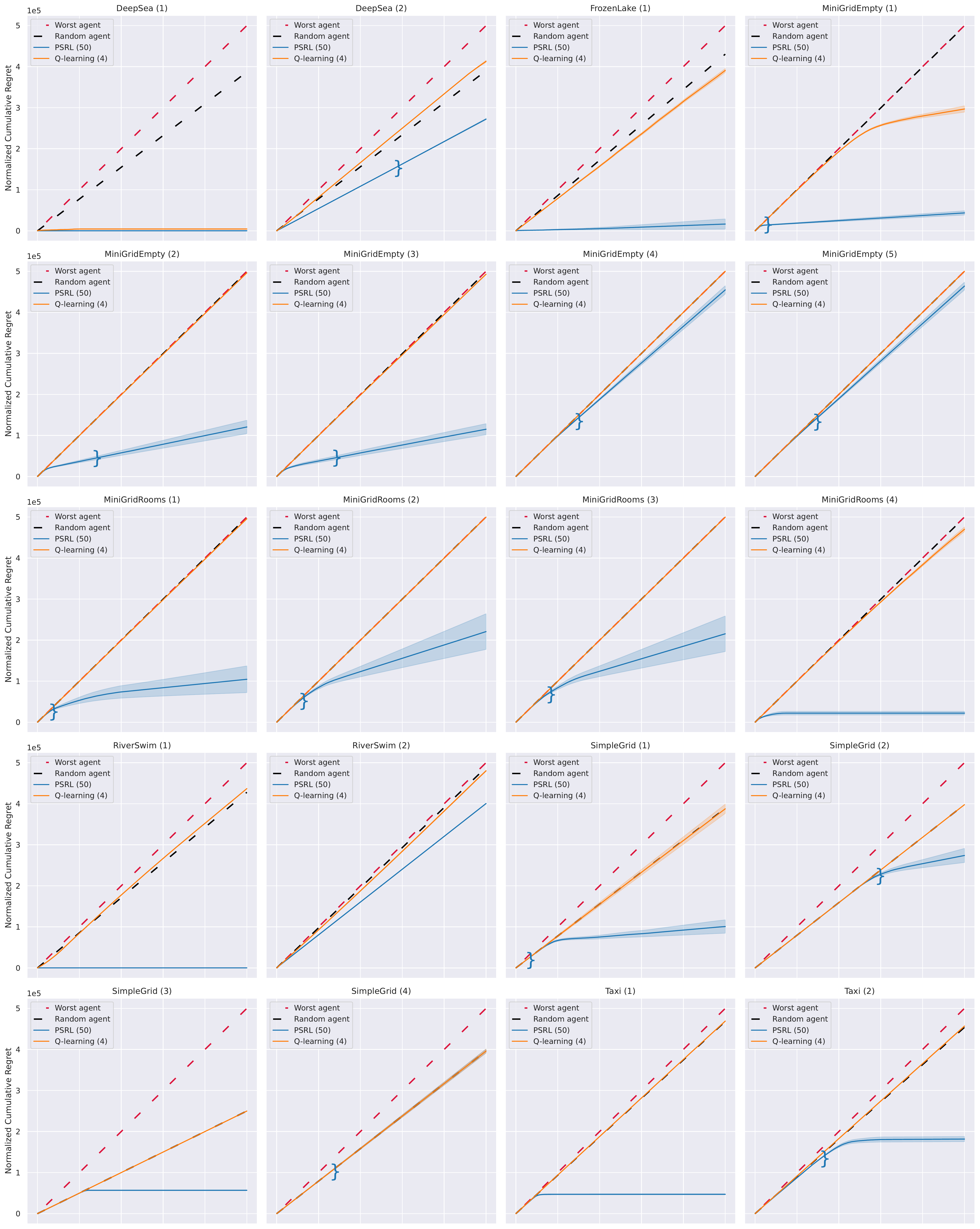}
    \caption{Episodic communicating benchmark results.}
    \label{fig:bec_cr_app}
\end{figure}

\begin{figure}[htb!]
    \centering
    \includegraphics[width=\linewidth]{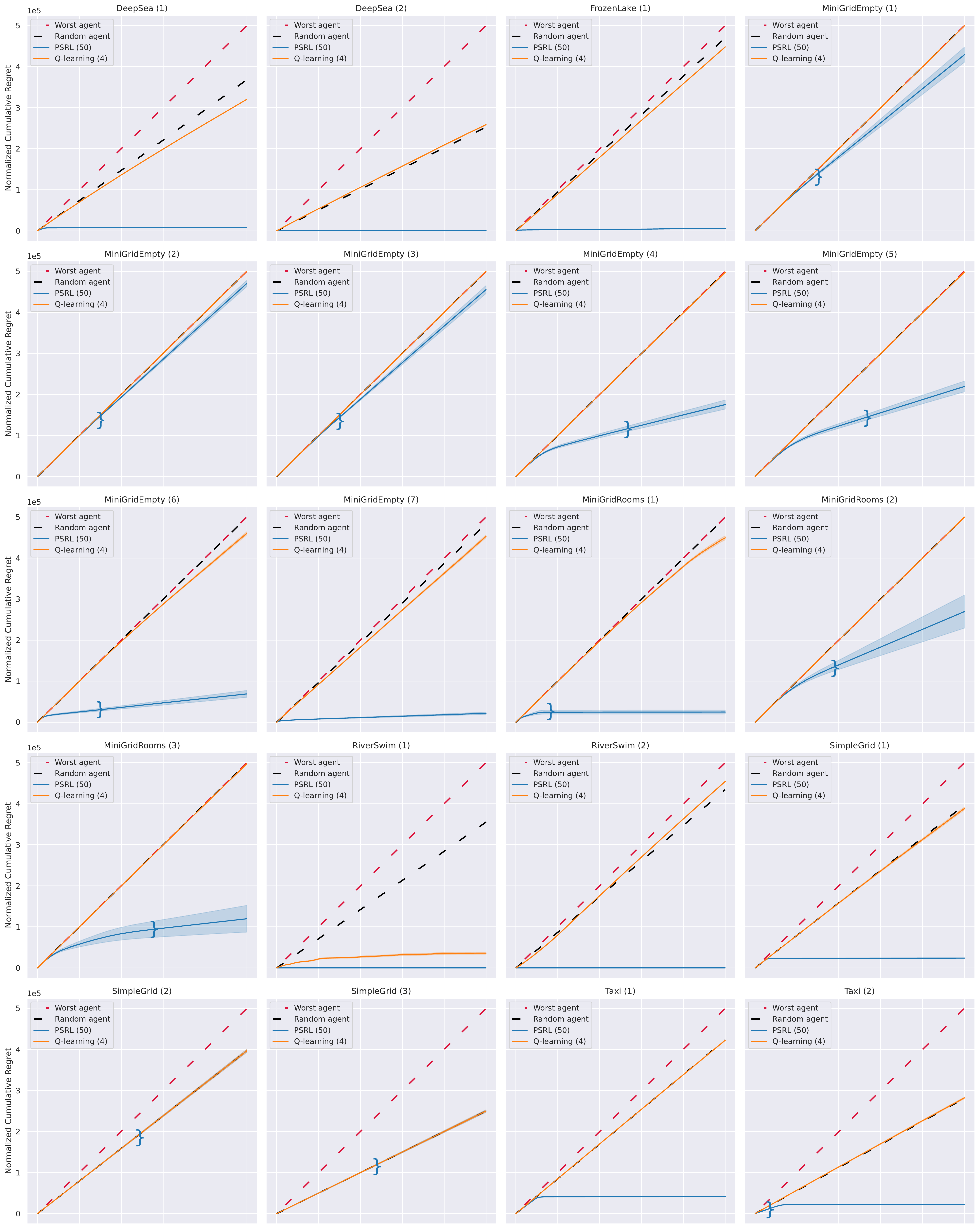}
    \caption{Episodic ergodic benchmark results.}
    \label{fig:bee_cr_app}
\end{figure}

\begin{figure}[htb!]
    \centering
    \includegraphics[width=\linewidth]{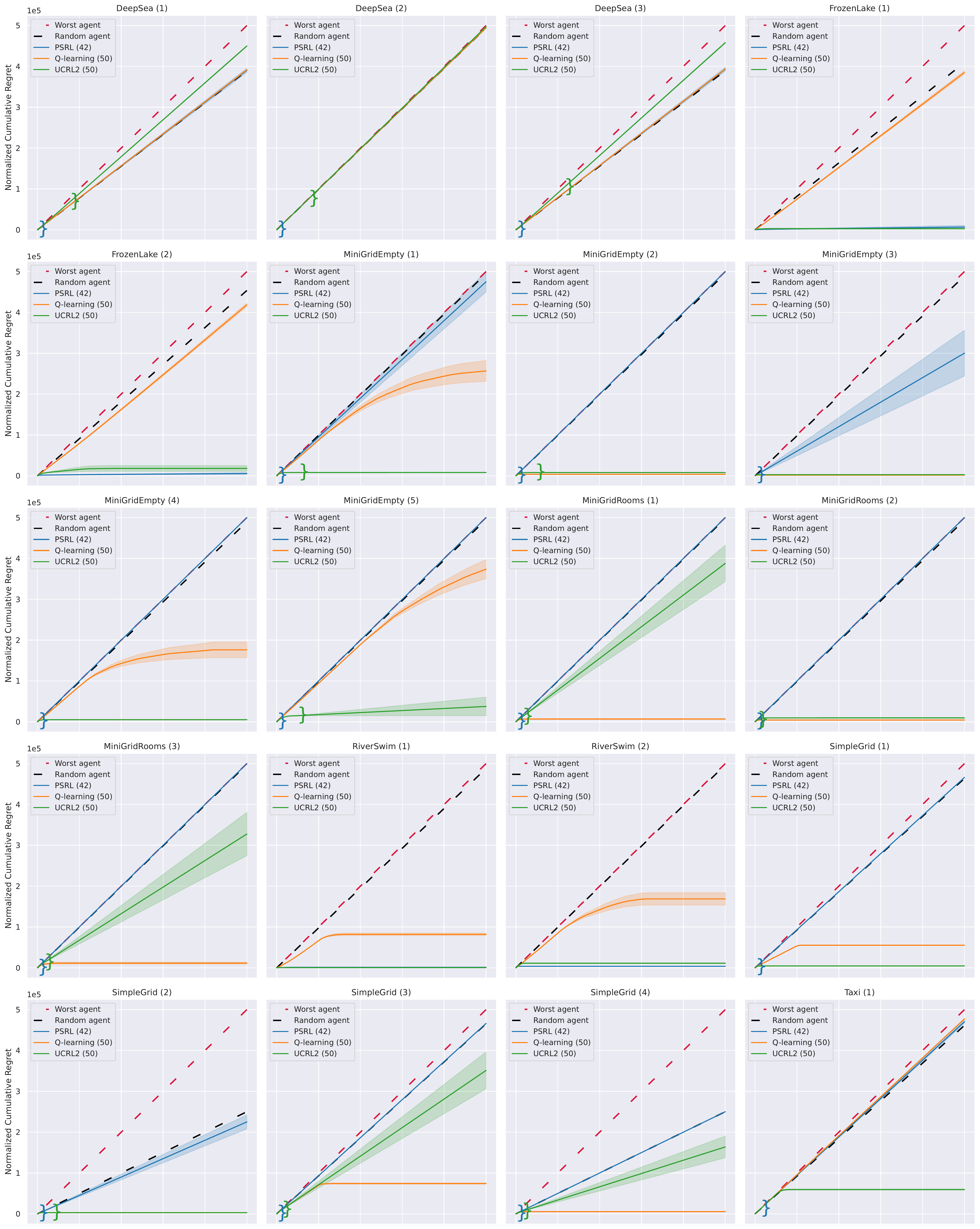}
    \caption{Continuous communicating benchmark results.}
    \label{fig:bcc_cr_app}
\end{figure}

\begin{figure}[htb!]
    \centering
    \includegraphics[width=\linewidth]{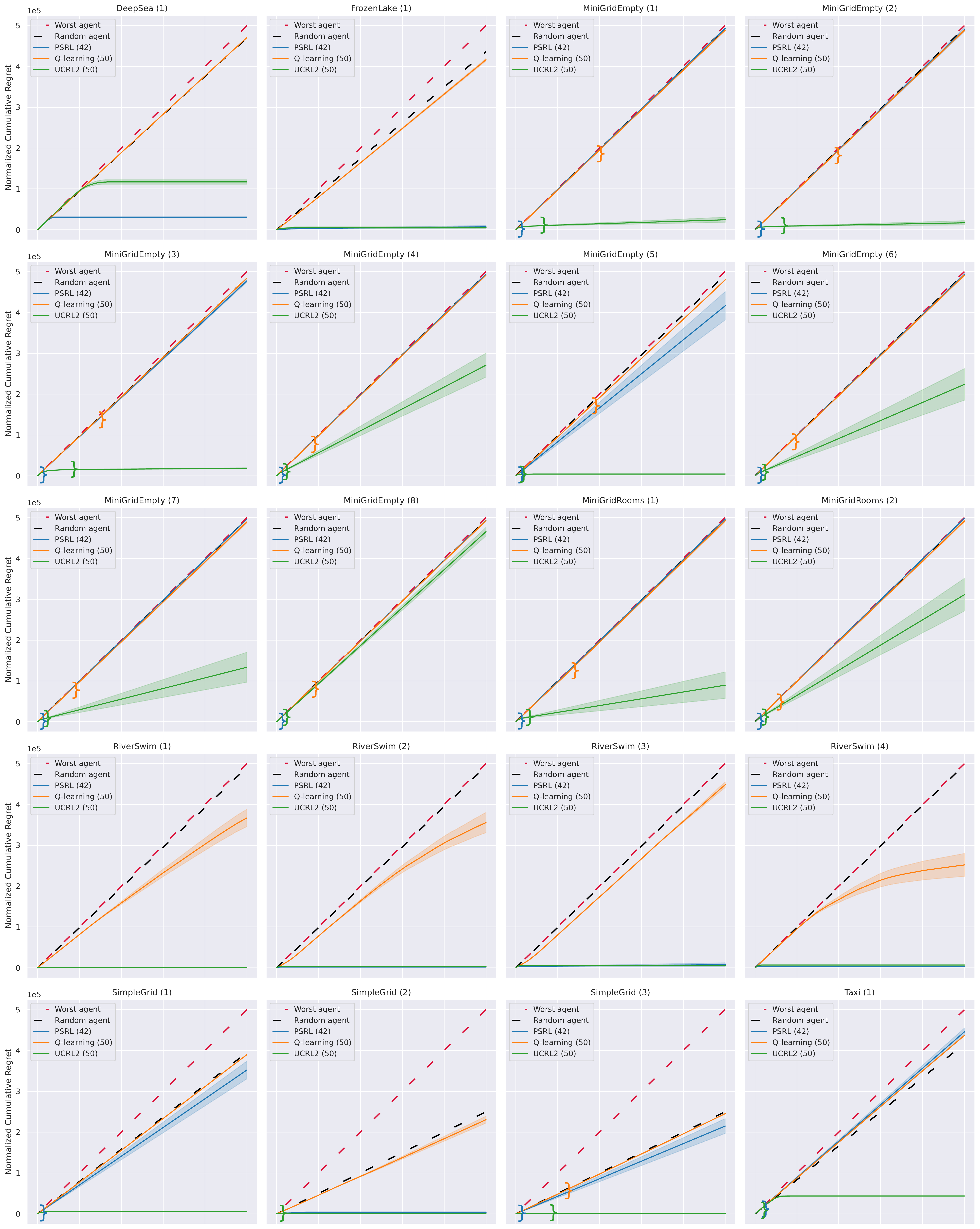}
    \caption{Continuous ergodic benchmark results.}
    \label{fig:bce_cr_app}
\end{figure}

\paragraph{Cumulative regret plots.}
Figures \ref{fig:bec_cr_app}, \ref{fig:bee_cr_app}, \ref{fig:bcc_cr_app}, and \ref{fig:bce_cr_app}
report the expected cumulative regrets for the agents during the agent/MDP interactions along with the cumulative regret of an agent that selects action at random, which provides an informative baseline.
Contrary to the episodic setting, in the continuous setting, the training of UCRL2 and PSRL is stopped for several MDPs of the benchmark.
For PSRL, this typically happens before reaching $10\,000$ time steps, which is particularly damaging.
At this point, the agent has not properly explored the MDP and so it is forced to continue the interaction following a policy that yields a regret similar to the one of the random agent.
UCRL2, instead, tends to terminate the allocated training time at later time steps, which penalizes the performance less.

\paragraph{Cumulative regret tables.}
In Tables 
\ref{tab:episodic_communicating_benchmark_result_table},
\ref{tab:episodic_ergodic_benchmark_result_table},
\ref{tab:continuous_communicating_benchmark_result_table}, and
\ref{tab:continuous_ergodic_benchmark_result_table},
we report the per-step normalized regrets with standard deviations along with the number of seeds for which the agent has been able to complete the total number of training time steps before exceeding the time limit.
We highlight in bold the best performing agent for each MDP.
The same information has been summarized in Table \ref{tab:benchmark_results_main} (Section \ref{sec:benchmark}).

\begin{table}[htb!]
  \centering
  \caption{Summary of benchmark results for the non-tabular \bsuite baselines.}%
  \subfloat[][Episodic communicating setting.]{
\resizebox{!}{4.5cm}{%
\begin{tabular}{lcc}
\toprule
{} &                    PSRL &              Q-learning \\
\midrule
DeepSea          &  $\mathbf{0.00}\pm0.00$ &           $0.01\pm0.01$ \\
                 &  $\mathbf{0.54}\pm0.01$ &           $0.83\pm0.02$ \\
\arrayrulecolor{black!15}\midrule%
FrozenLake       &  $\mathbf{0.03}\pm0.11$ &           $0.78\pm0.04$ \\
\arrayrulecolor{black!15}\midrule%
MG-Empty    &  $\mathbf{0.09}\pm0.05$ &           $0.59\pm0.07$ \\
                 &  $\mathbf{0.24}\pm0.15$ &           $0.99\pm0.00$ \\
                 &  $\mathbf{0.23}\pm0.12$ &           $0.99\pm0.01$ \\
                 &  $\mathbf{0.91}\pm0.09$ &           $1.00\pm0.00$ \\
                 &  $\mathbf{0.93}\pm0.09$ &           $1.00\pm0.00$ \\
\arrayrulecolor{black!15}\midrule%
MG-Rooms    &  $\mathbf{0.21}\pm0.29$ &           $0.99\pm0.01$ \\
                 &  $\mathbf{0.44}\pm0.39$ &           $1.00\pm0.00$ \\
                 &  $\mathbf{0.43}\pm0.39$ &           $1.00\pm0.00$ \\
                 &  $\mathbf{0.04}\pm0.04$ &           $0.94\pm0.05$ \\
\arrayrulecolor{black!15}\midrule%
RiverSwim        &  $\mathbf{0.00}\pm0.00$ &           $0.87\pm0.00$ \\
                 &  $\mathbf{0.80}\pm0.00$ &           $0.96\pm0.01$ \\
\arrayrulecolor{black!15}\midrule%
SimpleGrid       &  $\mathbf{0.20}\pm0.15$ &           $0.78\pm0.10$ \\
                 &  $\mathbf{0.55}\pm0.15$ &           $0.80\pm0.00$ \\
                 &  $\mathbf{0.11}\pm0.01$ &           $0.50\pm0.00$ \\
                 &  $\mathbf{0.79}\pm0.04$ &  $\mathbf{0.79}\pm0.04$ \\
\arrayrulecolor{black!15}\midrule%
Taxi             &  $\mathbf{0.09}\pm0.01$ &           $0.94\pm0.00$ \\
                 &  $\mathbf{0.36}\pm0.06$ &           $0.91\pm0.01$ \\
\arrayrulecolor{black!30}\midrule%
\textit{Average} &  $\mathbf{0.35}\pm0.30$ &           $0.83\pm0.23$ \\
\arrayrulecolor{black!15}\midrule%
\end{tabular}}
  }%
  \hfill
  \subfloat[][Continuous communicating setting.]{
\resizebox{!}{4.5cm}{%
\begin{tabular}{lccc}
\toprule
{} &                    PSRL &              Q-learning &                   UCRL2 \\
\midrule
DeepSea          &  $\mathbf{0.78}\pm0.05$ &  $\mathbf{0.78}\pm0.00$ &           $0.90\pm0.01$ \\
                 &  $\mathbf{0.99}\pm0.00$ &  $\mathbf{0.99}\pm0.00$ &  $\mathbf{0.99}\pm0.00$ \\
                 &  $\mathbf{0.79}\pm0.04$ &  $\mathbf{0.79}\pm0.00$ &           $0.92\pm0.01$ \\
\arrayrulecolor{black!15}\midrule%
FrozenLake       &  $\mathbf{0.01}\pm0.04$ &           $0.77\pm0.04$ &  $\mathbf{0.01}\pm0.01$ \\
                 &  $\mathbf{0.01}\pm0.02$ &           $0.84\pm0.04$ &           $0.04\pm0.06$ \\
\arrayrulecolor{black!15}\midrule%
MG-Empty    &           $0.95\pm0.22$ &           $0.51\pm0.23$ &  $\mathbf{0.02}\pm0.00$ \\
                 &           $1.00\pm0.00$ &  $\mathbf{0.01}\pm0.00$ &           $0.02\pm0.00$ \\
                 &           $0.60\pm0.50$ &  $\mathbf{0.00}\pm0.00$ &           $0.01\pm0.00$ \\
                 &           $1.00\pm0.00$ &           $0.35\pm0.17$ &  $\mathbf{0.01}\pm0.00$ \\
                 &           $1.00\pm0.00$ &           $0.75\pm0.21$ &  $\mathbf{0.08}\pm0.20$ \\
\arrayrulecolor{black!15}\midrule%
MG-Rooms    &           $1.00\pm0.00$ &  $\mathbf{0.01}\pm0.01$ &           $0.78\pm0.40$ \\
                 &           $1.00\pm0.00$ &  $\mathbf{0.01}\pm0.01$ &           $0.02\pm0.01$ \\
                 &           $1.00\pm0.00$ &  $\mathbf{0.02}\pm0.02$ &           $0.66\pm0.47$ \\
\arrayrulecolor{black!15}\midrule%
RiverSwim        &  $\mathbf{0.00}\pm0.01$ &           $0.16\pm0.03$ &  $\mathbf{0.00}\pm0.00$ \\
                 &  $\mathbf{0.01}\pm0.00$ &           $0.34\pm0.14$ &           $0.02\pm0.01$ \\
\arrayrulecolor{black!15}\midrule%
SimpleGrid       &           $0.93\pm0.00$ &           $0.11\pm0.01$ &  $\mathbf{0.01}\pm0.00$ \\
                 &           $0.45\pm0.15$ &  $\mathbf{0.01}\pm0.00$ &  $\mathbf{0.01}\pm0.00$ \\
                 &           $0.93\pm0.00$ &  $\mathbf{0.15}\pm0.01$ &           $0.70\pm0.40$ \\
                 &           $0.50\pm0.00$ &  $\mathbf{0.01}\pm0.00$ &           $0.33\pm0.24$ \\
\arrayrulecolor{black!15}\midrule%
Taxi             &           $0.94\pm0.04$ &           $0.95\pm0.00$ &  $\mathbf{0.12}\pm0.01$ \\
\arrayrulecolor{black!30}\midrule%
\textit{Average} &           $0.69\pm0.38$ &           $0.38\pm0.37$ &  $\mathbf{0.28}\pm0.37$ \\
\arrayrulecolor{black!15}\midrule%
\end{tabular}}
  }\\
  \subfloat[][Episodic ergodic setting.]{
\resizebox{!}{4.5cm}{%
\begin{tabular}{lcc}
\toprule
{} &                    PSRL &              Q-learning \\
\midrule
DeepSea          &  $\mathbf{0.01}\pm0.00$ &           $0.64\pm0.00$ \\
                 &  $\mathbf{0.00}\pm0.00$ &           $0.52\pm0.01$ \\
\arrayrulecolor{black!15}\midrule%
FrozenLake       &  $\mathbf{0.01}\pm0.00$ &           $0.90\pm0.01$ \\
\arrayrulecolor{black!15}\midrule%
MG-Empty    &  $\mathbf{0.86}\pm0.16$ &           $1.00\pm0.00$ \\
                 &  $\mathbf{0.94}\pm0.07$ &           $1.00\pm0.00$ \\
                 &  $\mathbf{0.91}\pm0.09$ &           $1.00\pm0.00$ \\
                 &  $\mathbf{0.35}\pm0.10$ &           $1.00\pm0.00$ \\
                 &  $\mathbf{0.44}\pm0.12$ &           $1.00\pm0.00$ \\
                 &  $\mathbf{0.14}\pm0.08$ &           $0.92\pm0.04$ \\
                 &  $\mathbf{0.04}\pm0.03$ &           $0.91\pm0.03$ \\
\arrayrulecolor{black!15}\midrule%
MG-Rooms    &  $\mathbf{0.05}\pm0.04$ &           $0.90\pm0.04$ \\
                 &  $\mathbf{0.54}\pm0.36$ &           $1.00\pm0.00$ \\
                 &  $\mathbf{0.24}\pm0.29$ &           $0.99\pm0.01$ \\
\arrayrulecolor{black!15}\midrule%
RiverSwim        &  $\mathbf{0.00}\pm0.00$ &           $0.07\pm0.02$ \\
                 &  $\mathbf{0.00}\pm0.00$ &           $0.91\pm0.01$ \\
\arrayrulecolor{black!15}\midrule%
SimpleGrid       &  $\mathbf{0.05}\pm0.01$ &           $0.78\pm0.03$ \\
                 &  $\mathbf{0.79}\pm0.03$ &  $\mathbf{0.79}\pm0.03$ \\
                 &  $\mathbf{0.50}\pm0.03$ &  $\mathbf{0.50}\pm0.03$ \\
\arrayrulecolor{black!15}\midrule%
Taxi             &  $\mathbf{0.08}\pm0.01$ &           $0.84\pm0.01$ \\
                 &  $\mathbf{0.05}\pm0.00$ &           $0.56\pm0.02$ \\
\arrayrulecolor{black!30}\midrule%
\textit{Average} &  $\mathbf{0.30}\pm0.33$ &           $0.81\pm0.24$ \\
\arrayrulecolor{black!15}\midrule%
\end{tabular}}
  }
  \hfill
  \subfloat[][Continuous ergodic setting.]{
\resizebox{!}{4.5cm}{%
\begin{tabular}{lccc}
\toprule
{} &                    PSRL &     Q-learning &                   UCRL2 \\
\midrule
DeepSea          &  $\mathbf{0.06}\pm0.01$ &  $0.94\pm0.00$ &           $0.23\pm0.05$ \\
\arrayrulecolor{black!15}\midrule%
FrozenLake       &  $\mathbf{0.01}\pm0.03$ &  $0.83\pm0.03$ &  $\mathbf{0.01}\pm0.02$ \\
\arrayrulecolor{black!15}\midrule%
MG-Empty    &           $0.99\pm0.01$ &  $0.98\pm0.02$ &  $\mathbf{0.05}\pm0.06$ \\
                 &           $0.98\pm0.04$ &  $0.98\pm0.02$ &  $\mathbf{0.03}\pm0.05$ \\
                 &           $0.95\pm0.03$ &  $0.97\pm0.00$ &  $\mathbf{0.04}\pm0.01$ \\
                 &           $0.99\pm0.01$ &  $0.98\pm0.01$ &  $\mathbf{0.54}\pm0.26$ \\
                 &           $0.83\pm0.31$ &  $0.96\pm0.01$ &  $\mathbf{0.01}\pm0.00$ \\
                 &           $0.99\pm0.02$ &  $0.98\pm0.02$ &  $\mathbf{0.45}\pm0.35$ \\
                 &           $0.99\pm0.01$ &  $0.98\pm0.03$ &  $\mathbf{0.27}\pm0.33$ \\
                 &           $0.99\pm0.01$ &  $0.98\pm0.01$ &  $\mathbf{0.93}\pm0.09$ \\
\arrayrulecolor{black!15}\midrule%
MG-Rooms    &           $0.99\pm0.02$ &  $0.98\pm0.03$ &  $\mathbf{0.18}\pm0.29$ \\
                 &           $1.00\pm0.00$ &  $0.98\pm0.02$ &  $\mathbf{0.62}\pm0.36$ \\
\arrayrulecolor{black!15}\midrule%
RiverSwim        &  $\mathbf{0.00}\pm0.00$ &  $0.73\pm0.19$ &  $\mathbf{0.00}\pm0.00$ \\
                 &  $\mathbf{0.00}\pm0.00$ &  $0.71\pm0.22$ &           $0.01\pm0.00$ \\
                 &           $0.02\pm0.04$ &  $0.90\pm0.06$ &  $\mathbf{0.01}\pm0.01$ \\
                 &  $\mathbf{0.01}\pm0.00$ &  $0.50\pm0.25$ &  $\mathbf{0.01}\pm0.01$ \\
\arrayrulecolor{black!15}\midrule%
SimpleGrid       &           $0.70\pm0.19$ &  $0.78\pm0.00$ &  $\mathbf{0.01}\pm0.01$ \\
                 &           $0.01\pm0.02$ &  $0.46\pm0.08$ &  $\mathbf{0.00}\pm0.00$ \\
                 &           $0.43\pm0.16$ &  $0.49\pm0.00$ &  $\mathbf{0.00}\pm0.00$ \\
\arrayrulecolor{black!15}\midrule%
Taxi             &           $0.89\pm0.08$ &  $0.87\pm0.01$ &  $\mathbf{0.09}\pm0.01$ \\
\arrayrulecolor{black!30}\midrule%
\textit{Average} &           $0.59\pm0.44$ &  $0.85\pm0.18$ &  $\mathbf{0.17}\pm0.26$ \\
\arrayrulecolor{black!15}\midrule%
\end{tabular}}
  }
  \label{tab:benchmark_table}%
\end{table}

\clearpage

\begin{table}[htb]
    \centering
    \caption{Episodic communicating benchmark final time step performance indicators.}
 \resizebox{\textwidth}{!}{%
\begin{tabular}{lllllll}
\toprule
         &            & Norm. cumulative reward & Norm. cumulative expected reward & Norm. cumulative regret & Steps per second & \# completed seeds \\
MDP & Agent &                         &                                  &                         &                  &                    \\
\midrule
DeepSea (1) & PSRL &           $1.00\pm0.00$ &                    $1.00\pm0.00$ &           $0.00\pm0.00$ &    $0.02\pm0.00$ &            $20/20$ \\
\arrayrulecolor{black!15}\cmidrule{2-6}
         & Q-learning &           $0.99\pm0.01$ &                    $0.99\pm0.01$ &           $0.01\pm0.01$ &    $0.03\pm0.01$ &            $20/20$ \\
\arrayrulecolor{black!15}\cmidrule{1-6}
DeepSea (2) & PSRL &           $0.23\pm0.06$ &                    $0.46\pm0.01$ &           $0.54\pm0.01$ &    $0.00\pm0.00$ &             $0/20$ \\
\arrayrulecolor{black!15}\cmidrule{2-6}
         & Q-learning &           $0.17\pm0.02$ &                    $0.17\pm0.02$ &           $0.83\pm0.02$ &    $0.01\pm0.00$ &            $20/20$ \\
\arrayrulecolor{black!15}\cmidrule{1-6}
FrozenLake (1) & PSRL &           $0.92\pm0.15$ &                    $0.97\pm0.11$ &           $0.03\pm0.11$ &    $0.01\pm0.01$ &            $20/20$ \\
\arrayrulecolor{black!15}\cmidrule{2-6}
         & Q-learning &           $0.22\pm0.04$ &                    $0.22\pm0.04$ &           $0.78\pm0.04$ &    $0.01\pm0.00$ &            $20/20$ \\
\arrayrulecolor{black!15}\cmidrule{1-6}
MG-Empty (1) & PSRL &           $0.88\pm0.05$ &                    $0.92\pm0.04$ &           $0.09\pm0.04$ &    $0.00\pm0.00$ &            $19/20$ \\
\arrayrulecolor{black!15}\cmidrule{2-6}
         & Q-learning &           $0.43\pm0.06$ &                    $0.43\pm0.06$ &           $0.59\pm0.07$ &    $0.01\pm0.00$ &            $20/20$ \\
\arrayrulecolor{black!15}\cmidrule{1-6}
MG-Empty (2) & PSRL &           $0.72\pm0.14$ &                    $0.83\pm0.08$ &           $0.24\pm0.14$ &    $0.00\pm0.00$ &            $14/20$ \\
\arrayrulecolor{black!15}\cmidrule{2-6}
         & Q-learning &           $0.01\pm0.01$ &                    $0.01\pm0.01$ &           $0.99\pm0.00$ &    $0.01\pm0.00$ &            $20/20$ \\
\arrayrulecolor{black!15}\cmidrule{1-6}
MG-Empty (3) & PSRL &           $0.75\pm0.09$ &                    $0.80\pm0.10$ &           $0.23\pm0.12$ &    $0.00\pm0.00$ &            $14/20$ \\
\arrayrulecolor{black!15}\cmidrule{2-6}
         & Q-learning &           $0.02\pm0.01$ &                    $0.02\pm0.01$ &           $0.99\pm0.01$ &    $0.01\pm0.00$ &            $20/20$ \\
\arrayrulecolor{black!15}\cmidrule{1-6}
MG-Empty (4) & PSRL &           $0.00\pm0.00$ &                    $0.12\pm0.12$ &           $0.91\pm0.09$ &    $0.00\pm0.00$ &             $0/20$ \\
\arrayrulecolor{black!15}\cmidrule{2-6}
         & Q-learning &           $0.00\pm0.00$ &                    $0.00\pm0.00$ &           $1.00\pm0.00$ &    $0.01\pm0.00$ &            $20/20$ \\
\arrayrulecolor{black!15}\cmidrule{1-6}
MG-Empty (5) & PSRL &           $0.00\pm0.00$ &                    $0.10\pm0.13$ &           $0.93\pm0.08$ &    $0.00\pm0.00$ &             $0/20$ \\
\arrayrulecolor{black!15}\cmidrule{2-6}
         & Q-learning &           $0.00\pm0.00$ &                    $0.00\pm0.00$ &           $1.00\pm0.00$ &    $0.01\pm0.00$ &            $20/20$ \\
\arrayrulecolor{black!15}\cmidrule{1-6}
MG-Rooms (1) & PSRL &           $0.22\pm0.33$ &                    $0.79\pm0.28$ &           $0.21\pm0.28$ &    $0.00\pm0.00$ &            $18/20$ \\
\arrayrulecolor{black!15}\cmidrule{2-6}
         & Q-learning &           $0.01\pm0.01$ &                    $0.01\pm0.01$ &           $0.99\pm0.01$ &    $0.01\pm0.00$ &            $20/20$ \\
\arrayrulecolor{black!15}\cmidrule{1-6}
MG-Rooms (2) & PSRL &           $0.00\pm0.00$ &                    $0.56\pm0.38$ &           $0.44\pm0.38$ &    $0.00\pm0.00$ &            $14/20$ \\
\arrayrulecolor{black!15}\cmidrule{2-6}
         & Q-learning &           $0.00\pm0.00$ &                    $0.00\pm0.00$ &           $1.00\pm0.00$ &    $0.01\pm0.00$ &            $20/20$ \\
\arrayrulecolor{black!15}\cmidrule{1-6}
MG-Rooms (3) & PSRL &           $0.00\pm0.00$ &                    $0.57\pm0.38$ &           $0.43\pm0.38$ &    $0.00\pm0.00$ &            $12/20$ \\
\arrayrulecolor{black!15}\cmidrule{2-6}
         & Q-learning &           $0.00\pm0.00$ &                    $0.00\pm0.00$ &           $1.00\pm0.00$ &    $0.01\pm0.00$ &            $20/20$ \\
\arrayrulecolor{black!15}\cmidrule{1-6}
MG-Rooms (4) & PSRL &           $0.62\pm0.34$ &                    $0.96\pm0.04$ &           $0.04\pm0.04$ &    $0.01\pm0.00$ &            $20/20$ \\
\arrayrulecolor{black!15}\cmidrule{2-6}
         & Q-learning &           $0.07\pm0.05$ &                    $0.07\pm0.05$ &           $0.94\pm0.05$ &    $0.01\pm0.00$ &            $20/20$ \\
\arrayrulecolor{black!15}\cmidrule{1-6}
RiverSwim (1) & PSRL &           $0.94\pm0.17$ &                    $1.00\pm0.00$ &           $0.00\pm0.00$ &    $0.02\pm0.00$ &            $20/20$ \\
\arrayrulecolor{black!15}\cmidrule{2-6}
         & Q-learning &           $0.13\pm0.00$ &                    $0.13\pm0.00$ &           $0.87\pm0.00$ &    $0.01\pm0.00$ &            $20/20$ \\
\arrayrulecolor{black!15}\cmidrule{1-6}
RiverSwim (2) & PSRL &           $0.17\pm0.00$ &                    $0.20\pm0.00$ &           $0.80\pm0.00$ &    $0.01\pm0.00$ &            $20/20$ \\
\arrayrulecolor{black!15}\cmidrule{2-6}
         & Q-learning &           $0.04\pm0.01$ &                    $0.04\pm0.01$ &           $0.96\pm0.01$ &    $0.01\pm0.00$ &            $20/20$ \\
\arrayrulecolor{black!15}\cmidrule{1-6}
SimpleGrid (1) & PSRL &           $0.63\pm0.18$ &                    $0.80\pm0.14$ &           $0.20\pm0.14$ &    $0.00\pm0.00$ &            $19/20$ \\
\arrayrulecolor{black!15}\cmidrule{2-6}
         & Q-learning &           $0.22\pm0.10$ &                    $0.22\pm0.10$ &           $0.78\pm0.10$ &    $0.01\pm0.00$ &            $20/20$ \\
\arrayrulecolor{black!15}\cmidrule{1-6}
SimpleGrid (2) & PSRL &           $0.37\pm0.15$ &                    $0.45\pm0.15$ &           $0.55\pm0.15$ &    $0.00\pm0.00$ &             $4/20$ \\
\arrayrulecolor{black!15}\cmidrule{2-6}
         & Q-learning &           $0.21\pm0.00$ &                    $0.20\pm0.00$ &           $0.80\pm0.00$ &    $0.01\pm0.00$ &            $20/20$ \\
\arrayrulecolor{black!15}\cmidrule{1-6}
SimpleGrid (3) & PSRL &           $0.51\pm0.03$ &                    $0.89\pm0.01$ &           $0.11\pm0.01$ &    $0.00\pm0.00$ &            $20/20$ \\
\arrayrulecolor{black!15}\cmidrule{2-6}
         & Q-learning &           $0.50\pm0.00$ &                    $0.50\pm0.00$ &           $0.50\pm0.00$ &    $0.01\pm0.00$ &            $20/20$ \\
\arrayrulecolor{black!15}\cmidrule{1-6}
SimpleGrid (4) & PSRL &           $0.21\pm0.04$ &                    $0.21\pm0.04$ &           $0.79\pm0.04$ &    $0.00\pm0.00$ &             $0/20$ \\
\arrayrulecolor{black!15}\cmidrule{2-6}
         & Q-learning &           $0.21\pm0.04$ &                    $0.21\pm0.04$ &           $0.79\pm0.04$ &    $0.00\pm0.00$ &            $20/20$ \\
\arrayrulecolor{black!15}\cmidrule{1-6}
Taxi (1) & PSRL &           $0.88\pm0.01$ &                    $0.91\pm0.01$ &           $0.09\pm0.01$ &    $0.01\pm0.00$ &            $20/20$ \\
\arrayrulecolor{black!15}\cmidrule{2-6}
         & Q-learning &           $0.06\pm0.00$ &                    $0.06\pm0.00$ &           $0.94\pm0.00$ &    $0.01\pm0.00$ &            $20/20$ \\
\arrayrulecolor{black!15}\cmidrule{1-6}
Taxi (2) & PSRL &           $0.51\pm0.11$ &                    $0.64\pm0.06$ &           $0.36\pm0.05$ &    $0.00\pm0.00$ &            $13/20$ \\
\arrayrulecolor{black!15}\cmidrule{2-6}
         & Q-learning &           $0.09\pm0.01$ &                    $0.09\pm0.01$ &           $0.91\pm0.01$ &    $0.01\pm0.00$ &            $20/20$ \\
\bottomrule
\end{tabular}}
    \label{tab:episodic_communicating_benchmark_result_table}
\end{table}

\begin{table}[htb]
    \centering
    \caption{Episodic ergodic benchmark final time step performance indicators.}
 \resizebox{\textwidth}{!}{%
\begin{tabular}{lllllll}
\toprule
         &            & Norm. cumulative reward & Norm. cumulative expected reward & Norm. cumulative regret & Steps per second & \# completed seeds \\
MDP & Agent &                         &                                  &                         &                  &                    \\
\midrule
DeepSea (1) & PSRL &           $0.69\pm0.17$ &                    $0.99\pm0.00$ &           $0.01\pm0.00$ &    $0.01\pm0.00$ &            $20/20$ \\
\arrayrulecolor{black!15}\cmidrule{2-6}
         & Q-learning &           $0.36\pm0.00$ &                    $0.36\pm0.00$ &           $0.64\pm0.00$ &    $0.01\pm0.00$ &            $20/20$ \\
\arrayrulecolor{black!15}\cmidrule{1-6}
DeepSea (2) & PSRL &           $0.79\pm0.07$ &                    $1.00\pm0.00$ &           $0.00\pm0.00$ &    $0.01\pm0.00$ &            $20/20$ \\
\arrayrulecolor{black!15}\cmidrule{2-6}
         & Q-learning &           $0.48\pm0.01$ &                    $0.48\pm0.01$ &           $0.52\pm0.01$ &    $0.01\pm0.00$ &            $20/20$ \\
\arrayrulecolor{black!15}\cmidrule{1-6}
FrozenLake (1) & PSRL &           $0.95\pm0.08$ &                    $0.99\pm0.00$ &           $0.01\pm0.00$ &    $0.01\pm0.00$ &            $20/20$ \\
\arrayrulecolor{black!15}\cmidrule{2-6}
         & Q-learning &           $0.10\pm0.01$ &                    $0.10\pm0.01$ &           $0.90\pm0.01$ &    $0.01\pm0.00$ &            $20/20$ \\
\arrayrulecolor{black!15}\cmidrule{1-6}
MG-Empty (1) & PSRL &           $0.00\pm0.00$ &                    $0.18\pm0.20$ &           $0.86\pm0.16$ &    $0.00\pm0.00$ &             $0/20$ \\
\arrayrulecolor{black!15}\cmidrule{2-6}
         & Q-learning &           $0.00\pm0.00$ &                    $0.00\pm0.00$ &           $1.00\pm0.00$ &    $0.01\pm0.00$ &            $20/20$ \\
\arrayrulecolor{black!15}\cmidrule{1-6}
MG-Empty (2) & PSRL &           $0.00\pm0.00$ &                    $0.08\pm0.10$ &           $0.94\pm0.07$ &    $0.00\pm0.00$ &             $0/20$ \\
\arrayrulecolor{black!15}\cmidrule{2-6}
         & Q-learning &           $0.00\pm0.00$ &                    $0.00\pm0.00$ &           $1.00\pm0.00$ &    $0.01\pm0.00$ &            $20/20$ \\
\arrayrulecolor{black!15}\cmidrule{1-6}
MG-Empty (3) & PSRL &           $0.00\pm0.00$ &                    $0.12\pm0.12$ &           $0.91\pm0.09$ &    $0.00\pm0.00$ &             $0/20$ \\
\arrayrulecolor{black!15}\cmidrule{2-6}
         & Q-learning &           $0.00\pm0.00$ &                    $0.00\pm0.00$ &           $1.00\pm0.00$ &    $0.01\pm0.00$ &            $20/20$ \\
\arrayrulecolor{black!15}\cmidrule{1-6}
MG-Empty (4) & PSRL &           $0.60\pm0.08$ &                    $0.69\pm0.09$ &           $0.35\pm0.10$ &    $0.00\pm0.00$ &             $0/20$ \\
\arrayrulecolor{black!15}\cmidrule{2-6}
         & Q-learning &           $0.01\pm0.00$ &                    $0.01\pm0.00$ &           $1.00\pm0.00$ &    $0.01\pm0.00$ &            $20/20$ \\
\arrayrulecolor{black!15}\cmidrule{1-6}
MG-Empty (5) & PSRL &           $0.52\pm0.10$ &                    $0.62\pm0.11$ &           $0.44\pm0.11$ &    $0.00\pm0.00$ &             $0/20$ \\
\arrayrulecolor{black!15}\cmidrule{2-6}
         & Q-learning &           $0.01\pm0.01$ &                    $0.01\pm0.00$ &           $1.00\pm0.00$ &    $0.01\pm0.00$ &            $20/20$ \\
\arrayrulecolor{black!15}\cmidrule{1-6}
MG-Empty (6) & PSRL &           $0.84\pm0.06$ &                    $0.87\pm0.06$ &           $0.14\pm0.07$ &    $0.00\pm0.00$ &            $14/20$ \\
\arrayrulecolor{black!15}\cmidrule{2-6}
         & Q-learning &           $0.10\pm0.05$ &                    $0.10\pm0.05$ &           $0.92\pm0.03$ &    $0.01\pm0.00$ &            $20/20$ \\
\arrayrulecolor{black!15}\cmidrule{1-6}
MG-Empty (7) & PSRL &           $0.95\pm0.02$ &                    $0.97\pm0.02$ &           $0.04\pm0.03$ &    $0.00\pm0.00$ &            $20/20$ \\
\arrayrulecolor{black!15}\cmidrule{2-6}
         & Q-learning &           $0.13\pm0.05$ &                    $0.13\pm0.05$ &           $0.91\pm0.03$ &    $0.01\pm0.00$ &            $20/20$ \\
\arrayrulecolor{black!15}\cmidrule{1-6}
MG-Rooms (1) & PSRL &           $0.80\pm0.03$ &                    $0.95\pm0.04$ &           $0.05\pm0.04$ &    $0.00\pm0.00$ &            $17/20$ \\
\arrayrulecolor{black!15}\cmidrule{2-6}
         & Q-learning &           $0.10\pm0.04$ &                    $0.10\pm0.04$ &           $0.90\pm0.04$ &    $0.01\pm0.00$ &            $20/20$ \\
\arrayrulecolor{black!15}\cmidrule{1-6}
MG-Rooms (2) & PSRL &           $0.00\pm0.00$ &                    $0.47\pm0.35$ &           $0.54\pm0.35$ &    $0.00\pm0.00$ &             $0/20$ \\
\arrayrulecolor{black!15}\cmidrule{2-6}
         & Q-learning &           $0.00\pm0.00$ &                    $0.00\pm0.00$ &           $1.00\pm0.00$ &    $0.01\pm0.00$ &            $20/20$ \\
\arrayrulecolor{black!15}\cmidrule{1-6}
MG-Rooms (3) & PSRL &           $0.39\pm0.25$ &                    $0.76\pm0.28$ &           $0.24\pm0.28$ &    $0.00\pm0.00$ &             $0/20$ \\
\arrayrulecolor{black!15}\cmidrule{2-6}
         & Q-learning &           $0.01\pm0.01$ &                    $0.01\pm0.01$ &           $0.99\pm0.01$ &    $0.01\pm0.00$ &            $20/20$ \\
\arrayrulecolor{black!15}\cmidrule{1-6}
RiverSwim (1) & PSRL &           $0.98\pm0.07$ &                    $1.00\pm0.00$ &           $0.00\pm0.00$ &    $0.02\pm0.00$ &            $20/20$ \\
\arrayrulecolor{black!15}\cmidrule{2-6}
         & Q-learning &           $0.93\pm0.02$ &                    $0.93\pm0.02$ &           $0.07\pm0.02$ &    $0.01\pm0.00$ &            $20/20$ \\
\arrayrulecolor{black!15}\cmidrule{1-6}
RiverSwim (2) & PSRL &           $0.93\pm0.17$ &                    $1.00\pm0.00$ &           $0.00\pm0.00$ &    $0.02\pm0.00$ &            $20/20$ \\
\arrayrulecolor{black!15}\cmidrule{2-6}
         & Q-learning &           $0.09\pm0.01$ &                    $0.09\pm0.01$ &           $0.91\pm0.01$ &    $0.01\pm0.00$ &            $20/20$ \\
\arrayrulecolor{black!15}\cmidrule{1-6}
SimpleGrid (1) & PSRL &           $0.93\pm0.01$ &                    $0.95\pm0.01$ &           $0.05\pm0.01$ &    $0.00\pm0.00$ &            $20/20$ \\
\arrayrulecolor{black!15}\cmidrule{2-6}
         & Q-learning &           $0.22\pm0.03$ &                    $0.22\pm0.03$ &           $0.78\pm0.03$ &    $0.01\pm0.00$ &            $20/20$ \\
\arrayrulecolor{black!15}\cmidrule{1-6}
SimpleGrid (2) & PSRL &           $0.21\pm0.03$ &                    $0.21\pm0.03$ &           $0.79\pm0.03$ &    $0.00\pm0.00$ &             $0/20$ \\
\arrayrulecolor{black!15}\cmidrule{2-6}
         & Q-learning &           $0.21\pm0.03$ &                    $0.21\pm0.03$ &           $0.79\pm0.03$ &    $0.01\pm0.00$ &            $20/20$ \\
\arrayrulecolor{black!15}\cmidrule{1-6}
SimpleGrid (3) & PSRL &           $0.51\pm0.04$ &                    $0.50\pm0.03$ &           $0.50\pm0.03$ &    $0.00\pm0.00$ &             $0/20$ \\
\arrayrulecolor{black!15}\cmidrule{2-6}
         & Q-learning &           $0.50\pm0.03$ &                    $0.50\pm0.03$ &           $0.50\pm0.03$ &    $0.01\pm0.00$ &            $20/20$ \\
\arrayrulecolor{black!15}\cmidrule{1-6}
Taxi (1) & PSRL &           $0.89\pm0.01$ &                    $0.92\pm0.01$ &           $0.08\pm0.01$ &    $0.01\pm0.00$ &            $20/20$ \\
\arrayrulecolor{black!15}\cmidrule{2-6}
         & Q-learning &           $0.15\pm0.01$ &                    $0.15\pm0.01$ &           $0.84\pm0.01$ &    $0.01\pm0.00$ &            $20/20$ \\
\arrayrulecolor{black!15}\cmidrule{1-6}
Taxi (2) & PSRL &           $0.93\pm0.01$ &                    $0.95\pm0.00$ &           $0.05\pm0.00$ &    $0.00\pm0.00$ &            $19/20$ \\
\arrayrulecolor{black!15}\cmidrule{2-6}
         & Q-learning &           $0.44\pm0.02$ &                    $0.44\pm0.02$ &           $0.56\pm0.02$ &    $0.01\pm0.00$ &            $20/20$ \\
\bottomrule
\end{tabular}}
    \label{tab:episodic_ergodic_benchmark_result_table}
\end{table}

\begin{table}[htb]
    \centering
    \caption{Continuous communicating benchmark final time step performance indicators.}
\resizebox{\textwidth}{!}{%
\begin{tabular}{lllllll}
\toprule
         &       & Norm. cumulative reward & Norm. cumulative expected reward & Norm. cumulative regret & Steps per second & \# completed seeds \\
MDP & Agent &                         &                                  &                         &                  &                    \\
\midrule
DeepSea (1) & PSRL &           $0.22\pm0.01$ &                    $0.22\pm0.05$ &           $0.78\pm0.05$ &    $0.00\pm0.00$ &             $0/20$ \\
\arrayrulecolor{black!15}\cmidrule{2-6}
         & Q-learning &           $0.22\pm0.00$ &                    $0.22\pm0.00$ &           $0.78\pm0.00$ &    $0.00\pm0.00$ &            $20/20$ \\
\arrayrulecolor{black!15}\cmidrule{2-6}
         & UCRL2 &           $0.19\pm0.03$ &                    $0.10\pm0.01$ &           $0.90\pm0.01$ &    $0.00\pm0.00$ &             $0/20$ \\
\arrayrulecolor{black!15}\cmidrule{1-6}
DeepSea (2) & PSRL &           $0.01\pm0.00$ &                    $0.01\pm0.00$ &           $0.99\pm0.00$ &    $0.00\pm0.00$ &             $0/20$ \\
\arrayrulecolor{black!15}\cmidrule{2-6}
         & Q-learning &           $0.01\pm0.00$ &                    $0.01\pm0.00$ &           $0.99\pm0.00$ &    $0.00\pm0.00$ &            $20/20$ \\
\arrayrulecolor{black!15}\cmidrule{2-6}
         & UCRL2 &           $0.01\pm0.00$ &                    $0.01\pm0.00$ &           $0.99\pm0.00$ &    $0.00\pm0.00$ &             $0/20$ \\
\arrayrulecolor{black!15}\cmidrule{1-6}
DeepSea (3) & PSRL &           $0.22\pm0.01$ &                    $0.21\pm0.04$ &           $0.79\pm0.04$ &    $0.00\pm0.00$ &             $0/20$ \\
\arrayrulecolor{black!15}\cmidrule{2-6}
         & Q-learning &           $0.21\pm0.00$ &                    $0.21\pm0.00$ &           $0.79\pm0.00$ &    $0.00\pm0.00$ &            $20/20$ \\
\arrayrulecolor{black!15}\cmidrule{2-6}
         & UCRL2 &           $0.17\pm0.03$ &                    $0.08\pm0.01$ &           $0.92\pm0.01$ &    $0.00\pm0.00$ &             $0/20$ \\
\arrayrulecolor{black!15}\cmidrule{1-6}
FrozenLake (1) & PSRL &           $0.19\pm0.20$ &                    $0.99\pm0.03$ &           $0.01\pm0.03$ &    $0.02\pm0.00$ &            $20/20$ \\
\arrayrulecolor{black!15}\cmidrule{2-6}
         & Q-learning &           $0.23\pm0.04$ &                    $0.23\pm0.04$ &           $0.77\pm0.04$ &    $0.01\pm0.00$ &            $20/20$ \\
\arrayrulecolor{black!15}\cmidrule{2-6}
         & UCRL2 &           $0.68\pm0.17$ &                    $0.99\pm0.01$ &           $0.01\pm0.01$ &    $0.02\pm0.01$ &            $20/20$ \\
\arrayrulecolor{black!15}\cmidrule{1-6}
FrozenLake (2) & PSRL &           $0.28\pm0.29$ &                    $0.99\pm0.02$ &           $0.01\pm0.02$ &    $0.01\pm0.00$ &            $20/20$ \\
\arrayrulecolor{black!15}\cmidrule{2-6}
         & Q-learning &           $0.14\pm0.04$ &                    $0.16\pm0.04$ &           $0.84\pm0.04$ &    $0.01\pm0.00$ &            $20/20$ \\
\arrayrulecolor{black!15}\cmidrule{2-6}
         & UCRL2 &           $0.34\pm0.20$ &                    $0.96\pm0.07$ &           $0.04\pm0.06$ &    $0.02\pm0.01$ &            $20/20$ \\
\arrayrulecolor{black!15}\cmidrule{1-6}
MG-Empty (1) & PSRL &           $0.05\pm0.22$ &                    $0.05\pm0.22$ &           $0.95\pm0.22$ &    $0.00\pm0.00$ &             $0/20$ \\
\arrayrulecolor{black!15}\cmidrule{2-6}
         & Q-learning &           $0.49\pm0.22$ &                    $0.49\pm0.22$ &           $0.51\pm0.22$ &    $0.01\pm0.00$ &            $20/20$ \\
\arrayrulecolor{black!15}\cmidrule{2-6}
         & UCRL2 &           $0.29\pm0.38$ &                    $0.98\pm0.00$ &           $0.02\pm0.00$ &    $0.00\pm0.00$ &             $7/20$ \\
\arrayrulecolor{black!15}\cmidrule{1-6}
MG-Empty (2) & PSRL &           $0.00\pm0.00$ &                    $0.00\pm0.00$ &           $1.00\pm0.00$ &    $0.00\pm0.00$ &             $0/20$ \\
\arrayrulecolor{black!15}\cmidrule{2-6}
         & Q-learning &           $0.99\pm0.00$ &                    $0.99\pm0.00$ &           $0.01\pm0.00$ &    $0.01\pm0.00$ &            $20/20$ \\
\arrayrulecolor{black!15}\cmidrule{2-6}
         & UCRL2 &           $0.91\pm0.02$ &                    $0.98\pm0.00$ &           $0.02\pm0.00$ &    $0.00\pm0.00$ &             $9/20$ \\
\arrayrulecolor{black!15}\cmidrule{1-6}
MG-Empty (3) & PSRL &           $0.00\pm0.00$ &                    $0.40\pm0.49$ &           $0.60\pm0.49$ &    $0.00\pm0.00$ &             $0/20$ \\
\arrayrulecolor{black!15}\cmidrule{2-6}
         & Q-learning &           $1.00\pm0.00$ &                    $1.00\pm0.00$ &           $0.00\pm0.00$ &    $0.02\pm0.01$ &            $20/20$ \\
\arrayrulecolor{black!15}\cmidrule{2-6}
         & UCRL2 &           $0.96\pm0.01$ &                    $0.99\pm0.00$ &           $0.01\pm0.00$ &    $0.01\pm0.00$ &            $20/20$ \\
\arrayrulecolor{black!15}\cmidrule{1-6}
MG-Empty (4) & PSRL &           $0.00\pm0.00$ &                    $0.00\pm0.00$ &           $1.00\pm0.00$ &    $0.00\pm0.00$ &             $0/20$ \\
\arrayrulecolor{black!15}\cmidrule{2-6}
         & Q-learning &           $0.64\pm0.17$ &                    $0.65\pm0.17$ &           $0.35\pm0.17$ &    $0.02\pm0.01$ &            $20/20$ \\
\arrayrulecolor{black!15}\cmidrule{2-6}
         & UCRL2 &           $0.57\pm0.39$ &                    $0.99\pm0.00$ &           $0.01\pm0.00$ &    $0.01\pm0.00$ &            $20/20$ \\
\arrayrulecolor{black!15}\cmidrule{1-6}
MG-Empty (5) & PSRL &          $0.00\pm0.00$ &                   $0.00\pm0.00$ &           $1.00\pm0.00$ &    $0.00\pm0.00$ &             $0/20$ \\
\arrayrulecolor{black!15}\cmidrule{2-6}
         & Q-learning &           $0.25\pm0.21$ &                    $0.25\pm0.20$ &           $0.75\pm0.20$ &    $0.01\pm0.00$ &            $20/20$ \\
\arrayrulecolor{black!15}\cmidrule{2-6}
         & UCRL2 &           $0.00\pm0.00$ &                    $0.92\pm0.20$ &           $0.08\pm0.20$ &    $0.00\pm0.00$ &             $0/20$ \\
\arrayrulecolor{black!15}\cmidrule{1-6}
MG-Rooms (1) & PSRL &          $0.00\pm0.00$ &                    $0.00\pm0.00$ &           $1.00\pm0.00$ &    $0.00\pm0.00$ &             $0/20$ \\
\arrayrulecolor{black!15}\cmidrule{2-6}
         & Q-learning &           $0.99\pm0.01$ &                    $0.99\pm0.01$ &           $0.01\pm0.01$ &    $0.01\pm0.00$ &            $20/20$ \\
\arrayrulecolor{black!15}\cmidrule{2-6}
         & UCRL2 &           $0.00\pm0.00$ &                    $0.22\pm0.39$ &           $0.78\pm0.39$ &    $0.00\pm0.00$ &             $0/20$ \\
\arrayrulecolor{black!15}\cmidrule{1-6}
MG-Rooms (2) & PSRL &           $0.00\pm0.00$ &                    $0.00\pm0.00$ &           $1.00\pm0.00$ &    $0.00\pm0.00$ &             $0/20$ \\
\arrayrulecolor{black!15}\cmidrule{2-6}
         & Q-learning &           $0.99\pm0.01$ &                    $0.99\pm0.01$ &           $0.01\pm0.01$ &    $0.02\pm0.01$ &            $20/20$ \\
\arrayrulecolor{black!15}\cmidrule{2-6}
         & UCRL2 &           $0.85\pm0.19$ &                    $0.98\pm0.01$ &           $0.02\pm0.01$ &    $0.00\pm0.00$ &            $18/20$ \\
\arrayrulecolor{black!15}\cmidrule{1-6}
MG-Rooms (3) & PSRL &          $0.00\pm0.00$ &                    $0.00\pm0.00$ &           $1.00\pm0.00$ &    $0.00\pm0.00$ &             $0/20$ \\
\arrayrulecolor{black!15}\cmidrule{2-6}
         & Q-learning &           $0.98\pm0.02$ &                    $0.98\pm0.02$ &           $0.02\pm0.02$ &    $0.01\pm0.00$ &            $20/20$ \\
\arrayrulecolor{black!15}\cmidrule{2-6}
         & UCRL2 &           $0.00\pm0.00$ &                    $0.34\pm0.46$ &           $0.66\pm0.46$ &    $0.00\pm0.00$ &             $0/20$ \\
\arrayrulecolor{black!15}\cmidrule{1-6}
RiverSwim (1) & PSRL &           $0.85\pm0.36$ &                    $1.00\pm0.01$ &           $0.00\pm0.01$ &    $0.01\pm0.00$ &            $20/20$ \\
\arrayrulecolor{black!15}\cmidrule{2-6}
         & Q-learning &           $0.83\pm0.03$ &                    $0.84\pm0.03$ &           $0.16\pm0.03$ &    $0.03\pm0.01$ &            $20/20$ \\
\arrayrulecolor{black!15}\cmidrule{2-6}
         & UCRL2 &           $0.99\pm0.00$ &                    $1.00\pm0.00$ &           $0.00\pm0.00$ &    $0.02\pm0.00$ &            $20/20$ \\
\arrayrulecolor{black!15}\cmidrule{1-6}
RiverSwim (2) & PSRL &           $0.93\pm0.22$ &                    $0.99\pm0.00$ &           $0.01\pm0.00$ &    $0.01\pm0.00$ &            $20/20$ \\
\arrayrulecolor{black!15}\cmidrule{2-6}
         & Q-learning &           $0.66\pm0.13$ &                    $0.66\pm0.13$ &           $0.34\pm0.13$ &    $0.02\pm0.01$ &            $20/20$ \\
\arrayrulecolor{black!15}\cmidrule{2-6}
         & UCRL2 &           $0.95\pm0.02$ &                    $0.98\pm0.01$ &           $0.02\pm0.01$ &    $0.01\pm0.00$ &            $20/20$ \\
\arrayrulecolor{black!15}\cmidrule{1-6}
SimpleGrid (1) & PSRL &           $0.07\pm0.00$ &                    $0.07\pm0.00$ &           $0.93\pm0.00$ &    $0.00\pm0.00$ &             $0/20$ \\
\arrayrulecolor{black!15}\cmidrule{2-6}
         & Q-learning &           $0.89\pm0.01$ &                    $0.89\pm0.01$ &           $0.11\pm0.01$ &    $0.02\pm0.00$ &            $20/20$ \\
\arrayrulecolor{black!15}\cmidrule{2-6}
         & UCRL2 &           $0.56\pm0.37$ &                    $0.99\pm0.00$ &           $0.01\pm0.00$ &    $0.00\pm0.00$ &            $20/20$ \\
\arrayrulecolor{black!15}\cmidrule{1-6}
SimpleGrid (2) & PSRL &           $0.50\pm0.00$ &                    $0.55\pm0.15$ &           $0.45\pm0.15$ &    $0.00\pm0.00$ &             $0/20$ \\
\arrayrulecolor{black!15}\cmidrule{2-6}
         & Q-learning &           $0.99\pm0.00$ &                    $0.99\pm0.00$ &           $0.01\pm0.00$ &    $0.02\pm0.00$ &            $20/20$ \\
\arrayrulecolor{black!15}\cmidrule{2-6}
         & UCRL2 &           $0.61\pm0.18$ &                    $0.99\pm0.00$ &           $0.01\pm0.00$ &    $0.00\pm0.00$ &            $13/20$ \\
\arrayrulecolor{black!15}\cmidrule{1-6}
SimpleGrid (3) & PSRL &           $0.07\pm0.00$ &                    $0.07\pm0.00$ &           $0.93\pm0.00$ &    $0.00\pm0.00$ &             $0/20$ \\
\arrayrulecolor{black!15}\cmidrule{2-6}
         & Q-learning &           $0.85\pm0.01$ &                    $0.85\pm0.01$ &           $0.15\pm0.01$ &    $0.01\pm0.00$ &            $20/20$ \\
\arrayrulecolor{black!15}\cmidrule{2-6}
         & UCRL2 &           $0.07\pm0.00$ &                    $0.30\pm0.39$ &           $0.70\pm0.39$ &    $0.00\pm0.00$ &             $0/20$ \\
\arrayrulecolor{black!15}\cmidrule{1-6}
SimpleGrid (4) & PSRL &           $0.50\pm0.00$ &                    $0.50\pm0.00$ &           $0.50\pm0.00$ &    $0.00\pm0.00$ &             $0/20$ \\
\arrayrulecolor{black!15}\cmidrule{2-6}
         & Q-learning &           $0.99\pm0.00$ &                    $0.99\pm0.00$ &           $0.01\pm0.00$ &    $0.01\pm0.00$ &            $20/20$ \\
\arrayrulecolor{black!15}\cmidrule{2-6}
         & UCRL2 &           $0.50\pm0.00$ &                    $0.67\pm0.23$ &           $0.33\pm0.23$ &    $0.00\pm0.00$ &             $0/20$ \\
\arrayrulecolor{black!15}\cmidrule{1-6}
Taxi (1) & PSRL &           $0.04\pm0.03$ &                    $0.06\pm0.03$ &           $0.94\pm0.03$ &    $0.00\pm0.00$ &             $0/20$ \\
\arrayrulecolor{black!15}\cmidrule{2-6}
         & Q-learning &           $0.10\pm0.01$ &                    $0.05\pm0.00$ &           $0.95\pm0.00$ &    $0.01\pm0.00$ &            $20/20$ \\
\arrayrulecolor{black!15}\cmidrule{2-6}
         & UCRL2 &           $0.06\pm0.03$ &                    $0.88\pm0.01$ &           $0.12\pm0.01$ &    $0.01\pm0.00$ &            $20/20$ \\
\bottomrule
\end{tabular}}
    \label{tab:continuous_communicating_benchmark_result_table}
\end{table}

\begin{table}[htb]
    \centering
    \caption{Continuous ergodic benchmark final time step performance indicators.}
 \resizebox{\textwidth}{!}{%
\begin{tabular}{lllllll}
\toprule
         &       & Norm. cumulative reward & Norm. cumulative expected reward & Norm. cumulative regret & Steps per second & \# completed seeds \\
MDP & Agent &                         &                                  &                         &                  &                    \\
\midrule
DeepSea (1) & PSRL &           $0.80\pm0.24$ &                    $0.93\pm0.01$ &           $0.06\pm0.01$ &    $0.00\pm0.00$ &            $20/20$ \\
\arrayrulecolor{black!15}\cmidrule{2-6}
         & Q-learning &           $0.06\pm0.00$ &                    $0.06\pm0.00$ &           $0.94\pm0.00$ &    $0.01\pm0.00$ &            $20/20$ \\
\arrayrulecolor{black!15}\cmidrule{2-6}
         & UCRL2 &           $0.04\pm0.01$ &                    $0.74\pm0.04$ &           $0.23\pm0.05$ &    $0.01\pm0.00$ &            $20/20$ \\
\arrayrulecolor{black!15}\cmidrule{1-6}
FrozenLake (1) & PSRL &           $0.15\pm0.12$ &                    $0.98\pm0.03$ &           $0.01\pm0.03$ &    $0.01\pm0.00$ &            $20/20$ \\
\arrayrulecolor{black!15}\cmidrule{2-6}
         & Q-learning &           $0.16\pm0.03$ &                    $0.17\pm0.03$ &           $0.83\pm0.03$ &    $0.01\pm0.00$ &            $20/20$ \\
\arrayrulecolor{black!15}\cmidrule{2-6}
         & UCRL2 &           $0.41\pm0.15$ &                    $0.99\pm0.02$ &           $0.01\pm0.02$ &    $0.02\pm0.00$ &            $20/20$ \\
\arrayrulecolor{black!15}\cmidrule{1-6}
MG-Empty (1) & PSRL &           $0.01\pm0.01$ &                    $0.01\pm0.01$ &           $0.99\pm0.01$ &    $0.00\pm0.00$ &             $0/20$ \\
\arrayrulecolor{black!15}\cmidrule{2-6}
         & Q-learning &           $0.02\pm0.01$ &                    $0.02\pm0.02$ &           $0.98\pm0.02$ &    $0.00\pm0.00$ &             $0/20$ \\
\arrayrulecolor{black!15}\cmidrule{2-6}
         & UCRL2 &           $0.04\pm0.02$ &                    $0.95\pm0.06$ &           $0.05\pm0.06$ &    $0.00\pm0.00$ &             $0/20$ \\
\arrayrulecolor{black!15}\cmidrule{1-6}
MG-Empty (2) & PSRL &           $0.01\pm0.01$ &                    $0.02\pm0.04$ &           $0.98\pm0.04$ &    $0.00\pm0.00$ &             $0/20$ \\
\arrayrulecolor{black!15}\cmidrule{2-6}
         & Q-learning &           $0.02\pm0.02$ &                    $0.02\pm0.02$ &           $0.98\pm0.02$ &    $0.00\pm0.00$ &             $0/20$ \\
\arrayrulecolor{black!15}\cmidrule{2-6}
         & UCRL2 &           $0.03\pm0.03$ &                    $0.97\pm0.05$ &           $0.03\pm0.05$ &    $0.00\pm0.00$ &             $0/20$ \\
\arrayrulecolor{black!15}\cmidrule{1-6}
MG-Empty (3) & PSRL &           $0.03\pm0.02$ &                    $0.05\pm0.03$ &           $0.95\pm0.03$ &    $0.00\pm0.00$ &             $0/20$ \\
\arrayrulecolor{black!15}\cmidrule{2-6}
         & Q-learning &           $0.03\pm0.00$ &                    $0.03\pm0.00$ &           $0.97\pm0.00$ &    $0.00\pm0.00$ &            $14/20$ \\
\arrayrulecolor{black!15}\cmidrule{2-6}
         & UCRL2 &           $0.07\pm0.02$ &                    $0.96\pm0.01$ &           $0.04\pm0.01$ &    $0.00\pm0.00$ &             $8/20$ \\
\arrayrulecolor{black!15}\cmidrule{1-6}
MG-Empty (4) & PSRL &           $0.01\pm0.01$ &                    $0.01\pm0.01$ &           $0.99\pm0.01$ &    $0.00\pm0.00$ &             $0/20$ \\
\arrayrulecolor{black!15}\cmidrule{2-6}
         & Q-learning &           $0.02\pm0.01$ &                    $0.02\pm0.01$ &           $0.98\pm0.01$ &    $0.00\pm0.00$ &             $0/20$ \\
\arrayrulecolor{black!15}\cmidrule{2-6}
         & UCRL2 &           $0.01\pm0.00$ &                    $0.46\pm0.26$ &           $0.54\pm0.26$ &    $0.00\pm0.00$ &             $0/20$ \\
\arrayrulecolor{black!15}\cmidrule{1-6}
MG-Empty (5) & PSRL &           $0.03\pm0.06$ &                    $0.17\pm0.30$ &           $0.83\pm0.30$ &    $0.00\pm0.00$ &             $0/20$ \\
\arrayrulecolor{black!15}\cmidrule{2-6}
         & Q-learning &           $0.05\pm0.01$ &                    $0.04\pm0.01$ &           $0.96\pm0.01$ &    $0.00\pm0.00$ &            $12/20$ \\
\arrayrulecolor{black!15}\cmidrule{2-6}
         & UCRL2 &           $0.26\pm0.23$ &                    $0.99\pm0.00$ &           $0.01\pm0.00$ &    $0.00\pm0.00$ &            $19/20$ \\
\arrayrulecolor{black!15}\cmidrule{1-6}
MG-Empty (6) & PSRL &           $0.01\pm0.01$ &                    $0.01\pm0.02$ &           $0.99\pm0.02$ &    $0.00\pm0.00$ &             $0/20$ \\
\arrayrulecolor{black!15}\cmidrule{2-6}
         & Q-learning &           $0.02\pm0.02$ &                    $0.02\pm0.02$ &           $0.98\pm0.02$ &    $0.00\pm0.00$ &             $0/20$ \\
\arrayrulecolor{black!15}\cmidrule{2-6}
         & UCRL2 &           $0.01\pm0.00$ &                    $0.55\pm0.34$ &           $0.45\pm0.34$ &    $0.00\pm0.00$ &             $0/20$ \\
\arrayrulecolor{black!15}\cmidrule{1-6}
MG-Empty (7) & PSRL &           $0.01\pm0.01$ &                    $0.01\pm0.01$ &           $0.99\pm0.01$ &    $0.00\pm0.00$ &             $0/20$ \\
\arrayrulecolor{black!15}\cmidrule{2-6}
         & Q-learning &           $0.02\pm0.02$ &                    $0.02\pm0.03$ &           $0.98\pm0.03$ &    $0.00\pm0.00$ &             $0/20$ \\
\arrayrulecolor{black!15}\cmidrule{2-6}
         & UCRL2 &           $0.01\pm0.00$ &                    $0.73\pm0.32$ &           $0.27\pm0.32$ &    $0.00\pm0.00$ &             $0/20$ \\
\arrayrulecolor{black!15}\cmidrule{1-6}
MG-Empty (8) & PSRL &           $0.01\pm0.01$ &                    $0.01\pm0.01$ &           $0.99\pm0.01$ &    $0.00\pm0.00$ &             $0/20$ \\
\arrayrulecolor{black!15}\cmidrule{2-6}
         & Q-learning &           $0.02\pm0.01$ &                    $0.02\pm0.01$ &           $0.98\pm0.01$ &    $0.00\pm0.00$ &             $0/20$ \\
\arrayrulecolor{black!15}\cmidrule{2-6}
         & UCRL2 &           $0.01\pm0.00$ &                    $0.07\pm0.09$ &           $0.93\pm0.09$ &    $0.00\pm0.00$ &             $0/20$ \\
\arrayrulecolor{black!15}\cmidrule{1-6}
MG-Rooms (1) & PSRL &           $0.01\pm0.04$ &                    $0.01\pm0.02$ &           $0.99\pm0.02$ &    $0.00\pm0.00$ &             $0/20$ \\
\arrayrulecolor{black!15}\cmidrule{2-6}
         & Q-learning &           $0.04\pm0.02$ &                    $0.02\pm0.03$ &           $0.98\pm0.03$ &    $0.00\pm0.00$ &             $0/20$ \\
\arrayrulecolor{black!15}\cmidrule{2-6}
         & UCRL2 &           $0.01\pm0.00$ &                    $0.82\pm0.28$ &           $0.18\pm0.28$ &    $0.00\pm0.00$ &             $0/20$ \\
\arrayrulecolor{black!15}\cmidrule{1-6}
MG-Rooms (2) & PSRL &           $0.00\pm0.00$ &                    $0.00\pm0.00$ &           $1.00\pm0.00$ &    $0.00\pm0.00$ &             $0/20$ \\
\arrayrulecolor{black!15}\cmidrule{2-6}
         & Q-learning &           $0.01\pm0.02$ &                    $0.02\pm0.02$ &           $0.98\pm0.02$ &    $0.00\pm0.00$ &             $0/20$ \\
\arrayrulecolor{black!15}\cmidrule{2-6}
         & UCRL2 &           $0.00\pm0.00$ &                    $0.38\pm0.35$ &           $0.62\pm0.35$ &    $0.00\pm0.00$ &             $0/20$ \\
\arrayrulecolor{black!15}\cmidrule{1-6}
RiverSwim (1) & PSRL &           $0.77\pm0.38$ &                    $1.00\pm0.00$ &           $0.00\pm0.00$ &    $0.01\pm0.00$ &            $20/20$ \\
\arrayrulecolor{black!15}\cmidrule{2-6}
         & Q-learning &           $0.21\pm0.15$ &                    $0.27\pm0.19$ &           $0.73\pm0.19$ &    $0.01\pm0.01$ &            $20/20$ \\
\arrayrulecolor{black!15}\cmidrule{2-6}
         & UCRL2 &           $0.85\pm0.24$ &                    $1.00\pm0.00$ &           $0.00\pm0.00$ &    $0.02\pm0.00$ &            $20/20$ \\
\arrayrulecolor{black!15}\cmidrule{1-6}
RiverSwim (2) & PSRL &           $0.90\pm0.30$ &                    $1.00\pm0.00$ &           $0.00\pm0.00$ &    $0.01\pm0.00$ &            $20/20$ \\
\arrayrulecolor{black!15}\cmidrule{2-6}
         & Q-learning &           $0.29\pm0.21$ &                    $0.29\pm0.21$ &           $0.71\pm0.21$ &    $0.01\pm0.00$ &            $20/20$ \\
\arrayrulecolor{black!15}\cmidrule{2-6}
         & UCRL2 &           $0.90\pm0.20$ &                    $0.99\pm0.00$ &           $0.01\pm0.00$ &    $0.02\pm0.00$ &            $20/20$ \\
\arrayrulecolor{black!15}\cmidrule{1-6}
RiverSwim (3) & PSRL &           $0.88\pm0.24$ &                    $0.98\pm0.04$ &           $0.02\pm0.04$ &    $0.01\pm0.00$ &            $20/20$ \\
\arrayrulecolor{black!15}\cmidrule{2-6}
         & Q-learning &           $0.09\pm0.06$ &                    $0.10\pm0.06$ &           $0.90\pm0.06$ &    $0.01\pm0.00$ &            $20/20$ \\
\arrayrulecolor{black!15}\cmidrule{2-6}
         & UCRL2 &           $0.80\pm0.25$ &                    $0.99\pm0.01$ &           $0.01\pm0.01$ &    $0.01\pm0.00$ &            $20/20$ \\
\arrayrulecolor{black!15}\cmidrule{1-6}
RiverSwim (4) & PSRL &           $0.89\pm0.26$ &                    $0.99\pm0.00$ &           $0.01\pm0.00$ &    $0.01\pm0.00$ &            $20/20$ \\
\arrayrulecolor{black!15}\cmidrule{2-6}
         & Q-learning &           $0.82\pm0.20$ &                    $0.50\pm0.24$ &           $0.50\pm0.25$ &    $0.02\pm0.01$ &            $20/20$ \\
\arrayrulecolor{black!15}\cmidrule{2-6}
         & UCRL2 &           $0.82\pm0.29$ &                    $0.99\pm0.01$ &           $0.01\pm0.01$ &    $0.01\pm0.00$ &            $20/20$ \\
\arrayrulecolor{black!15}\cmidrule{1-6}
SimpleGrid (1) & PSRL &           $0.22\pm0.07$ &                    $0.30\pm0.19$ &           $0.70\pm0.19$ &    $0.00\pm0.00$ &             $0/20$ \\
\arrayrulecolor{black!15}\cmidrule{2-6}
         & Q-learning &           $0.22\pm0.00$ &                    $0.22\pm0.00$ &           $0.78\pm0.00$ &    $0.01\pm0.00$ &            $20/20$ \\
\arrayrulecolor{black!15}\cmidrule{2-6}
         & UCRL2 &           $0.24\pm0.04$ &                    $0.99\pm0.01$ &           $0.01\pm0.01$ &    $0.00\pm0.00$ &            $20/20$ \\
\arrayrulecolor{black!15}\cmidrule{1-6}
SimpleGrid (2) & PSRL &           $0.78\pm0.22$ &                    $0.99\pm0.02$ &           $0.01\pm0.02$ &    $0.00\pm0.00$ &            $20/20$ \\
\arrayrulecolor{black!15}\cmidrule{2-6}
         & Q-learning &           $0.67\pm0.04$ &                    $0.54\pm0.07$ &           $0.46\pm0.07$ &    $0.01\pm0.00$ &            $20/20$ \\
\arrayrulecolor{black!15}\cmidrule{2-6}
         & UCRL2 &           $0.79\pm0.16$ &                    $1.00\pm0.00$ &           $0.00\pm0.00$ &    $0.01\pm0.00$ &            $19/20$ \\
\arrayrulecolor{black!15}\cmidrule{1-6}
SimpleGrid (3) & PSRL &           $0.48\pm0.10$ &                    $0.57\pm0.16$ &           $0.43\pm0.16$ &    $0.00\pm0.00$ &             $0/20$ \\
\arrayrulecolor{black!15}\cmidrule{2-6}
         & Q-learning &           $0.51\pm0.00$ &                    $0.51\pm0.00$ &           $0.49\pm0.00$ &    $0.00\pm0.00$ &            $15/20$ \\
\arrayrulecolor{black!15}\cmidrule{2-6}
         & UCRL2 &           $0.50\pm0.00$ &                    $1.00\pm0.00$ &           $0.00\pm0.00$ &    $0.00\pm0.00$ &             $0/20$ \\
\arrayrulecolor{black!15}\cmidrule{1-6}
Taxi (1) & PSRL &           $0.13\pm0.08$ &                    $0.11\pm0.08$ &           $0.89\pm0.08$ &    $0.00\pm0.00$ &             $0/20$ \\
\arrayrulecolor{black!15}\cmidrule{2-6}
         & Q-learning &           $0.19\pm0.01$ &                    $0.13\pm0.01$ &           $0.87\pm0.01$ &    $0.01\pm0.00$ &            $20/20$ \\
\arrayrulecolor{black!15}\cmidrule{2-6}
         & UCRL2 &           $0.14\pm0.06$ &                    $0.91\pm0.01$ &           $0.09\pm0.01$ &    $0.00\pm0.00$ &            $18/20$ \\
\bottomrule
\end{tabular}}
    \label{tab:continuous_ergodic_benchmark_result_table}
\end{table}

\clearpage
\subsection{Non-tabular setting} \label{app:nt_benchmarking}

We now present the full results of the benchmarking procedure for the non-tabular baseline agents from \bsuite, ActorCritic, ActorCriticRNN, BootDQN, and DQN.

In order to keep the hardness induced by the emission map as low as possible while still testing the effective non-tabular capabilities of the agent, we employ the deterministic state information map, which provides clear state-identifying information.
The experimental procedure is the same as for the tabular case (described in the previous Section) with the only difference being that the total training time given to the agent is $40$ minutes both for the continuous and episodic settings.

The performances of the agents are in line with the results reported in \citet{osband2020bsuite}, with the exception of BootDQN.
Differently from \bsuite, the \colosseum benchmarking procedure penalizes particularly computationally expensive algorithms by limiting the training time.
Being BootDQN the most computationally intensive algorithm due to the ensemble training, this agent often breaks the time limit, which consequently worsens its overall performance.

Tables~\ref{tab:nt_benchmark_table_e}a,~\ref{tab:nt_benchmark_table_e}b,~\ref{tab:nt_benchmark_table_e}c, and ~\ref{tab:nt_benchmark_table_e}d
report a summary of the performance of the agents in terms of normalized cumulative regrets.
DQN is the best performing agent on average for all the different settings.
Unsurprisingly, it performs relatively better in the ergodic setting compared to the communicating one, in which the exploration challenge is harder.
This is especially clear from the results for the \texttt{DeepSea} family in the continuous setting.
DQN is not able to perform well in any of the instances in the communicating case (Table~\ref{tab:nt_benchmark_table_e}c).
Although BootDQN generally performs better than the ActorCritic agents, it is not able to fully express its potential due to the associated computational burden.
Interestingly, the ActorCritic agent without the recurrent component performs better than the version equipped with it.
Note that this is not due to a different computational cost as ActorCriticRNN is always able to complete the interactions with the MDPs within the given time limit.

\begin{figure}[htb!]
    \centering
    \includegraphics[width=\linewidth]{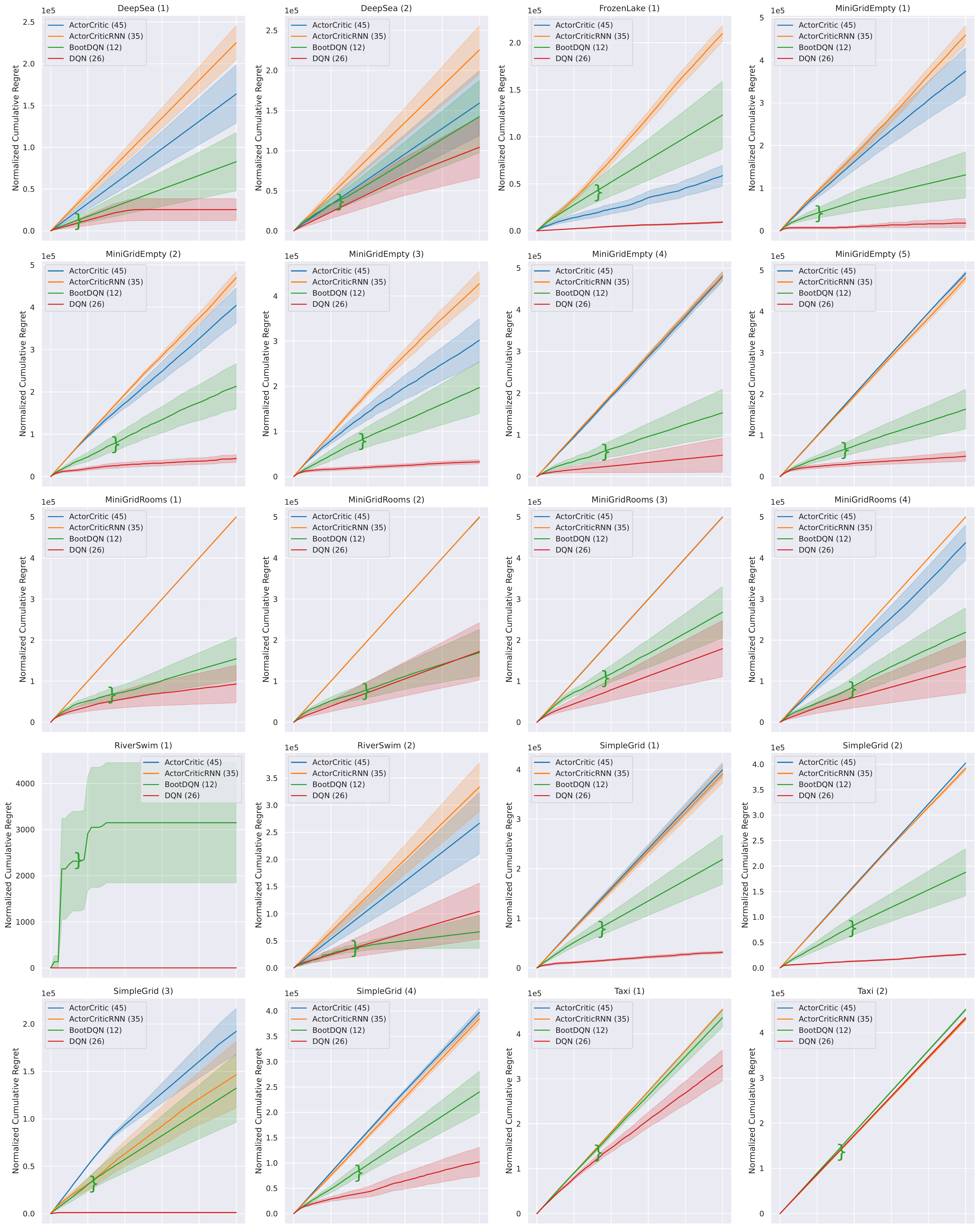}
\caption{Full interaction performances for the non-tabular baseline agents in the episodic communicating setting.}
    \label{fig:nt_bench_ec}
\end{figure}

\begin{figure}[htb!]
    \centering
    \includegraphics[width=\linewidth]{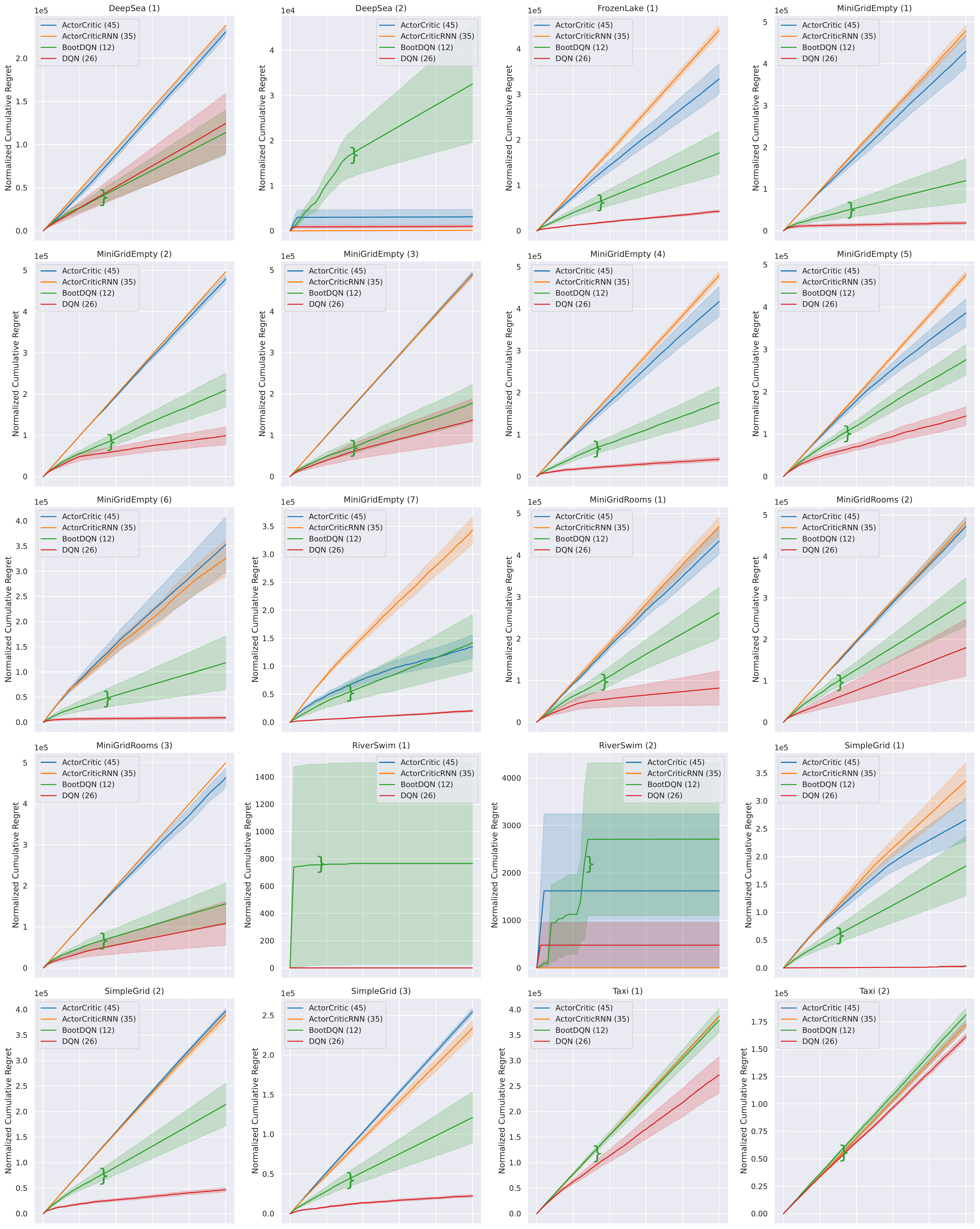}
\caption{Full interaction performances for the non-tabular baseline agents in the episodic ergodic setting.}
    \label{fig:nt_bench_ee}
\end{figure}

\begin{figure}[htb!]
    \centering
    \includegraphics[width=1\linewidth]{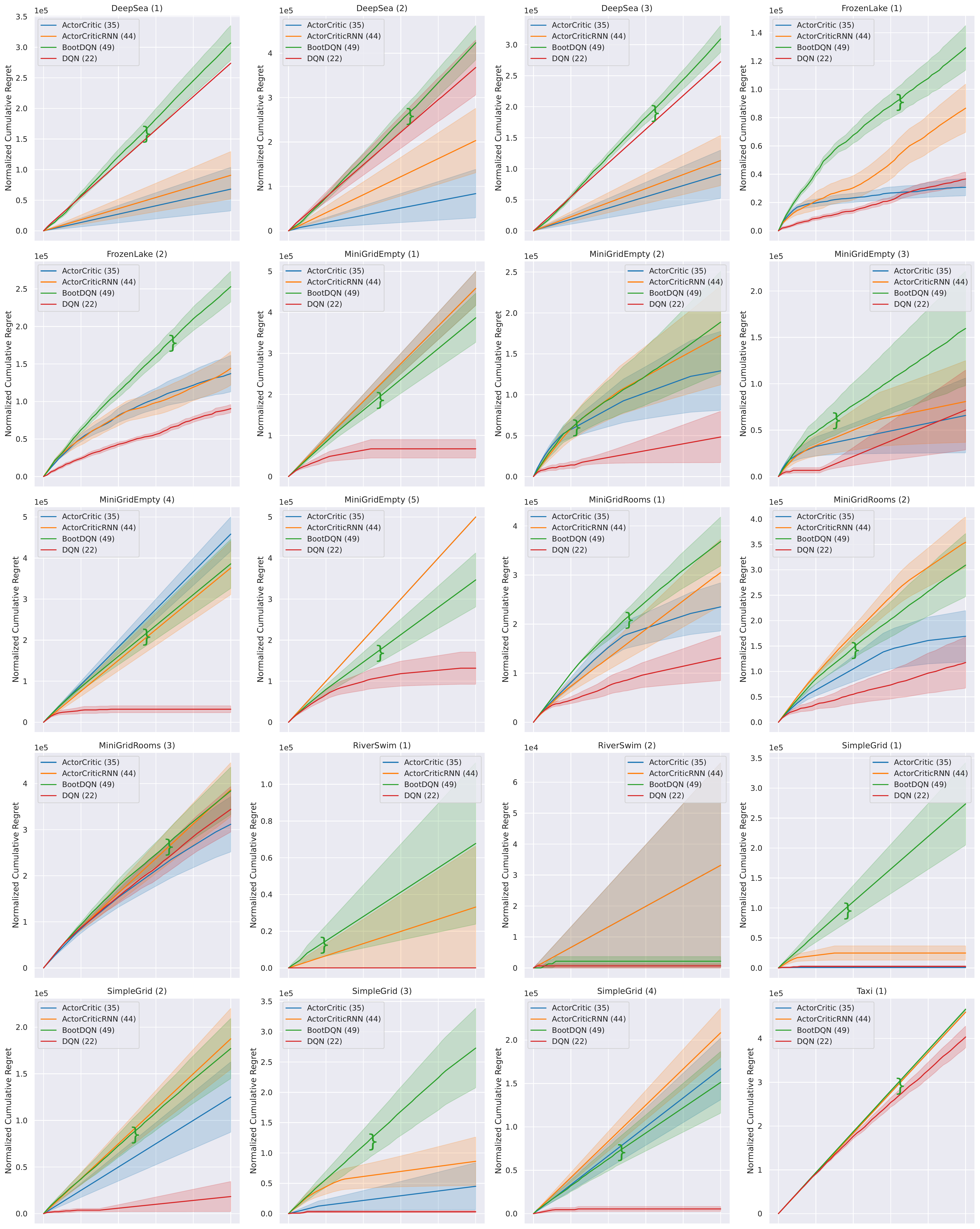}
\caption{Full interaction performances for the non-tabular baseline agents in the continuous communicating setting.}
    \label{fig:nt_bench_cc}
\end{figure}

\begin{figure}[htb!]
    \centering
    \includegraphics[width=1\linewidth]{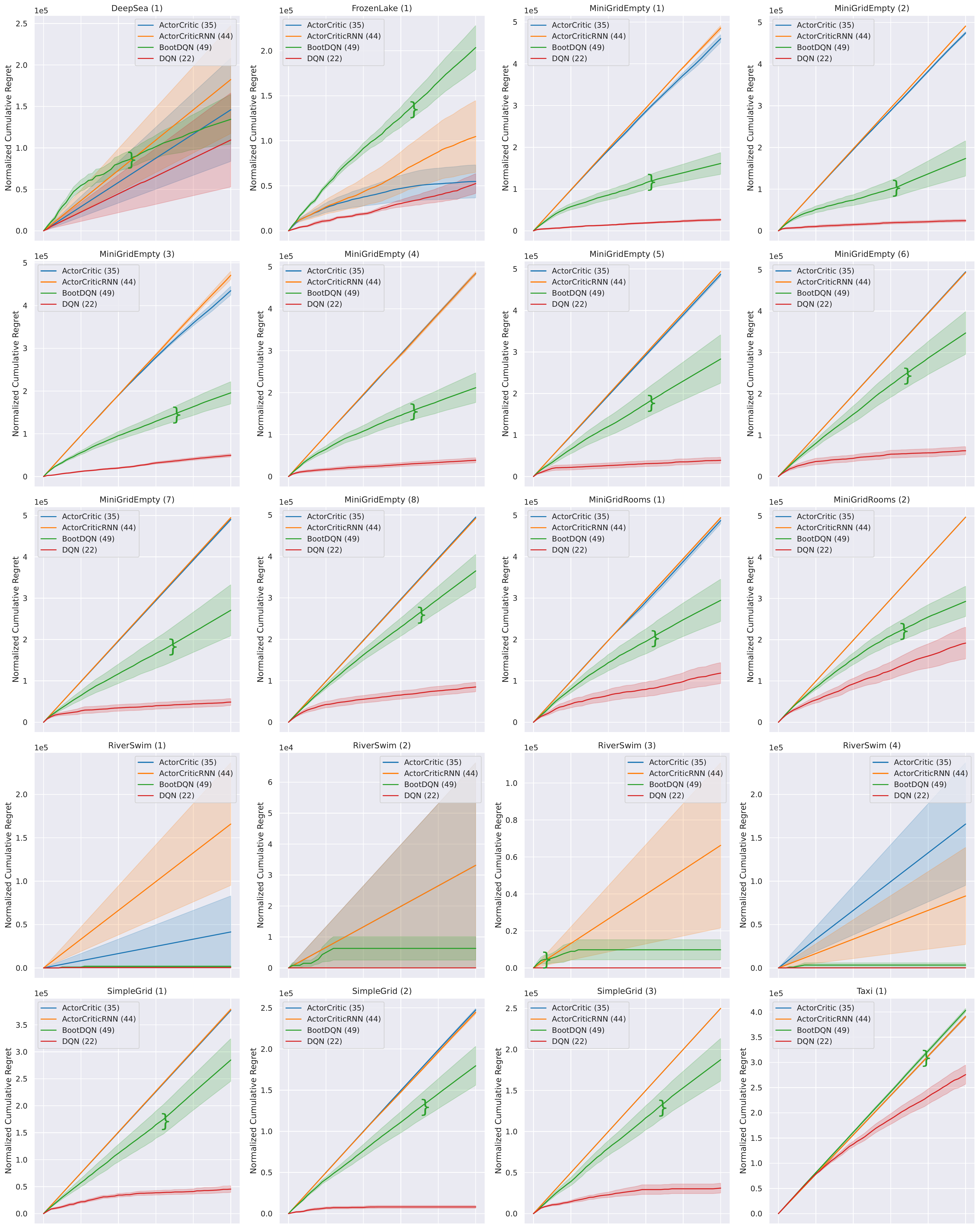}
\caption{Full interaction performances for the non-tabular baseline agents in the continuous ergodic setting.}
    \label{fig:nt_bench_ce}
\end{figure}

Figure~\ref{fig:nt_bench_ec},~\ref{fig:nt_bench_ee},~\ref{fig:nt_bench_cc}, and ~\ref{fig:nt_bench_ce}
show how the cumulative regret, along with standard error intervals, of the agents evolve during their interaction with the benchmark environments.
Note that the $\}$ symbol is always reported for BootDQN, meaning that, for at least one of the seeds, the agent ran out of time.
We note that, overall, the variability of the performances across the different seeds is low, with the exception of a few cases.
In particular, the \texttt{RiverSwim} family is associated with moderate variability.
This is not surprising since, for this MDP family, the agent can easily get trapped in a sub-optimal policy if it does not fully explore the state space.

\begin{figure}[htb!]
    \captionsetup[subfloat]{position=bottom}
    \centering
    \subfloat[][Episodic communicating setting.]{
    \includegraphics[width=\linewidth]{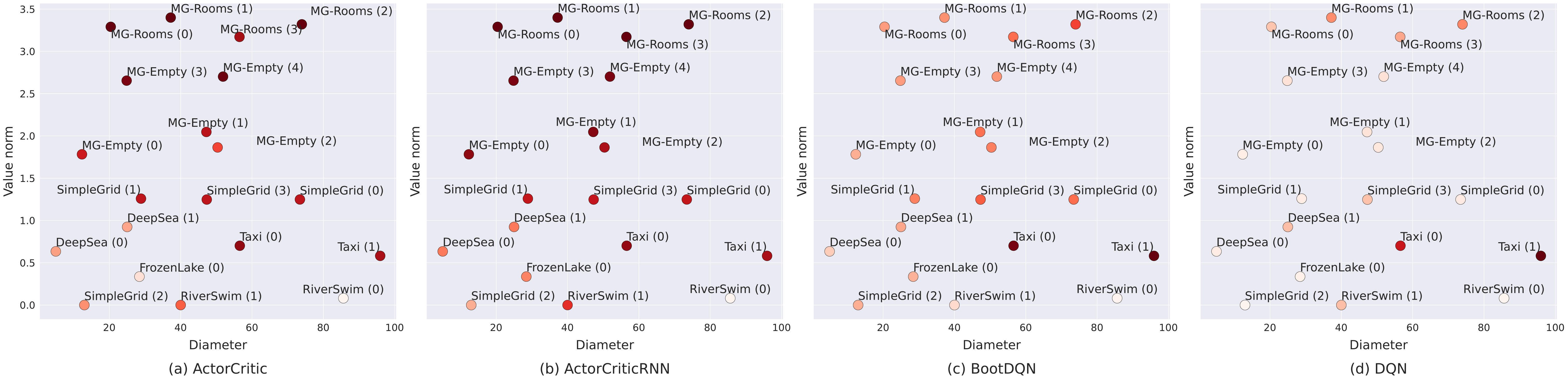}}\\
    \subfloat[][Episodic ergodic setting.]{
    \includegraphics[width=\linewidth]{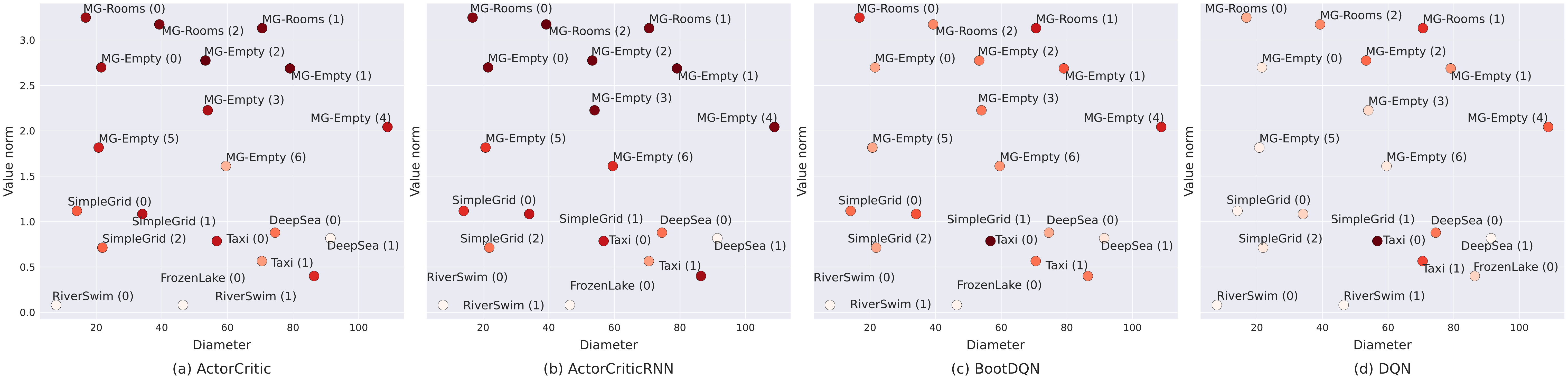}}\\
    \subfloat[][Continuous communicating setting.]{
    \includegraphics[width=\linewidth]{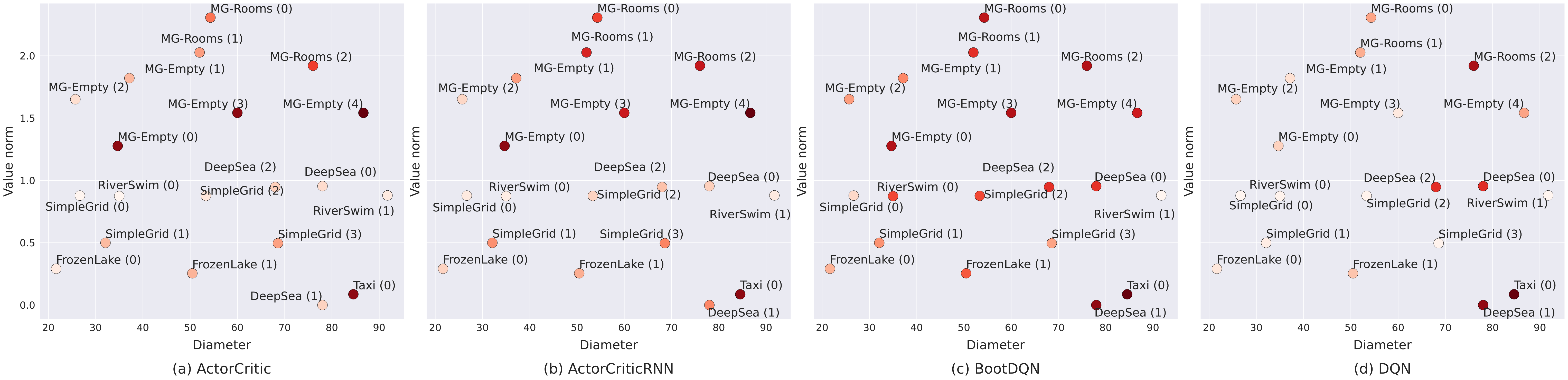}}\\
    \subfloat[][Continuous ergodic setting.]{
    \includegraphics[width=\linewidth]{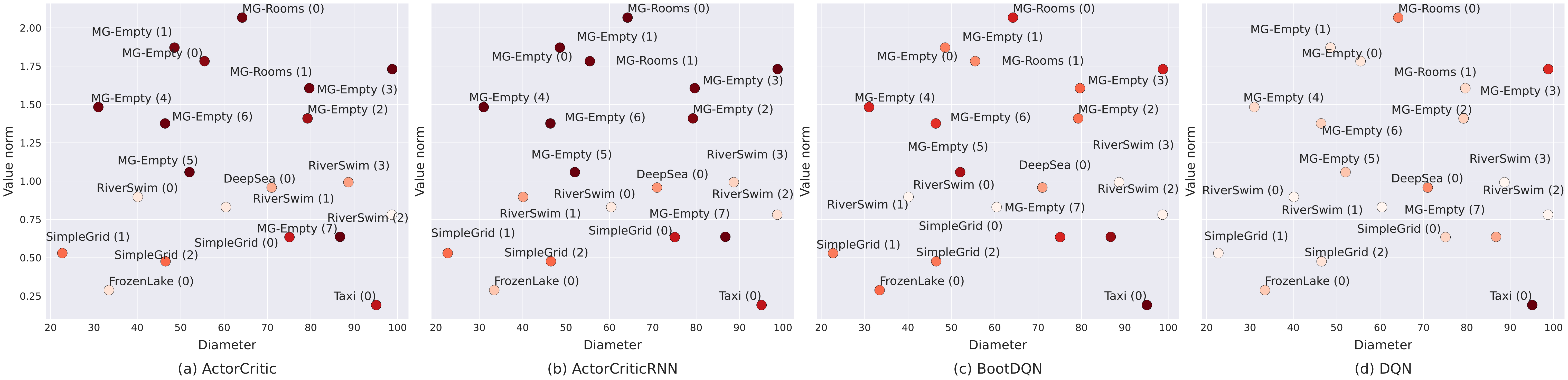}}
\caption{Average cumulative regret obtained by the agents in the continuous ergodic setting placed according to the diameter and the value norm values of the benchmark MDPs.} 
\label{fig:nt_hardness_and_crs}
\end{figure}

Figures~\ref{fig:nt_hardness_and_crs}a, ~\ref{fig:nt_hardness_and_crs}b, ~\ref{fig:nt_hardness_and_crs}c, and ~\ref{fig:nt_hardness_and_crs}d
place the regret of the agents on a position corresponding to the diameter and value norm of each MDP respectively for the
episodic communicating,
episodic ergodic,
continuous communicating, and
continuous ergodic settings.
Overall, we observe that the harder environments effectively induce higher regret.
This relationship is visibly stronger when the agents perform better across the environments, (see the ActorCritic agents in Figure~\ref{fig:nt_hardness_and_crs}c and DQN in all settings).
Interestingly, and similarly to the tabular case, while the best performing agent (DQN) is evidently impacted more by the diameter, the opposite holds for the other agents. Regardless of the visitation complexity, this suggests that an agent that fails to handle the estimation complexity of an environment is bound to perform badly both in tabular and non-tabular settings.

\begin{table}[htb!]
  \centering
  \caption{Summary of benchmark results for the non-tabular \bsuite baselines.}%
  \subfloat[][Episodic communicating setting.]{
  \resizebox{0.5\textwidth}{!}{%
\begin{tabular}{lcccc}
\toprule
{} &             ActorCritic &          ActorCriticRNN &                 BootDQN &                     DQN \\
\midrule
DeepSea          &           $0.33\pm0.24$ &           $0.45\pm0.14$ &           $0.17\pm0.24$ &  $\mathbf{0.05}\pm0.09$ \\
                 &           $0.32\pm0.28$ &           $0.45\pm0.21$ &           $0.28\pm0.31$ &  $\mathbf{0.21}\pm0.26$ \\
\arrayrulecolor{black!15}\midrule%
FrozenLake       &           $0.12\pm0.08$ &           $0.42\pm0.06$ &           $0.25\pm0.25$ &  $\mathbf{0.02}\pm0.01$ \\
\arrayrulecolor{black!15}\midrule%
MG-Empty    &           $0.75\pm0.38$ &           $0.92\pm0.15$ &           $0.26\pm0.38$ &  $\mathbf{0.04}\pm0.07$ \\
                 &           $0.81\pm0.29$ &           $0.94\pm0.11$ &           $0.43\pm0.37$ &  $\mathbf{0.09}\pm0.06$ \\
                 &           $0.60\pm0.33$ &           $0.85\pm0.19$ &           $0.39\pm0.40$ &  $\mathbf{0.07}\pm0.03$ \\
                 &           $0.96\pm0.08$ &           $0.96\pm0.07$ &           $0.31\pm0.39$ &  $\mathbf{0.10}\pm0.28$ \\
                 &           $0.99\pm0.03$ &           $0.96\pm0.06$ &           $0.33\pm0.33$ &  $\mathbf{0.10}\pm0.09$ \\
\arrayrulecolor{black!15}\midrule%
MG-Rooms    &           $1.00\pm0.00$ &           $1.00\pm0.00$ &           $0.31\pm0.37$ &  $\mathbf{0.19}\pm0.32$ \\
                 &           $1.00\pm0.01$ &           $1.00\pm0.00$ &  $\mathbf{0.34}\pm0.39$ &  $\mathbf{0.34}\pm0.48$ \\
                 &           $1.00\pm0.00$ &           $1.00\pm0.01$ &           $0.54\pm0.43$ &  $\mathbf{0.36}\pm0.48$ \\
                 &           $0.87\pm0.30$ &           $1.00\pm0.00$ &           $0.44\pm0.41$ &  $\mathbf{0.27}\pm0.44$ \\
\arrayrulecolor{black!15}\midrule%
RiverSwim        &  $\mathbf{0.00}\pm0.00$ &  $\mathbf{0.00}\pm0.00$ &           $0.01\pm0.01$ &  $\mathbf{0.00}\pm0.00$ \\
                 &           $0.53\pm0.39$ &           $0.67\pm0.31$ &  $\mathbf{0.13}\pm0.21$ &           $0.21\pm0.36$ \\
\arrayrulecolor{black!15}\midrule%
SimpleGrid       &           $0.80\pm0.10$ &           $0.78\pm0.13$ &           $0.44\pm0.35$ &  $\mathbf{0.06}\pm0.02$ \\
                 &           $0.80\pm0.01$ &           $0.78\pm0.03$ &           $0.38\pm0.32$ &  $\mathbf{0.05}\pm0.01$ \\
                 &           $0.38\pm0.17$ &           $0.29\pm0.24$ &           $0.26\pm0.25$ &  $\mathbf{0.00}\pm0.00$ \\
                 &           $0.79\pm0.06$ &           $0.77\pm0.07$ &           $0.48\pm0.28$ &  $\mathbf{0.20}\pm0.20$ \\
\arrayrulecolor{black!15}\midrule%
Taxi             &           $0.91\pm0.01$ &           $0.91\pm0.01$ &           $0.87\pm0.14$ &  $\mathbf{0.66}\pm0.24$ \\
                 &  $\mathbf{0.86}\pm0.01$ &  $\mathbf{0.86}\pm0.01$ &           $0.90\pm0.02$ &           $0.87\pm0.01$ \\
\arrayrulecolor{black!30}\midrule%
\textit{Average} &           $0.69\pm0.30$ &           $0.75\pm0.28$ &           $0.38\pm0.21$ &  $\mathbf{0.19}\pm0.22$ \\
\arrayrulecolor{black!15}\midrule%
\end{tabular}}
  }%
  \subfloat[][Episodic ergodic setting.]{
 \resizebox{0.5\textwidth}{!}{%
\begin{tabular}{lcccc}
\toprule
{} &             ActorCritic &          ActorCriticRNN &                 BootDQN &                     DQN \\
\midrule
DeepSea          &           $0.46\pm0.04$ &           $0.48\pm0.00$ &  $\mathbf{0.23}\pm0.18$ &           $0.25\pm0.24$ \\
                 &           $0.01\pm0.01$ &  $\mathbf{0.00}\pm0.00$ &           $0.07\pm0.09$ &  $\mathbf{0.00}\pm0.00$ \\
\arrayrulecolor{black!15}\midrule%
FrozenLake       &           $0.67\pm0.23$ &           $0.88\pm0.07$ &           $0.34\pm0.33$ &  $\mathbf{0.09}\pm0.02$ \\
\arrayrulecolor{black!15}\midrule%
MG-Empty    &           $0.86\pm0.27$ &           $0.96\pm0.09$ &           $0.24\pm0.36$ &  $\mathbf{0.04}\pm0.03$ \\
                 &           $0.96\pm0.07$ &           $0.99\pm0.01$ &           $0.42\pm0.29$ &  $\mathbf{0.20}\pm0.15$ \\
                 &           $0.98\pm0.05$ &           $0.98\pm0.03$ &           $0.35\pm0.31$ &  $\mathbf{0.27}\pm0.36$ \\
                 &           $0.83\pm0.25$ &           $0.96\pm0.07$ &           $0.35\pm0.26$ &  $\mathbf{0.08}\pm0.03$ \\
                 &           $0.77\pm0.23$ &           $0.95\pm0.06$ &           $0.55\pm0.25$ &  $\mathbf{0.29}\pm0.15$ \\
                 &           $0.71\pm0.38$ &           $0.65\pm0.25$ &           $0.24\pm0.37$ &  $\mathbf{0.02}\pm0.02$ \\
                 &           $0.27\pm0.14$ &           $0.69\pm0.16$ &           $0.28\pm0.35$ &  $\mathbf{0.04}\pm0.01$ \\
\arrayrulecolor{black!15}\midrule%
MG-Rooms    &           $0.87\pm0.22$ &           $0.94\pm0.16$ &           $0.52\pm0.43$ &  $\mathbf{0.16}\pm0.28$ \\
                 &           $0.94\pm0.16$ &           $0.96\pm0.08$ &           $0.58\pm0.40$ &  $\mathbf{0.36}\pm0.47$ \\
                 &           $0.93\pm0.16$ &           $1.00\pm0.00$ &           $0.31\pm0.35$ &  $\mathbf{0.22}\pm0.37$ \\
\arrayrulecolor{black!15}\midrule%
RiverSwim        &  $\mathbf{0.00}\pm0.00$ &  $\mathbf{0.00}\pm0.00$ &  $\mathbf{0.00}\pm0.01$ &  $\mathbf{0.00}\pm0.00$ \\
                 &  $\mathbf{0.00}\pm0.01$ &  $\mathbf{0.00}\pm0.00$ &           $0.01\pm0.01$ &  $\mathbf{0.00}\pm0.00$ \\
\arrayrulecolor{black!15}\midrule%
SimpleGrid       &           $0.53\pm0.27$ &           $0.67\pm0.23$ &           $0.37\pm0.37$ &  $\mathbf{0.01}\pm0.01$ \\
                 &           $0.79\pm0.04$ &           $0.78\pm0.06$ &           $0.43\pm0.29$ &  $\mathbf{0.09}\pm0.03$ \\
                 &           $0.51\pm0.03$ &           $0.47\pm0.07$ &           $0.24\pm0.22$ &  $\mathbf{0.04}\pm0.01$ \\
\arrayrulecolor{black!15}\midrule%
Taxi             &           $0.77\pm0.01$ &           $0.77\pm0.01$ &           $0.76\pm0.16$ &  $\mathbf{0.54}\pm0.24$ \\
                 &           $0.34\pm0.02$ &           $0.34\pm0.02$ &           $0.36\pm0.04$ &  $\mathbf{0.32}\pm0.02$ \\
\arrayrulecolor{black!30}\midrule%
\textit{Average} &           $0.61\pm0.32$ &           $0.67\pm0.34$ &           $0.33\pm0.18$ &  $\mathbf{0.15}\pm0.15$ \\
\arrayrulecolor{black!15}\midrule%
\end{tabular}}
  }\\
  \subfloat[][Continuous communicating setting.]{
  \resizebox{0.5\textwidth}{!}{%
\begin{tabular}{lcccc}
\toprule
{} &             ActorCritic & ActorCriticRNN &                 BootDQN &                     DQN \\
\midrule
DeepSea          &  $\mathbf{0.14}\pm0.25$ &  $0.18\pm0.27$ &           $0.61\pm0.20$ &           $0.55\pm0.00$ \\
                 &  $\mathbf{0.17}\pm0.38$ &  $0.41\pm0.50$ &           $0.85\pm0.27$ &           $0.74\pm0.43$ \\
                 &  $\mathbf{0.18}\pm0.27$ &  $0.23\pm0.28$ &           $0.62\pm0.15$ &           $0.54\pm0.00$ \\
\arrayrulecolor{black!15}\midrule%
FrozenLake       &  $\mathbf{0.06}\pm0.04$ &  $0.17\pm0.12$ &           $0.26\pm0.11$ &           $0.07\pm0.03$ \\
                 &           $0.27\pm0.17$ &  $0.29\pm0.16$ &           $0.51\pm0.14$ &  $\mathbf{0.18}\pm0.04$ \\
\arrayrulecolor{black!15}\midrule%
MG-Empty    &           $0.92\pm0.29$ &  $0.92\pm0.29$ &           $0.77\pm0.41$ &  $\mathbf{0.14}\pm0.16$ \\
                 &           $0.26\pm0.33$ &  $0.34\pm0.42$ &           $0.38\pm0.43$ &  $\mathbf{0.10}\pm0.22$ \\
                 &  $\mathbf{0.13}\pm0.28$ &  $0.16\pm0.30$ &           $0.32\pm0.43$ &           $0.14\pm0.30$ \\
                 &           $0.92\pm0.29$ &  $0.75\pm0.45$ &           $0.77\pm0.41$ &  $\mathbf{0.06}\pm0.06$ \\
                 &           $1.00\pm0.00$ &  $1.00\pm0.00$ &           $0.69\pm0.46$ &  $\mathbf{0.26}\pm0.27$ \\
\arrayrulecolor{black!15}\midrule%
MG-Rooms    &           $0.47\pm0.34$ &  $0.61\pm0.47$ &           $0.74\pm0.35$ &  $\mathbf{0.26}\pm0.32$ \\
                 &           $0.34\pm0.35$ &  $0.71\pm0.34$ &           $0.62\pm0.43$ &  $\mathbf{0.24}\pm0.35$ \\
                 &  $\mathbf{0.62}\pm0.42$ &  $0.77\pm0.41$ &           $0.77\pm0.36$ &           $0.69\pm0.34$ \\
\arrayrulecolor{black!15}\midrule%
RiverSwim        &  $\mathbf{0.00}\pm0.00$ &  $0.07\pm0.23$ &           $0.14\pm0.30$ &  $\mathbf{0.00}\pm0.00$ \\
                 &           $0.07\pm0.23$ &  $0.07\pm0.23$ &  $\mathbf{0.00}\pm0.01$ &  $\mathbf{0.00}\pm0.00$ \\
\arrayrulecolor{black!15}\midrule%
SimpleGrid       &  $\mathbf{0.00}\pm0.01$ &  $0.05\pm0.08$ &           $0.55\pm0.48$ &           $0.01\pm0.01$ \\
                 &           $0.25\pm0.26$ &  $0.38\pm0.23$ &           $0.35\pm0.22$ &  $\mathbf{0.04}\pm0.11$ \\
                 &           $0.09\pm0.27$ &  $0.17\pm0.28$ &           $0.55\pm0.45$ &  $\mathbf{0.01}\pm0.01$ \\
                 &           $0.33\pm0.25$ &  $0.42\pm0.19$ &           $0.30\pm0.25$ &  $\mathbf{0.01}\pm0.02$ \\
\arrayrulecolor{black!15}\midrule%
Taxi             &           $0.92\pm0.01$ &  $0.92\pm0.01$ &           $0.93\pm0.02$ &  $\mathbf{0.81}\pm0.17$ \\
\arrayrulecolor{black!30}\midrule%
\textit{Average} &           $0.36\pm0.33$ &  $0.43\pm0.31$ &           $0.54\pm0.24$ &  $\mathbf{0.24}\pm0.26$ \\
\arrayrulecolor{black!15}\midrule%
\end{tabular}}
  }%
  \subfloat[][Continuous ergodic setting.]{
 \resizebox{0.5\textwidth}{!}{%
\begin{tabular}{lcccc}
\toprule
{} &             ActorCritic & ActorCriticRNN &                 BootDQN &                     DQN \\
\midrule
DeepSea          &           $0.29\pm0.43$ &  $0.36\pm0.45$ &           $0.27\pm0.21$ &  $\mathbf{0.22}\pm0.39$ \\
\arrayrulecolor{black!15}\midrule%
FrozenLake       &  $\mathbf{0.11}\pm0.13$ &  $0.21\pm0.28$ &           $0.41\pm0.17$ &  $\mathbf{0.11}\pm0.08$ \\
\arrayrulecolor{black!15}\midrule%
MG-Empty    &           $0.92\pm0.07$ &  $0.97\pm0.04$ &           $0.32\pm0.18$ &  $\mathbf{0.05}\pm0.02$ \\
                 &           $0.95\pm0.02$ &  $0.98\pm0.01$ &           $0.35\pm0.29$ &  $\mathbf{0.05}\pm0.02$ \\
                 &           $0.87\pm0.07$ &  $0.94\pm0.06$ &           $0.39\pm0.18$ &  $\mathbf{0.10}\pm0.02$ \\
                 &           $0.97\pm0.02$ &  $0.97\pm0.03$ &           $0.42\pm0.24$ &  $\mathbf{0.08}\pm0.04$ \\
                 &           $0.97\pm0.03$ &  $0.99\pm0.00$ &           $0.57\pm0.40$ &  $\mathbf{0.08}\pm0.05$ \\
                 &           $0.99\pm0.01$ &  $0.99\pm0.01$ &           $0.69\pm0.36$ &  $\mathbf{0.12}\pm0.07$ \\
                 &           $0.98\pm0.02$ &  $0.99\pm0.01$ &           $0.54\pm0.43$ &  $\mathbf{0.10}\pm0.06$ \\
                 &           $0.99\pm0.00$ &  $0.98\pm0.01$ &           $0.73\pm0.28$ &  $\mathbf{0.17}\pm0.08$ \\
\arrayrulecolor{black!15}\midrule%
MG-Rooms    &           $0.97\pm0.05$ &  $0.99\pm0.01$ &           $0.59\pm0.35$ &  $\mathbf{0.24}\pm0.17$ \\
                 &           $0.99\pm0.00$ &  $0.99\pm0.00$ &           $0.59\pm0.25$ &  $\mathbf{0.38}\pm0.27$ \\
\arrayrulecolor{black!15}\midrule%
RiverSwim        &           $0.08\pm0.29$ &  $0.33\pm0.49$ &  $\mathbf{0.00}\pm0.01$ &  $\mathbf{0.00}\pm0.00$ \\
                 &           $0.07\pm0.23$ &  $0.07\pm0.23$ &           $0.01\pm0.03$ &  $\mathbf{0.00}\pm0.00$ \\
                 &  $\mathbf{0.00}\pm0.00$ &  $0.13\pm0.31$ &           $0.02\pm0.04$ &  $\mathbf{0.00}\pm0.00$ \\
                 &           $0.33\pm0.49$ &  $0.17\pm0.39$ &           $0.01\pm0.02$ &  $\mathbf{0.00}\pm0.00$ \\
\arrayrulecolor{black!15}\midrule%
SimpleGrid       &           $0.75\pm0.01$ &  $0.76\pm0.01$ &           $0.57\pm0.27$ &  $\mathbf{0.09}\pm0.04$ \\
                 &           $0.49\pm0.01$ &  $0.49\pm0.01$ &           $0.36\pm0.16$ &  $\mathbf{0.02}\pm0.01$ \\
                 &           $0.50\pm0.00$ &  $0.50\pm0.00$ &           $0.37\pm0.18$ &  $\mathbf{0.06}\pm0.04$ \\
\arrayrulecolor{black!15}\midrule%
Taxi             &           $0.78\pm0.02$ &  $0.78\pm0.02$ &           $0.81\pm0.03$ &  $\mathbf{0.55}\pm0.13$ \\
\arrayrulecolor{black!30}\midrule%
\textit{Average} &           $0.65\pm0.36$ &  $0.68\pm0.34$ &           $0.40\pm0.24$ &  $\mathbf{0.12}\pm0.14$ \\
\arrayrulecolor{black!15}\midrule%
\end{tabular}}
  }
  \label{tab:nt_benchmark_table_e}%
\end{table}

\end{document}